\def\year{2022}\relax
   \documentclass[letterpaper]{article}
   \usepackage{aaai22} % DO NOT CHANGE THIS
   \usepackage{times} % DO NOT CHANGE THIS
   \usepackage{helvet} % DO NOT CHANGE THIS
   \usepackage{courier} % DO NOT CHANGE THIS
   \usepackage[hyphens]{url} % DO NOT CHANGE THIS
   \usepackage{graphicx} % DO NOT CHANGE THIS
   \usepackage{graphicx}  % DO NOT CHANGE THIS
   \usepackage{natbib}  % DO NOT CHANGE THIS
   \usepackage{caption}  % DO NOT CHANGE THIS
   \DeclareCaptionStyle{ruled}%
{labelfont=normalfont,labelsep=colon,strut=off} 

\usepackage[utf8]{inputenc}
\newcommand{\nciteyear}[1]{[\citeyear{#1}]}
\newcommand{\fullv}[1]{#1}
\newcommand{\shortv}{\commentout}

\usepackage{amsmath}
\usepackage{todonotes}
\usepackage{amssymb}

\usepackage{enumitem}

\newtheorem{THEOREM}{Theorem}[section]
\newenvironment{theorem}{\begin{THEOREM} \hspace{-.85em} {\bf :} }%
                        {\end{THEOREM}}
\newtheorem{LEMMA}[THEOREM]{Lemma}
\newenvironment{lemma}{\begin{LEMMA} \hspace{-.85em} {\bf :} }%
                      {\end{LEMMA}}
\newtheorem{COROLLARY}[THEOREM]{Corollary}
\newenvironment{corollary}{\begin{COROLLARY} \hspace{-.85em} {\bf :} }%
                          {\end{COROLLARY}}
\newtheorem{PROPOSITION}[THEOREM]{Proposition}
\newenvironment{proposition}{\begin{PROPOSITION} \hspace{-.85em} {\bf :} }%
                            {\end{PROPOSITION}}
\newtheorem{DEFINITION}[THEOREM]{Definition}
\newenvironment{definition}{\begin{DEFINITION} \hspace{-.85em} {\bf :} \rm}%
                            {\end{DEFINITION}}
\newtheorem{CLAIM}[THEOREM]{Claim}
\newenvironment{claim}{\begin{CLAIM} \hspace{-.85em} {\bf :} \rm}%
                            {\end{CLAIM}}
\newtheorem{EXAMPLE}[THEOREM]{Example}
\newenvironment{example}{\begin{EXAMPLE} \hspace{-.85em} {\bf :} \rm}%
                            {\end{EXAMPLE}}
\newtheorem{REMARK}[THEOREM]{Remark}
\newenvironment{remark}{\begin{REMARK} \hspace{-.85em} {\bf :} \rm}%
                            {\end{REMARK}}
\newcommand{\thm}{\begin{theorem}}
\newcommand{\lem}{\begin{lemma}}
\newcommand{\pro}{\begin{proposition}}
\newcommand{\dfn}{\begin{definition}}
\newcommand{\rem}{\begin{remark}}
\newcommand{\xam}{\begin{example}}
\newcommand{\cor}{\begin{corollary}}
\newcommand{\prf}{\noindent{\bf Proof:} }
\newcommand{\ethm}{\end{theorem}}
\newcommand{\elem}{\end{lemma}}
\newcommand{\epro}{\end{proposition}}
\newcommand{\edfn}{\bbox\end{definition}}
\newcommand{\erem}{\bbox\end{remark}}
\newcommand{\exam}{\bbox\end{example}}
\newcommand{\ecor}{\end{corollary}}
\newcommand{\eprf}{\bbox\vspace{0.1in}}
\newcommand{\beqn}{\begin{equation}}
\newcommand{\eeqn}{\end{equation}}
\newcommand{\bbox}{\vrule height7pt width4pt depth1pt}
\newcommand{\qed}{\eprf}
\newcommand{\clm}{\begin{claim}}
\newcommand{\eclm}{\end{claim}}
\newcommand{\sat}{\models}

\newcommand{\stur}{\vdash}
\newcommand{\rimp}{\Rightarrow}
\newcommand{\dimp}{\Leftrightarrow}
\newcommand{\bor}{\bigvee}
\newcommand{\band}{\bigwedge}
\newcommand{\union}{\cup}
\newcommand{\inter}{\cap}
\newcommand{\xx}{{\bf x}}
\newcommand{\yy}{{\bf y}}
\newcommand{\uu}{{\bf u}}
\newcommand{\vv}{{\bf v}}
\newcommand{\FF}{{\bf F}}

\newcommand{\IN}{\mbox{$I\!\!N$}}

\renewcommand{\phi}{\varphi}
\newcommand{\A}{{\cal A}}

\newcommand{\C}{{\cal C}}

\newcommand{\F}{{\cal F}}

\newcommand{\G}{{\cal G}}
\newcommand{\I}{{\cal I}}

\newcommand{\M}{{\cal M}}

\renewcommand{\P}{{\cal P}}

\newcommand{\R}{{\cal R}}

\newcommand{\U}{{\cal U}}
\newcommand{\V}{{\cal V}}

\newcommand{\Y}{{\cal Y}}

\newcommand{\WW}{{\bf W}}
\newcommand{\XX}{{\bf X}}
\newcommand{\YY}{{\bf Y}}
\newcommand{\ZZ}{{\bf Z}}
\newcommand{\zz}{{\bf z}}
\newcommand{\ww}{{\bf w}}

\newcommand{\<}{\langle}
\renewcommand{\>}{\rangle}

\newcommand{\ol}{\setlength{\itemsep}{0pt}\begin{enumerate}}
\newcommand{\eol}{\end{enumerate}\setlength{\itemsep}{-\parsep}}
\newcommand{\ul}{\setlength{\itemsep}{0pt}\begin{itemize}}
\newcommand{\dl}{\setlength{\itemsep}{0pt}\begin{description}}
\newcommand{\edl}{\end{description}\setlength{\itemsep}{-\parsep}}
\newcommand{\eul}{\end{itemize}\setlength{\itemsep}{-\parsep}}

\newcommand{\true}{{\it true}}

\newcommand{\commentout}[1]{}

\newcommand{\bi}{\begin{itemize}}
\newcommand{\ei}{\end{itemize}}
\newcommand{\be}{\begin{enumerate}}
\newcommand{\ee}{\end{enumerate}}

\newcommand{\shows}{\vdash}

\renewcommand{\L}{{\cal L}}
\renewcommand{\S}{{\cal S}}

\newcommand{\ttt}{\mathbf{t}}
\newcommand{\TT}{\mathbf{T}}

\renewcommand{\YY}{\vec{Y}}
\renewcommand{\yy}{\vec{y}}
\renewcommand{\XX}{\vec{X}}
\renewcommand{\xx}{\vec{x}}
\renewcommand{\ZZ}{\vec{Z}}
\renewcommand{\zz}{\vec{z}}
\renewcommand{\WW}{\vec{W}}
\renewcommand{\ww}{\vec{w}}
\renewcommand{\TT}{\vec{T}}
\renewcommand{\ttt}{\vec{t}}
\newcommand{\affects}{\twoheadrightarrow}

\newcommand{\propertyge}{{\ge}\, 1}
\newcommand{\propertyle}{{\le}\, 1}

\newcommand{\LL}{\L_{\WW, \R', \I'}(\S)}

\usepackage{thmtools, thm-restate}
\title{
	Reasoning About Causal Models With
 	Infinitely Many Variables
}
\author{
    Joseph Y. Halpern,\textsuperscript{\rm 1}
        Spencer Peters\textsuperscript{\rm 1}
}
\affiliations {
    \textsuperscript{\rm 1} Cornell University\\
    halpern@cs.cornell.edu, sp2473@cornell.edu
}

\begin{document}

\maketitle

%spencer11: remove this
%% \setlength{\parskip}{5pt}

%joe3*: I rewrote the abstract and intro.  I left in the old version,
%commented out, at the end of the section.  Since I made so many
%change, I didn't both
%marking most of my individual changes.  My goal was to put in more of
%a story, %less of a listing %of theorems.  I've put in comments on
%more significant issues
\begin{abstract}
	\emph{Generalized structural equations models (GSEMs)} \cite{PH20},
	are, as the name suggests, a generalization of structural equations
	models (SEMs).  They can deal with (among other things) infinitely
	many variables with infinite ranges, which is critical for capturing
	dynamical systems.  We provide a sound and complete axiomatization of
	causal reasoning in GSEMs that is an extension of the sound and
	complete axiomatization provided by Halpern \nciteyear{Hal20} for SEMs.
        %spencer25*: Added. Which one do you like better?
%joe23: I slightly prefer your version, with one further change
%    Looking at GSEMs helps clarify what properties each of Halpern's
%joe25: I would cut ``each of'', since we don't end up looking at all
%the axioms
  %spencer27: Sounds good.
        %Considering GSEMs helps clarify what properties each of Halpern's axioms captures.
Considering GSEMs helps clarify what properties Halpern's axioms capture.
  % Alternatively:
  % Looking at GSEMs helps clarify what each of Halpern's axioms is doing.
	%joe3*: I didn't really understand this sentence.  I'm not sure what
	%you're trying to say here.  Let's discuss
	%spencer3*: I was trying to bring in the roll-forward property.
	% Not sure how best to do this.
	%joe4: I think for now, w\e  should just cut it.  If we  can prove a
	%completeness result, we can try to come up with a useful sentence.
	%We also capture interesting features of
	%the example continuous-time GSEMs given in \cite{PH20} in a general
	%definition, and show how this definition can be used to prove other
	%properties shared by these GSEMs.
\end{abstract}

%spencer9*: It might be worth telling a more specific story about
% using our less structured models (GSEMs) to understand better what
% each of the SEM axioms is saying.
% What are the bare minimum guarantees any reasonable causal model offers? (AX_{basic})
% What does having an explicit mechanism
% (the structural equations) guarantee us over (AX_{basic})? (You gave the answer in 2000, AX/AX_{rec})
% Are these guarantees independent? (e.g., can one be imposed without the others holding?) (Yes)
% What properties do reasonable continuous-time models actually satisfy? (Roll-forward?)
% How do these properties relate to the guarantees provided by (acyclic?) structural equations? (Not sure yet, but we want to answer this)
% What do you think?

\section{Introduction}
Systems that evolve in continuous time are ubiquitous in all areas of
science and engineering.
%joe1
%Modeling causality in these systems can yield
%new insight into pathways in molecular biology
A number of approaches have been used to model causality in such systems,
%joe26
%ranging from dynamical systems involving differential equations,
ranging from dynamical systems involving differential equations
%spencer9:
%\emph{rule-based models}
to \emph{rule-based models}
\cite{laurent_counterfactual_2018} for
%joe26
%capturing complex interactions in molecular biology, and \emph{hybrid
capturing complex interactions in molecular biology and \emph{hybrid
  automata} \cite{alur_hybrid_1992}
%spencer28: flows better without this
% , a popular approach
for describing
mixed discrete-continuous systems.
%joe3: We can add this in too.  Does Friston have a name for his
%systems?  If we add it, I need the reference
%basic neuroscience \cite{friston2003dynamic}
%In molecular biology, \emph{rule-based models}
%\cite{laurent_counterfactual_2018} have been used to model causality
%and have yielded new insight into pathways of
%%new insight into pathways in molecular biology
%\cite{laurent_counterfactual_2018}, basic neuroscience
%joe3: I''ll need this reference (although I can probably figure it out
%\cite{friston2003dynamic}, and the engineering of hybrid
%discrete-continuous systems.
%joe3
%spencer3: Friston's approach
% https://www.ncbi.nlm.nih.gov/pubmed/12948688
% is actually more like Bayesian inference on differential equation parameters,
% so on second thought, I don't think we should include it.
The standard approach to modeling causality, \emph{structural-equations
	models} (SEMs), introduced by Pearl \nciteyear{pearl:2k}, cannot
handle such systems, since it allows only finitely many variables,
which each have finite ranges.  But continuous systems typically have
real-valued variables indexed by time, which ranges over the reals
(e.g., the temperature at time $t$).

%joe3: since it's not blind, you can actually say ``we''
%joe17: if it's blind, you'll have to change this back
%spencer18: again, this is moot
%Peters and Halpern recently proposed a
%joe24
%We recently proposed an extension of SEMs, \emph{generalized
%spencer26*: Question: why can we not say 'We recently proposed' given
%that the other submission is also anonymized?
%joe25: It's not a big deal, but it does show that both papers have
%the same set of authors.  I was trying to be completely anonymous
%spencer27*: Got it. But then is it problematic that we put
%[Submission 576] in the CMT abstract and in our bib entry? It seems
%like we need that so that they can find the GSEM paper. But that
%gives away that we know it's also being submitted to IJCAI, and how
%would we know that, unless we're the authors?
%joe26: It could be that there is an overlap in authors, without it
%being the identical set.  That said, I'm not sure why we need the
%Submission number.  On the other hand, we shouldhave the title.  I'll
%revise joe.bib
%spencer28*: Should the citation in the abstract in CMT have the
%submission number? That is, [Submission 576]? Or should it be exactly
%the same as the abstract in the paper, that is, the citation should
%look like [Anonymous, 2020]?
%joe27: I prefer the latter
%spencer29*: Sorry to keep checking on this, but if we do that, are you sure the reviewers for this paper will be able to find the other paper?
\fullv{We recently proposed an} \shortv{An} extension of SEMs,
%spencer26: added comma
% \emph{generalized structural-equations models (GSEMs)}, that can capture such systems
%joe25
%\emph{generalized structural-equations models (GSEMs)}, that can
\emph{generalized structural-equations models (GSEMs)}, which can
%spencer44:
%capture such systems,
capture such systems
%joe24
%\cite{PH20}.
\fullv{\cite{PH20}.}
%spencer26:
% \shortv{was recently proposition \cite{PH20}.}
%joe29
%\shortv{was recently proposed \cite{PH20}.}
\shortv{was recently proposed \cite{PH20}.}%
The goal of this paper is to provide a sound and complete
axiomatization of GSEMs, in the spirit of that provided for SEMs by
%joe6
%Halpern~\nciteyear{Hal20}.  However, there are a number of features of
Halpern~\nciteyear{Hal20}.  There are a number of features of
GSEMs that make reasoning about them
subtle.  We briefly discuss some of them here.

Like SEMs, GSEMs are defined with respect to a signature
$\S$ that describes the variables in the model, their possible values, and the
allowed interventions. The language $\L(\S)$ of causal formulas that
we consider (like that of Halpern \nciteyear{Hal20}) is
%joe3
%the axiom systems $AX^+(\S)$ and $AX^+_{rec}(\S)$ given in
%\cite{Hal20} for general SEMs and acyclic SEMs respectively, are all
parameterized by $\S$.
%As we shall see, The choice of $\S$ has a significant impact on the
%axiom system.
%spencer6*: I feel pretty strongly about the "equations" thing.
A GSEM is a mapping that,
%spencer42:
%given an intervention, produces a set of assignments
%joe36
%given an intervention (and a context, see Section \ref{sec:review}),
given an intervention (and a \emph{context}, see Section
\ref{sec:review} for details),
produces a set of assignments
%joe26: following your suggestion
%spencer28: We should do this in GSEM too. Adding a to-do there.
%to the variables given in $\S. These assignments are called
%\emph{solutions}, and
%$
to the variables given in $\S$, called \emph{outcomes};
intuitively, they correspond to possible outcomes after the intervention is performed.
%A GSEM is characterized by equations that describe the effect of
%interventions.  There may in general be multiple solutions to these
%equations.
If the signature  $\S$ is finite (i.e., there are
finitely many variables, each of which can take on only finitely many
%joe26
%values), there can be only finitely many solutions.
values), there can be only finitely many outcomes.
%spencer6: cut
% to these equations.
This is, in particular, the case with SEMs.
%spencer9: Think this might be too early to bring up recursivity.
%Indeed, in a
%\emph{recursive} or \emph{acyclic} SEM,
%where there are no cyclic dependencies between the variables,
%there is always a unique solution.
%spencer6: cut
%to the equations.
But in general, in a GSEM, there may be infinitely many
%spencer20*: Again, I don't like mentioning equations, because there aren't any!
% solutions to these equations.
%joe26
%solutions.
outcomes.
This complicates reasoning about them,
as we shall see.

%joe3: slowing down.  We need to tell a story
Another complication
involves \emph{allowed interventions}.  In SEMs,
all possible interventions are allowed; that is, we can
intervene by setting any subset of the variables to any of the values
in their ranges.  In GSEMs, we have more expressive power: we can
%joe6
%specify which subset of the possible interventions is allowed.
specify which interventions are allowed.
The idea of limiting the set of interventions has already appeared in
earlier work \cite{BH19,Rub17}.  Intuitively, allowed interventions
are the ones that are feasible or meaningful.   The set of allowed
interventions is part of the signature; it also has an impact on the language.
In the language $\L(\S)$,
we allow a formula of the form $[\XX \gets \xx]\phi$ (which can be
read ``after intervening
by setting the variables in $\XX$ to $\xx$, $\phi$ holds'') only if
$\XX \gets \xx$ is an allowed intervention: if an intervention is not
allowed, we cannot talk about it in the language.
%spencer3:
%As we showed
%joe4
%As we showed in
%spencer4: oops, sorry!1
%joe24
%As we showed
\fullv{As we showed}
%spencer45:
%\shortv{As shown in}
\shortv{As shown by the shell game example in}
\cite{PH20}, restricting to
allowed interventions is useful even when the signature is finite; we
can describe interesting situations that are inconsistent with all
interventions being
%spencer44:
%(see Example~\ref{example:shell-game-SEM}).
%spencer45:
%(see Example~\ref{example:shell-game-SEM}, taken from \cite{PH20}).
\shortv{allowed.}
\fullv{allowed (see Example~\ref{example:shell-game-SEM}, taken from \cite{PH20}).}

%joe6
%The situation gets more complicated with non-finite signatures;
%spencer9: This reads a little vague to me.
%The situation gets more complicated with infinitary signatures;
%joe26
%Besides creating the possibility for infinitely many solutions,
%spencer28: typo
%Besides creating the possibility for infinitely many outcomes.
%joe27
Besides creating the possibility for infinitely many outcomes,
%spencer30*: I can't parse this, I'm guessing it's a typo. Reverting for now.
% Besides creating the possibility for infinitely of outcomes,
the infinitary signatures required for continuous-time systems pose
certain technical problems.
%joe6: not any more
%For one thing, given a variable with
%an infinite range,
%%spencer3: I don't quite understand this sentence.
%we want to be able to say that there is some setting
%of the
%%spencer3*: should this be "variable"?
%%joe4: we may want to talk about more than one variable at a time
%%(although I don't feel strongly about this)
%%spencer4: You're right. It just sounds weird "given a
%%variable... variables" rather than "...variable"
%variables
%with certain properties.
% Thus, we need to extend the
%(essentially propositional) language $\L(\S)$ considered by Halpern
%\nciteyear{Hal20}  to allow some limited
%existential quantification.  (For variables with finite
%range, this existential quantification can be replaced with finite
%disjunction.)
%joe6*: added next three sentences
%spencer9: reverted
%spencer7:
If,
%if,
for example, we have variables ranging over the reals, and we can
%joe26
%refer to all possible real numbers in the language, then we have an
%uncountable language.  Although we believe that all our results
refer to all possible real numbers in the language, then the
%spencer28: typo
%langauge
language
must be uncountable.  Although we believe that all our results
continue to hold for uncountable languages,
%spencer9:
%it does make
having uncountably many formulas makes
%spencer9:
%things
soundness and completeness arguments
much more complicated.  We
thus restrict the language so that it can refer explicitly to only
countably many values
%spencer40: shaving
% and so that only countably many interventions
% are allowed
and countably many interventions.
This still leaves us with an extremely rich language,
which easily suffices to characterize systems that occur in practice.
%joe6* Is this still true? It should be.
%spencer7*: Yes, but you have to replace "same solutions under all
%allowed interventions" with "same solutions under all intervention in
%the language" (e.g. \I').
% added this below.
%spencer25*: We proved this in PH20 for the full uncountable
%language. We haven't proved a version of it for the countable
%languages yet. It is true for some countable languages; which ones
%depend on M and M'. Specifically, it is true for the languages that
%include the intervention and the finitely many variables and values
%appearing in the formula $\phi$ in the proof of Theorem 2.1 in
%GSEM.tex. Given M and M' with different solutions, any countable
%language can be extended to another countable language that can tell
%the difference between them. But there isn't a single countable
%language that can distinguish all M and M' (because they might differ
%on a variable not in the language). DONE address this.
%joe24*: there's still a sense in which it's true.  I don't think we
%need to go into it here
%spencer26: Ok, sounds good.
%The version of it which is true for countable languages is

%joe24
%We show that the resulting language is rich enough to
It is shown in \cite[Theorem 2.1]{PH20} that  $\L(\S)$ is rich enough to
completely characterize
%spencer6: added
%joe6
%SEMs, and GSEMs where there are only finitely many solutions
SEMs, as well as GSEMs over infinitary signatures where
%joe26
%there are only finitely many solutions to each intervention
there are only finitely many outcomes to each intervention
%spencer3*: I'm not sure we should talk about "equations";
% there really aren't any!
%joe4: but then what are they solutions to?
%spencer4: They are just solutions of the model (to the model?) given a context and intervention--but we did use "equations" in PH20.
%joe5: For what it's worth, I don't think of a model as having
%solutions; I think of equations as having solutions.
%spencer5: Hmm, but we can define the solutions of the model M(u, I) under intervention I and context u arbitrarily. What has solutions here, if not the model? I guess we can say we have "equations" solutions_{u, I} = M(u, I) for each u and I... but this doesn't seem very satisfying.
%spencer27*: Crazy idea. Rename 'solutions' to 'outcomes'
%globally. Just revisited this, and I agree that models don't have
%solutions; equations do. What do you think? So for example, GSEMs and
%SEMs both have outcomes; in SEMs these outcomes are specified as
%solutions to the structural equations; in GSEMs they are directly
%specified.
%joe26: I could go for that :-)  I did it.
%spencer6*: cut
%to the equations
(so, in particular, GSEMs with finite
%spencer24*: changed \M to M everywhere
signatures);
%joe24
%specifically, we show that if $\M$ and $\M'$
%spencer27: as suggested
%specifically, it is shown that if $\M$ and $\M'$
specifically, it is shown that if each of $\M$ and $\M'$
%joe25*: I don't think that Theorem 2.1 in our GSEM paper talks about
%GSEMs; it only talks about SEMs (although it should aply without
%change to GSEMs). The statement of the theorem just says ``causal
%models''.  We should clearify it.  Also, it should be possible for M
%to be a SEM and G' to be a GSEM, so perhaps we can say ``If each of  M and M'
%is either a SEM or a GSEM with only finitely many, and M and M' have
%the same signature, then ...''
%spencer27: I agree. Clarified this in the other paper
%spencer27: as suggested
% are two SEMs with the same signature, or two GSEMs
% with the same signature for which there are only finitely many solutions,
%joe26
%is either a SEM or a GSEM for which there are only finitely many solutions
is either a SEM or a GSEM for which there are only finitely many outcomes
%spencer27:
%to each intervention
to each intervention,
%joe3*: I think this will prove to be useful notation
%spencer3: I like this
then $\M$ and $\M'$ are \emph{$\L(\S)$-equivalent}, that is, they agree on
all formulas in $\L(\S)$,
%spencer6:
iff they are \emph{equivalent}, that is, iff
%joe26
%they have the same solutions
they have the same outcomes
%spencer6*: cut; added mention of allowed interventions
%to the equations
%joe25: we should cut the reference Theorem 4.2 here.
%spencer27: Done.
%under all allowed interventions(Theorem \ref{theorem:formulas-equiv-solutions}).
under all allowed interventions.
(We remark that
this is no longer the case if we consider GSEMs for which
%joe26
%there may be infinitely many solutions
there may be infinitely many outcomes
%spencer27: more 'equations'
%to the equations
%joe26
%for a given context and intervention (which can
%certainly be the case in some dynamical systems);
%see Example~\ref{example:infinite-solutions-not-characterized}.
%}
for a given context and intervention,
%spencer40: shaving
% something that can
which can
%joe26
%spencer40: shaving
%certainly be the case in some dynamical systems;
certainly be the case in dynamical systems;
%joe26
see Example~\ref{example:infinite-solutions-not-characterized}.)
%joe6
%\todo{}We also show that every SEM is $\L(\S)$-equivalent to a GSEM with the
%same signature.
%spencer7: not sure what this todo was for
%\todo{}
%spencer9: added paragraph break

%spencer9: this result was also from PH20
%We also show that
%spencer25: moving this paragraph below

%joe25: Why not just cut the next sentence?  It's not clear what
%connction this paragraph has with the previous paragraph
%spencer27: Sounds good.
% This, in turn, has implications for axiomatizing GSEMs.
Halpern \nciteyear{Hal20} provided axiom systems $AX^+(\S)$ and
$AX^+_{rec}(\S)$ that he showed were sound and complete for
%joe6: we should be consistent about whether we use ``acyclic'' or ``recursive''
%spencer23*: switching 'recursive' to 'acyclic' everywhere for
%consistency with the other paper.
%general SEMs and acyclic SEMs, respectively.  We extend these systems
general SEMs and acyclic SEMs, respectively.
In this paper,
we extend
%spencer27:
% these systems
$AX^+(\S)$
to arbitrary GSEMs, and several interesting subclasses of GSEMs
%spencer3: I actually think that axiomatizing recursive GSEMs in
%\L(\S) is impossible, because I think the "variables that have no
%dependencies have unique solutions" proviso from PH20 can't be
%expressed in \L(\S).
%joe4: this doesn't follow (I''m not saying it's false; I'm just
%saying that it doesn't follow from your argument).  It could be that
%we have a sound and complete axiomatization of recursive GSEMs even
%though you can't experss this proviso.  For example, there is a sound
%and complete axiomatization of the real numbers in the first-order
%language of arithemtic (0,1, +, x), even though you can't express the
%proviso that there are no infinitesimals.
%spencer23*: This is no longer correct
% (such as acyclic GSEMs).
%joe26
%(such as GSEMs with unique solutions).
(such as GSEMs with unique outcomes).
%spencer7*: I would naturally interpret this as saying that AX^+
%(Halpern) is equivalent to AX_{basic}^+ (our axiomatization for GSEMs
%with no conditions imposed) which is not what we mean to say. What do
%you think of this?
%where $\S$ is a finite signature and all interventions are allowed,
%joe7: I think it's clear enough for the intro; we make it completely
%precise later
%spencer8: Makes sense.
%spencer9: I still think this is confusing. I want to make the punchy statement from PH20 that imposing SEM axioms on the class of GSEMs recovers the class of SEMs.
%spencer25: Moved from above
%joe24: removed paragraph break, since we haven't yet talked about the
%implications for axiomatizaing GSEMs.
%
%joe24
%As shown by Peters and Halpern \nciteyear{PH20}, every
%spencer27: added
First, we show that $AX^+(\S)$ is sound and complete for the class of GSEMs
% (that is, GSEMs over a finite signature $\S$)
satisfying $AX^+(\S)$, if $\S$ is finite and $\S$ is \emph{universal}; that is, if all interventions are allowed. This is an easy corollary of \cite[Theorem 3.4]{PH20}, which states that if $\S$ is finite and universal, then every SEM with signature $\S$ is equivalent to a GSEM satisfying $AX^+(\S)$, and vice versa.
The assumption that $\S$ is universal is critical here.
%spencer44:
%Example 3.6 in \cite{PH20} gives
Example \ref{example:infinite-solutions-not-characterized} (from
\cite{PH20}) gives
a GSEM over a finite signature $\S$ that satisfies all the axioms of
$AX^+(\S)$ but is not equivalent to any SEM.  This
implies that $AX^+(\S)$ is no longer complete for SEMs when $\S$ is
not universal (Theorem \ref{theorem:axioms-not-complete-for-SEMs}).
We then show that a subsystem of $AX^+(\S)$ that we call
$AX^+_{basic}(\S)$ is sound and complete for arbitrary GSEMs over a
finite signature $\S$.
%spencer39*: added
%joe34
%We also show that as in SEMs, extending $AX^+(\S)$ with one more axiom
We also show that, as in SEMs, extending $AX^+(\S)$ with one more axiom
gives a sound and complete system for
%spencer44*: No longer a conjecture!
% finite \emph{acyclic} GSEMs; we
% conjecture a similar result holds for general acyclic GSEMs.
\emph{acyclic} GSEMs.
%
%spencer39:
%Extending this result to arbitrary (possibly
%spencer44: I think this is clearer
%Extending these results to arbitrary
Proving these results for arbitrary
(possibly infinite) signatures $\S$ is nontrivial, because one of the axioms of
$AX^+_{basic}(\S)$ is no longer in the language $\L(\S)$ when $\S$ is
%joe26
%infinite. We show that this axiom, D2, can be replaced with a new
%inference rule, D2$^+$, that is equivalent when $\S$ is finite; and
%spencer44:
%infinite.
and the axiom corresponding to acyclicity must be strengthened.
We show that this axiom can be replaced with a new
inference rule that gives an equivalent system when $\S$ is finite.  Moreover,
the resulting axiom system, $AX^*_{basic}(\S)$, is sound and
complete for arbitrary GSEMs (Theorem \ref{theorem:GSEMs}). We further
%joe26
%show that several properties of SEMs (such as unique solutions to all
%contexts and interventions) can be recovered from arbitrary GSEMs by
show that several properties of SEMs (such as having unique outcomes for all
%joe34
%interventions in all contexts) can be recovered from arbitrary GSEMs by
interventions in all contexts) can be enforced in GSEMs by
adding axioms from $AX^+(\S)$ to $AX^*_{basic}(\S)$ (Theorem
\ref{theorem:GSEMs}).
%spencer27:
%Doing so helps clarify what properties Halpern's axioms capture.
Doing so helps clarify what properties the added axioms capture.

\section{SEMs: a review}\label{sec:review}
Formally, a \emph{structural-equations model} $M$
is a pair $(\S,\F)$, where $\S$ is a \emph{signature}, which explicitly
lists the endogenous and exogenous variables  and characterizes
their possible values, and $\F$ defines a set of \emph{modifiable
%joe35
%structural equations}, relating the values of the variables.
structural equations}, relating the values of variables.
We extend the signature to include a set of \emph{allowed
	interventions}, as
was done in earlier work  \cite{BH19,Rub17}.
Intuitively, allowed interventions are the ones that are feasible or
meaningful.
%joe6
%A signature $\S$ is a tuple $(\U,\V,\R,\I)$.
A signature $\S$ is a tuple $(\U,\V,\R,\I)$, where
$\U$ is a set of exogenous variables, $\V$ is a set
of endogenous variables, and $\R$ associates with every variable $Y \in
	\U \union \V$ a
nonempty, finite
set $\R(Y)$ of possible values for
$Y$ (i.e., the
set of
values over which $Y$ {\em ranges}). We
assume (as is typical for SEMs) that $\U$ and $\V$ are finite sets, and adopt the convention
that for $\YY \subseteq \U \cup \V$, $\R(\YY)$ denotes the product of
the ranges of the variables appearing in $\YY$; that is, $\R(\YY)
	:= \times_{Y \in \YY} \R(Y)$.
Finally, an intervention $I \in \I$ is a set of pairs $(X, x)$, where
$X \in \V$ and $x \in \R(X)$.
%joe6*: this doesn't seem right (or reasonable)
%For each $X \in \V$,
%there is at most one $x \in \R(X)$ with $(X, x) \in I$.
%spencer7*: why not? Any intervention should set any given variable to at most one value.
We abbreviate an intervention $I$ by $\XX \gets \xx$, where
$\XX \subseteq \V$.
%joe24
%joe26: reinstated; I think we have the room
%\fullv{
%joe35
\fullv{Although this notation makes most sense
%joe35: typo
%if $\XX$ is nonempty, we allow $\XX$ to be empty (
  if $\XX$ is nonempty, we}
\shortv{We} allow $\XX$ to be empty
%spencer40: shaving
%joe35: undid for now
(which amounts to
%that is,
not
intervening at all).
%}
%joe6: cut; this seems like a special case of \XX <- \xx
%If $I$ consists of exactly one pair $(Y, y)$, we abbreviate $I$ as $Y
%\gets y$.

$\F$ associates with each endogenous variable $X \in \V$ a
function denoted $F_X$ such that $F_X: \R(\U \union \V - \{X\})
	\rightarrow \R(X)$.
This mathematical notation just makes precise the fact that
$F_X$ determines the value of $X$,
given the values of all the other variables in $\U \union \V$.
If there is one exogenous variable $U$ and three endogenous
variables, $X$, $Y$, and $Z$, then $F_X$ defines the values of $X$ in
terms of the values of $Y$, $Z$, and $U$.  For example, we might have
$F_X(u,y,z) = u+y$, which is usually written as
$X = U+Y$.   Thus, if $Y = 3$ and $U = 2$, then
$X=5$, regardless of how $Z$ is set.

The structural equations define what happens in the presence of external
interventions.
Setting the value of some variable $X$ to $x$ in a SEM
$M = (\S,\F)$ results in a new SEM, denoted $M_{X
			\gets x}$, which is identical to $M$, except that the
equation for $X$ in $\F$ is replaced by $X = x$. Interventions on
subsets $\XX$ of $\V$ are defined similarly. Notice that $M_{\XX
			\gets \xx}$ is always well defined, even if $(\XX \gets \xx) \notin \I$.
In earlier work, the reason that the model included allowed
interventions was that, for example, relationships between two models
were required to hold only for allowed interventions (i.e., the
interventions that were meaningful).
%spencer40: shaving
% As we shall see, here, the set of allowed
Here, the set of allowed
interventions plays a different role, influencing the language (what
we are allowed to talk about).

%joe26
%Given context $\uu \in \R(\U)$, the \emph{solutions} of a
%spencer42:
%Given context $\uu \in \R(\U)$, the \emph{outcomes} of a
Given a \emph{context} $\uu \in \R(\U)$, the \emph{outcomes} of a
SEM $M$ under intervention $\XX \gets \xx$ are all
assignments of values $\vv \in \R(\V)$ such that the assignments $\uu$
and $\vv$ together satisfy the structural equations of $M_{\XX \gets
%joe26
%  \xx}$. This set of  solutions is denoted
  %$
    \xx}$. This set of  outcomes is denoted
$M(\uu, \XX \gets \xx)$.
%joe26
%Given a solution $\vv$, we denote by $\vv[X]$ and $\vv[\XX]$ the value
%spencer28*: Do we need this paragraph?
%joe27: Only if we use it (and I think we do, at least the \vv[X] notation
%spencer30*: Oops, I meant after this sentence. Starting with 'As
%discussed in the introduction'. What do you think about that one?
%joe28: Since we make a bit of a fuss about acylicity, and have plenty
%of room, and think it's fine to keep it
Given an outcome $\vv$, we denote by $\vv[X]$ and $\vv[\XX]$ the value
that $\vv$ assigns to $X$ and the restriction of $\vv$ to $\R(\XX)$
respectively.
%spencer42:
(That is, $\vv[\XX] \in \R(\XX)$ is the assignment to the variables
$\XX$ that agrees with $\vv$.)
%spencer39: I don't think this was discussed in the introduction!
%As discussed in the introduction, an important special case
An important special case of SEMs
%spencer23: switching this back
%joe6
are acyclic (or recursive) SEMs.   Formally, an acyclic SEM is one
% are recursive (or acyclic) SEMs.   Formally, a recursive SEM is one
for which,
for every context $\uu \in \R(U)$,
there is some total ordering $\prec_\uu$ of the endogenous variables
(the ones in $\V$)
such that if $X \prec_\uu Y$, then $X$ is independent of $Y$,
that is,
$F_X(\uu, \ldots, y, \ldots) = F_X(\uu, \ldots, y', \ldots)$
for all $y, y' \in \R(Y)$.
%joe6
%Intuitively, if a theory is acyclic, there is no
%feedback. Acyclic models always have unique solutions; this is a
%consequence of assuming that $\V$ is finite.

%joe6: added label
\section{Axiomatizing SEMs}\label{sec:SEMaxioms}
In order to talk about SEMs and the information they represent more
precisely, we use the formal language $\L(\S)$ for SEMs
having signature $\S$, introduced by Halpern
%spencer1: the citations are also clobbered by the
% given style files. Not sure what the best way to fix is.
%joe24
%\nciteyear{Hal20}; see also \cite{GallesPearl98}.
\nciteyear{Hal20}.
%spencer1*: I want the language description to be formal, since it is
%a core part of the main topic of this paper. Not sure where it needs
%to be tightened up though.
%An informal description of this language
%follows; for details more, see \cite{Hal20}.
%joe1*: it needs to be tightened up:-)
%Note that \cite{Hal20} did not consider allowed interventions.
%joe10

%We restrict the language given in \cite{Hal20} to formulas containing
We restrict the language used by Halpern \nciteyear{Hal20} to formulas
containing
%spencer17: added
only allowed interventions.
Fix a signature $\S = (\U, \V, \R, \I)$.
%joe6
%An \emph{primitive event} (over signature $\S$) has the form $X=x$,
%joe18*: It may be better to call this a primitive formla (see my comment
%later in response to your comment about conjunctive formulas below),
%although I did call it a primitive event in the book.
%joe21: cut ``signature'' in the next line to save a line.
A \emph{primitive event} (over signature $\S$) has the form $X=x$,
where $X \in \V$ and $x \in \R(\V)$.
%spencer6: added
An \emph{event} is a Boolean combination of primitive events.
%joe34: removed paragraph break
%
% spencer14: I'd like to denote events by e_1, e_2, ... everywhere
% to avoid confusion with formulas.
An \emph{atomic formula} (over
$\S$) has the form $[\YY \gets \yy]\phi$, where $\YY \gets \yy \in
	\I$ (i.e., it is an allowed intervention), and $\phi$
%spencer6: slight refactor
%  is a Boolean
%  combination of primitive events.
is an event.
A \emph{causal formula} (over $\S$) is a Boolean combination of atomic formulas.
The language $\L(\S)$ consists
%spencer6: slight refactor
%of all Boolean combinations of atomic formulas.
of all causal formulas over $\S$.
There are a number of minor differences between the language
considered here and that considered by Halpern \nciteyear{Hal20}.
First, since Halpern implicitly assumed that all interventions were
allowed, he did not have the restriction to allowed interventions.
Second, Halpern considered a slightly
richer language, where the context $\uu$ was part of the formula,
%spencer42:
%not on the left-hand side of the $\models$.
not on the left-hand side of the $\models$ (see below).
Specifically, a
%joe6
%  primitive event had the  $X(\uu) = x$.  We explain the reasons that
%joe24
%primitive event had the form $X(\uu) = x$.  We explain the reasons that
%Halpern did this below, and its impact.  For now we just note that
primitive event had the form $X(\uu) = x$.  \fullv{We explain the reasons that
Halpern did this below, and its impact.  For now we just note that it}
\shortv{It} has become standard not to include the context $\uu$ in the
%joe6
%  formula (see, e.g., \cite{HP01a,Hal48}.
formula (see, e.g., \cite{HP01a,Hal48}).

%joe1*: You should first define what it means for (M,u,v) to satisfy a
%primitive event and then a Boolean combination of primitive events.
%Finally, you can give the semantics of arbitrary formulas.  You
%should then explain how X(u) = x is defined in my earlier paper.
%spencer6: This sounds good, but I'm pretty sure I only want (M, u)
% We only wanted (M, u, v) when we were talking about actual causes
% and v referred to "what actually happened".
%  We restrict the language used by Halpern to formulas containing
%only allowed interventions.
%spencer6*: added
Next we define the semantics of $\L(\S)$.
%joe6: don't need this; we're already talking about \L(\S)
%    Fix a signature $\S$.  Given
%a primitive event $X = x$, we say that an assignment
%$\vv \in \R(\V)$ satisfies $X = x$ if $\vv[X] = x$. We extend this
%definition to events by structural induction in the obvious way, that
An assignment     $\vv \in \R(\V)$ \emph{satisfies} the primitive
%joe7: you use this notation later
%spencer8: thank you!
%    event $X = x$ if $\vv[X] = x$. We extend this
event $X = x$, written
%spencer13:
% $\vv \sat \phi$,
%joe12
$\vv \sat (X = x)$
if $\vv[X] = x$. We extend this
definition to Boolean combinations of primitive events by structural
induction in the obvious way, that
%joe7
%is, say that $\vv$ satisfies $e_1 \wedge e_2$ iff $\vv$ satisfies
%$e_1$ and $\vv$ satisfies $e_2$, and similarly for the other Boolean
is, say that $\vv \models e_1 \wedge e_2$ iff $\vv \models e_1$ and
$\vv \models e_2$, and similarly for the other Boolean
connectives $\vee$ and $\neg$.
Fix a SEM $M$ with signature $\S$. Given a context $\uu \in \R(U)$,
we say that $M$ satisfies the atomic formula $[\YY \gets \yy]\varphi$
in context $\uu$, written $(M, \uu) \models [\YY \gets \yy]\varphi$,
%joe26
%if all solutions $\vv \in M(\uu, \YY \gets \yy)$ satisfy
if all outcomes $\vv \in M(\uu, \YY \gets \yy)$ satisfy
$\varphi$. Finally, we extend this definition to causal formulas by
%joe26
%structural induction as above. That is, say that $M, \uu \models [\YY
%$
structural induction as above. That is, $M, \uu \models [\YY
%joe6
		%  \gets \yy]\varphi \wedge [\ZZ \gets \zz]\psi$ iff $M, \uu
		%\models[\YY \gets \yy]\varphi$ and $M, \uu \models [\ZZ \gets
%$
		\gets \yy]\varphi \wedge [\ZZ \gets \zz]\psi$ iff $(M, \uu)
	\models[\YY \gets \yy]\varphi$ and $(M, \uu) \models [\ZZ \gets
		\zz]\psi$, and similarly for $\vee$ and $\neg$.
%joe26
%        As usual, we define
        %$\< \YY \gets \yy \> \varphi$ as $\neg [\YY \gets
                As usual,
$\< \YY \gets \yy \> \varphi$ is taken to be an abbreviation for $\neg [\YY \gets
		\yy] (\neg \phi)$.  It is easy to check that $(M,\uu) \sat \< \YY
	\gets \yy \> \varphi$ iff $\varphi$ is true of at
%joe26
%        least one solution $\vv \in M(\uu, \YY \gets \yy)$.
        least one outcome $\vv \in M(\uu, \YY \gets \yy)$.
        % spencer14: added
%joe26: cut; no need (this is quite standard
%        Notice that we could just as well have started with the formulas
%$\<\YY \gets \yy\> \varphi$ and defined $[\YY \gets \yy] \varphi =
%	\neg \< \YY \gets \yy \> (\neg \varphi)$. For this reason, the
%formulas $\<\YY \gets \yy\> \varphi$ are called \emph{dual} basic
%causal formulas.
%joe6: this is nonstandard terminology; put in the standard version
%We say that
%$M$ satisfies the causal formula $\phi$, written $M \models \phi$, if
%spencer28: changed \phi to \psi here to avoid confusion
% The causal formula $\phi$ is \emph{valid in $M$}, written $M \models \phi$, if
% $M, \uu \models \phi$ for all $\uu \in \R(\U)$; $\phi$ is
% \emph{satisfied in $M$} if $(M,\uu) \sat \phi$ for some context $\uu
% 	\in \R(\U)$.
 The causal formula $\psi$ is \emph{valid in $M$}, written $M \models \psi$, if
$M, \uu \models \psi$ for all $\uu \in \R(\U)$; $\psi$ is
\emph{satisfied in $M$} if $(M,\uu) \sat \psi$ for some context $\uu
	\in \R(\U)$.
  \fullv{
%spencer28:
%The language $\L(\S)$ completely characterizes the solutions of a SEM
The language $\L(\S)$ completely characterizes the outcomes of a SEM
%joe26
%or GSEM with finitely many solutions (Theorem
or GSEM with finitely many outcomes (Theorem
\ref{theorem:formulas-equiv-solutions}). However, this is not true in
%joe26
%general for GSEMs with infinitely many solutions;
general for GSEMs with infinitely many outcomes;
%joe6*: this example is misplaced, since we haven't given the
%semantics of GSEMs yet
%as we show in the following example.
%spencer7: good catch!
%joe24
%we show this in Section~\ref{sec:GSEMreview}, after giving the
%semantics of GSEMs.
%spencer36:
%see Example~\ref{example:shell-game-SEM}.
see Example~\ref{example:shell-game-SEM} (taken from \cite{PH20}).
}

%joe6*: slowing down, defining \leadsto.
We now review Halpern's axiomatization of SEMs where all interventions
%joe24
%are allowed
are allowed (which is based on that of Galles and Pearl
\nciteyear{GallesPearl98}).
%joe24
%
To axiomatize acyclic SEMs, following Halpern, we define $Y
\leadsto Z$, read ``$Y$ affects $Z$'', as an abbreviation for the
formula
% spencer12: DONE fix this overfull hbox!
% spencer15: done.
% $$\begin{array}{ll}\bor_{\XX \subseteq \V, \xx \in \R(\XX), y \in
%   \R(y), z \ne z' \in \R(Z)}
% &([\XX \gets \xx](Z = z) \land\\
% &[\XX \gets \xx, Y \gets y](Z = z')).\end{array}$$  Thus, $(M,\uu)
% \models Y\leadsto Z$
\begin{equation} \nonumber
	\begin{split}
%joe40*: is there some way to get the equation on one line, by
%splitting just the subscript of the \lor?
	  &\lor_{\XX \subseteq \V, \xx \in \R(\XX), y \in \R(y), z \ne z' \in \R(Z)} \\
		%joe21: the period at the end of the line shold be a semicolon or colon
		%spencer24: changed to semicolon
		&\quad ([\XX \gets \xx](Z = z) \land [\XX \gets \xx, Y \gets y](Z = z'));
	\end{split}
\end{equation}
%spencer24: added
that is, $Y$ affects $Z$
if there is some setting of some endogenous variables $\XX$
for which changing the value of $Y$ changes the
value of $Z$.  This definition is used in axiom D6 below, which
characterizes
%spencer7:
%acyclicity.
%joe22: do we want to say ``acyclicity'', for consistency?
%spencer25: Yes!
% recursiveness.
acyclicity.

Consider the following axioms:

%joe1
%\begin{definition}
%$AX^+$ consists of axiom schema D0-D5 and D7-D9, and inference rule MP.
%$AX^+_{rec}$ results from adding D6 to $AX^+$.
%\end{definition}
%spencer26: formatting so D10(a) and D10(b) dont stick out into the
%column margin.
%joe25: this helps :-)
%\begin{itemize}
\begin{itemize}[leftmargin=\parindent + 5pt, align=left, labelwidth=\parindent, labelsep=5pt, itemsep=2pt]
	\item[D0.] All instances of propositional tautologies.
	\item[D1.] $[\YY \gets \yy](X = x \rimp
		      X \ne x')$  if $x, x' \in \R(X)$, $x \ne x'$ \hfill
	      (functionality)
	      %spencer1: reverted to old axiom for this section.
	\item[D2.] $[\YY \gets \yy](\bor_{x \in \R(X)} X = x)$
	      %joe6
	      %  \hfill (definitenes)
	      \hfill (definiteness)
	\item[D3.] $\<\XX \gets \xx\>(W = w
		      %joe6: replaced \YY=\yy by \phi
		      %  \land \YY = \yy)
		      \land \phi)
		      \rimp \<\XX \gets \xx;W \gets
		      %joe6
		      %  w\>(\YY = \yy)
		      w\>(\phi)$
	      \hfill
	      (composition)
	      %joe6* seems more natural
	      %\item[D4.] $[\boldW \gets \boldw; X \gets x](X = x)$ \hfill
	\item[D4.] $[\XX \gets \xx](\XX = \xx)$ \hfill
	      (effectiveness)
	\item[D5.] $(\<\XX \gets \xx; Y \gets y\> (W = w \land
		      \ZZ = \zz)  \land
		      \<\XX \gets \xx; W \gets w\> (Y = y \land
		      \ZZ = \zz))$\\
	      $\mbox{ }\ \ \ \rimp \<\XX \gets \xx\> (W
		      = w \land Y = y \land \ZZ = \zz)
		      %joe6
		      %, where $\ZZ = \V - (\XX \union \{W,Y\})$\\
	      $, if $\ZZ = \V - (\XX \union \{W,Y\})$
	      \mbox{ } \hfill (reversibility)
	      %spencer2:
	\item[D6.]
	      %(X_0 \leadsto X_1 \land \ldots \land X_{k-1} \leadsto X_k)
	      %\rimp
	      %\neg (X_k \leadsto X_0)$
	      %$\neg \Bigg(
	      %joe6*: more transparent and *much* easier to parse
	      %\band_{i = 0}^{k - 1} \Big( [\YY_i \gets \yy_i](X_{i + 1} = z_i)
	      %\wedge [\YY_i \gets \yy_i; X_i \gets x_i](X_{i+1} = z_i') \Big)
	      %\wedge [\YY_k \gets \yy_k](X_0 = z_k) \wedge [\YY_k \gets \yy_k; X_k
	      %\gets x_k](X_0 = z_k') \Bigg)
	      $(X_0 \leadsto X_1 \land \ldots \land X_{k-1} \leadsto X_k)
		      \rimp
		      \neg (X_k \leadsto X_0)$ \hfill (recursiveness)
	      %\neg(\land_{i=0}^{k-1}(X_{i+1})_{\yy_ix_i} = z_i \land
	      %(X_{i+1})_{\yy_i} = z_i') \land
	      %(X_{0})_{\yy_kx_k} = z_k \land
	      %(X_{0})_{\yy_k} = z_k'),
	      %where $z_i \neq z_i'$ for $i = 0, 1,
	      %\ldots, k$.
	      %\hfill
	      %(recursiveness)
	\item[D7.] $([\XX \gets \xx]\phi \land [\XX \gets
			      \xx](\phi \rimp \psi)) \rimp  [\XX \gets \xx]\psi$
	      \hfill
	      (distribution)
	\item[D8.] $[\XX \gets \xx]\phi$ if $\phi$ is a propositional
	      tautology  \hfill (generalization)
	\item[D9.]
	      %joe6*: simplified and added Y = \V -\{X\}
	      %$\<\YY \gets \yy\>true \wedge \big( \<\YY \gets \yy\>(X = x)
	      %  \Rightarrow \<\YY \gets \yy\>(X \neq x') \big)$,
	      $\<\YY \gets \yy\>true \wedge (\<\YY \gets \yy\>\phi
		      \Rightarrow [\YY\gets \yy]\phi)$
	      %if $x \neq x'$
	      %spencer7*: I think you meant to remove this
	      %\ if $x \neq x'$ and
	      \ if
	      $\YY = \V - \{X\}$
	      %> D9. <Y \gets y>true ^ (<Y \gets y>(X = x) => <Y \gets y>(X != x')) if
	      %> x' != x
	      \hfill (unique
%joe26
              %	      solutions for $\V - \{X\}$)
              	      outcomes for $\V - \{X\}$)
	      %\item[D10.] $\langle \YY \gets \yy \rangle \true \land
	      %\lor_{x
	      %\in \R(X)} [\YY \gets \yy](X = x)$
	      %\\ \mbox{}
	      %\hfill
	      %(unique solutions)
	      %\item[D11.] $\langle \YY \gets \yy \rangle (\phi_1(\vec{u}_1)
	      %\land \ldots \land \phi_k(\vec{u}_k)) \dimp
	      %(\langle \YY \gets \yy \rangle \phi_1(\vec{u}_1)
	      %\land \ldots \land
	      %\langle \YY \gets \yy \rangle \phi_k(\vec{u}_k)$, if
	      %$\phi_i(\vec{u}_i)$ is a Boolean combination of formulas of the form
	      %$X(\vec{u}_i) = x$ and
	      %$\vec{u}_i \ne \vec{u}_j$ for $i \ne j$
	      %\hfill (separability)
	      %joe6*: added
	\item[D10(a).]
%joe26
          %	  $\<\YY \gets \yy\>true$ \hfill (at least one solution)
          	  $\<\YY \gets \yy\>true$ \hfill (at least one outcome)
	\item[D10(b).]  $\<\YY \gets \yy\>\phi \rimp [\YY \gets
%joe26
          %          \yy]\phi$ \hfill (at most one solution)
                    \yy]\phi$ \hfill (at most one outcome)

	      %spencer1: fixed typo
	      %\Rightarrow \<Y \gets y\>(X \neq x') \big)$,
	      %  \Rightarrow \<\YY \gets \yy\>(X \neq x') \big)$,
	      %if $x \neq x'$.
	      %   \Rightarrow [\YY \gets \yy](X = x) \big)$,
	      %> D9. <Y \gets y>true ^ (<Y \gets y>(X = x) => <Y \gets y>(X != x')) if
	      %> x' != x
	      %joe6
	      %\\ \mbox{ } \hfill (unique
	      %solutions for $\V - \{X\}$)
	      %\item[D10.] $\langle \YY \gets \yy \rangle \true \land
	      %\lor_{x
	      %\in \R(X)} [\YY \gets \yy](X = x)$
	      %\\ \mbox{}
	      %\hfill
	      %(unique solutions)
	      %\item[D11.] $\langle \YY \gets \yy \rangle (\phi_1(\vec{u}_1)
	      %\land \ldots \land \phi_k(\vec{u}_k)) \dimp
	      %(\langle \YY \gets \yy \rangle \phi_1(\vec{u}_1)
	      %\land \ldots \land
	      %\langle \YY \gets \yy \rangle \phi_k(\vec{u}_k)$, if
	      %$\phi_i(\vec{u}_i)$ is a Boolean combination of formulas of the form
	      %$X(\vec{u}_i) = x$ and
	      %$\vec{u}_i \ne \vec{u}_j$ for $i \ne j$
	      %\hfill (separability)
	\item[MP.] From $\phi$ and $\phi \rimp \psi$, infer $\psi$
	      \hfill
	      (modus ponens)
\end{itemize}

%joe1: moved here
Let
%spencer39: typo
%$AX^+$ consists
$AX^+$ consist
of axiom schema D0-D5 and D7-D9, and inference rule MP;
%joe6*
%and let $AX^+_{rec}$ be the result of adding D6 to $AX^+$.
%joe26
%and
%joe37*
%let $AX^+_{rec}$ be the result of adding D6 and D10 to $AX^+$.
let $AX^+_{rec}$ be the result of adding D6 and D10 to $AX^+$, and
removing D5.

%joe6*: Spencer, isn't D10 also sound for us in ODEs and for recursive
%systems? Added discussion below
%spencer7*: Yes!
%joe7*: Actually, apparently not.  Please let's del with that later
%spencer8*: Yes, sorry. We'll deal with it. I'm still thinking of it as "sound, but with restrictions on the allowed interventions".
These are not quite the same axioms that Halpern \nciteyear{Hal20}
used,
%spencer7:
%(which were inspired by those of Galles and Pearl \nciteyear{GallesPearl98},
%joe24: now mention GP above.
%(which were inspired by those of Galles and Pearl \nciteyear{GallesPearl98}),
although they are equivalent for SEMs.  In more
detail:
\begin{itemize}
	%joe14
	%\item Instead of an arbitrary formulas $\phi$ in D3,
	\item Instead of an arbitrary formula $\phi$ in D3,
	      Halpern had just formulas of the form
	      %spencer7:
	      %$\YY \gets \yy$.
	      $\YY =\yy$.
	      But since in
%joe26
%	      the case of SEMs, every formula $\phi$ is equivalent to
%spencer28*: is 'propositional formula' clearer than 'event'? I'd be
              %tempted to change this to 'event' since we define that earlier.
%joe27: I don't feel strongly about it, but I think of a formulas as
%being equivalent to a disjunction, not an event.  Technically, an
%event is a set of states, while a formula is a syntactic object.
%spencer30: Makes sense. This is fine here.
 the case of SEMs, every propositional formula $\phi$ is equivalent to
              a disjunction of
	      %joe14
	      %formulas of this form
	      formulas of the form
	      %spencer7:
	      %$\YY \gets \yy$,
	      $\YY =\yy$,
	      and
	      $\<\XX \gets \xx\>(\phi \lor \psi) \rimp \<\XX \gets \xx\>\phi \lor \<\XX
		      %joe14
		      %\gets \xx\>\psi$ is provable from the axioms (see Footnote 1 below), our
		      %$
        \gets \xx\>\psi$ is provable from the axioms
        %spencer45:
	      %(cf. Lemma~\ref{lemma:and-distrib-box}), our
        \fullv{(cf. Lemma~\ref{lemma:and-distrib-box}),}
        \shortv{(see the full paper),}
        our
	      version of
	      D3 is easily seen to be equivalent to the original version for SEMs,
	      %joe14
	      %but is more general in the case of GSEMs.
	      but is stronger in the case of GSEMs.
	      %joe14
	      % \item D5 follows from D2, D3, D6, D7, D8, D10, and MP.  (This was
	    \item D5 follows from D2, D3, D6, D7, D8, D10, and MP, so
%joe37
%              Halpern did not include it in his axiomatization.
it is not needed for $AX^+_{rec}$.
              %spencer15*: Unless I'm missing something, D5 is included in your "Axiomatizing Causal Reasoning" paper--in both the C and D axioms, there is a "reversibility" axiom (C5/D5), which looks like our D5.
    %spencer42*: Did you ever address this question? One of the
    %reviewers asked about D5.
%joe36*: Hmmm ... I'm OK with cutting D5 altogether, and then
%replacing this bullet by one that says that I had the reversibility
%axiom but it followed from the others, so we're not including it (and
%I didn't include it as part of my earlier axiomatization either for
%that reason).  We would then have
%to renumber everything carefully, and cut Proposition C.2 and all
%comments about D5.  I would actually be happy to do that (and we
%have a  bit of a space problem anyway.)
%spencer43*: That doesn't work, though! We need D5 if we don't include
%D6, since D5 only follows from the other axioms in the presence of
%D6. (And in Theorem 5.2, we don't include D6.) I'm pretty sure you
%have D5 in your axiomatization paper (contrary to what we currently
    %say) for that reason!
%joe37: You're right; see my (suggested) changes above.
(This was
	      already essentially observed by Galles and Pearl
	      \nciteyear{GallesPearl98}.) Indeed, as we show
	      %spencer15:
	      % below (Proposition~\ref{??}}),
        %spencer45:
	      %below (Proposition~\ref{theorem:d5}),
        \fullv{below (Proposition~\ref{theorem:d5}),}
        \shortv{in the full paper,}
	      % \todo{}
	      in the presence of these other axioms, D5
	      holds even without the requirement that $\ZZ = \V -\{\XX,\YY\}$.
	      %spencer7: I can't immediately see how to prove this.
	      %joe7*:   Here's the idea: Using D6, there's some ordering on the
	      %variables.  Suppose without loss of generality that Y has no effect
	      %on W.  By D10, we can replace <> by [].  Thus, from <\XX<-\xx,
	      %Y<y>W=w & Z=z), we get [\XX <- \xx](W=w). Now by D2, there must be
	      %some \zz' such that [\XX <- \xx](W=w & \ZZ = \zz').  By D3, it follows
	      %that [\XX <- \xx, W <-x](\ZZ = \zz'). Since [\XX <- \xx, W <-x](Y = y
	      %& \ZZ =\zz), we must have \zz' = \zz.  Similarly, there must be a y'
	      %such that [\XX <- \xx](W=w & \ZZ = \zz & Y=y').  Again, by D3, we
	      %have [\XX <- \xx, W <- w](\ZZ = \zz & Y=y'), so we must have y=y'.
	      %Does that make sense?
	      %spencer8: Yes, this makes perfect sense, thank you!
	      %\zz
      \item Halpern's version of D4 said
                %spencer45:
        %$[\boldW \gets \boldw; X \gets x](X= x)$.
$[\WW \gets \ww; X \gets x](X = x)$.
        Using D0, D7, and D8 (and some standard modal logic reasoning),
	      it is easy to see that the two versions are equivalent.
	\item Halpern had slightly different versions of D9 and D10.
	      Specifically, the second conjunct is Halpern's version of D9 is
	      $\lor_{x
			      \in \R(X)} [\YY \gets \yy](X = x)$.
	      %spencer17:
	      % Our version of D9 is equivalent
	      For finite signatures, our version of D9 is equivalent
	      %spencer14: removed extra .
	      %joe12: undid
	      %spencer15: pointing to the appendix.
	      % to Halpern's in the presence of the other axioms.
	      to Halpern's in the presence of the other axioms, as we prove in
	      %joe14
	      %appendix (Theorem \ref{theorem:d9-equivalence}).
        %spencer39:
	      %appendix (see Theorem \ref{theorem:d9-equivalent}).
%joe36: no more supplementary material; we should refer to the full
%paper on arxiv
%spencer43: Makes sense.
% DONE put in dummy citation for shortv.
% For the fullv, decide where this proof should go. Main paper?
        %spencer44:
        % the supplementary material (see Theorem \ref{theorem:d9-equivalent}).
        \shortv{the full paper \cite{HP21}.}
        %spencer46: wrong reference
        %\fullv{Appendix \ref{appendix:additional-proofs}.}
\fullv{Appendix \ref{appendix:d5-and-d9}.}
	\item Finally, Halpern also had an additional
	      axiom D11; we discuss this below.
\end{itemize}

%There are a number of minor differences between the language
%considered here and that considered by Halpern \nciteyear{Hal20}.
%First, since Halpern implicitly assumed that all interventions were
%allowed, he did not have the restriction to allowed interventions.
%Second, Halpern considered a slightly
%  richer language, where the context $\uu$ was part of the formula,
%  not on the left-hand side of the $\models$.  Specifically, a
%  primitive event had the  $X(\uu) = x$.  We explain the reasons that
%  Halpern did this below, and its impact.  For now we just note that
%  it has become standard not to include the context $\uu$ in the
%  formula (see, e.g., \cite{HP01a,Hal48}.

%spencer6*: Added 3 paragraphs about differences with Hal20
As mentioned before, the language $\L(\S)$ considered here differs
from the language considered by Halpern \nciteyear{Hal20}, which we
%joe26
%denote by $\L_H(\S)$, in two ways. First, Halpern implicitly assumed
denote $\L_H(\S)$, in two ways. First, Halpern implicitly assumed
that all interventions were allowed, so he did not have the
%joe6
%restriction to allowed interventions. That is, all formulas [\YY
restriction to allowed interventions. That is, all formulas of the
form $[\YY
			\gets \yy]\phi$ were included in $\L_H(\S)$, where $\YY \subseteq
	\V$ and $\yy \subseteq \R(\YY)$. Second, the causal formulas in
$\L_H(\S)$ were built from atomic events of the form $X(\uu) = x$ as
opposed to the form $X = x$.
%joe6: added next sentence
Halpern \nciteyear{Hal20} gave semantics to formulas with respect to
models $M$, not with respect to pairs $(M,\uu)$.
%joe34: line shaving
%The semantics of atomic formulas in
%$\L_H(\S)$ is given by $M \models [\YY \gets \yy](X(\uu) = x)$
In Halpern's semantics,
$M \models [\YY \gets \yy](X(\uu) = x)$
%joe26
%if all solutions $\vv \in M(\uu, \YY \gets \yy)$ satisfy $X = x$.
if all outcomes $\vv \in M(\uu, \YY \gets \yy)$ satisfy $X = x$.
%joe6
%At the level of atomic formulas, $\L(\S)$ and $\L_H(\S)$ are
%equivalent--the above judgment is equivalent to the judgment $(M,
%\uu) \models [\YY \gets \yy](X = x)$.
%However, $\L_H(\S)$ is more
%expressive at the level of causal formulas---the judgment $M \models
%[\YY \gets \yy](X(\uu) = x) \vee [\YY \gets \yy](X(\uu') = x)$ has no
%analogue in $\L(\S)$. This makes no difference to the axioms given
%above, which are implicitly universally quantified over contexts
%$\uu$. However, to prove completeness in the richer language, Halpern
%had an additional axiom D11, called separability, as follows.
%This makes no difference to the axioms given
%above, which are implicitly universally quantified over contexts
%$\uu$. However, to prove completeness in the richer language, Halpern
%had an additional axiom D11, called separability, as follows.
It is easy to see that $M \models [\YY \gets \yy](X(\uu) = x)$ iff, in
%joe34
%the semantics we are using for this paper, $(M,\uu) \models [\YY \gets $
%spencer40: Not sure what you intended here--maybe:
%the semantics we are of this paper, $(M,\uu) \models [\YY \gets \yy](X = x)$.
the semantics of this paper, $(M,\uu) \models [\YY \gets \yy](X = x)$.
%joe24
\fullv{However, $\L_H(\S)$ is strictly more
expressive than $\L(S)$.  In particular, a formula in $\L_H(\S)$ can
refer to several contexts simultaneously, and this cannot be done in
$\L(\S)$.  For example, the formula
$[\YY \gets \yy](X(\uu) = x) \vee [\YY \gets \yy](X(\uu') = x)$ has no
analogue in $\L(\S)$.}
To deal with the richer language, Halpern
\nciteyear{Hal20} had an additional axiom:
%spencer28: Replacing $\vec{u}$ with $\uu$ for consistency.

% \begin{enumerate}
% 	\item[D11.] $\langle \YY \gets \yy \rangle (\phi_1(\vec{u}_1)
% 		      \land \ldots \land \phi_k(\vec{u}_k)) \dimp
% 		      (\langle \YY \gets \yy \rangle \phi_1(\vec{u}_1)
% 		      \land \ldots \land
% 		      \langle \YY \gets \yy \rangle \phi_k(\vec{u}_k)$, if
% 	      $\phi_i(\vec{u}_i)$ is a Boolean combination of formulas of the form
% 	      $X(\vec{u}_i) = x$ and
% 	      $\vec{u}_i \ne \vec{u}_j$ for $i \ne j$.
% \end{enumerate}
%joe33:
%\begin{enumerate}
\begin{itemize}[leftmargin=\parindent + 5pt, align=left,
    labelwidth=\parindent, labelsep=5pt, itemsep=2pt]
	\item[D11.] $\langle \YY \gets \yy \rangle (\phi_1(\uu_1)
		      \land \ldots \land \phi_k(\uu_k)) \dimp
		      (\langle \YY \gets \yy \rangle \phi_1(\uu_1)
		      \land \ldots \land
		      \langle \YY \gets \yy \rangle \phi_k(\uu_k)$, if
	      $\phi_i(\uu_i)$ is a Boolean combination of formulas of the form
	      $X(\uu_i) = x$ and
	      $\uu_i \ne \uu_j$ for $i \ne j$.
        %joe33:
%\end{enumerate}
\end{itemize}
%However, in the completeness proofs in \cite{Hal20}, D11 was only used
%to prove statements of the form of the LHS of D11, which are not
%expressible in the language $\L(\S)$, so we omit D11 from the versions
%of $AX^+$ and $AX^+_{rec}$ presented here. With this modification, the
%soundness and completeness results of \cite{Hal20} hold for SEMs over
%$\L(\S)$. Namely, $AX^+$ is a sound and complete axiomatization for
%SEMs over $\L(\S)$, and $AX^+_{rec}$ is a sound and complete
%axiomatization for recursive GSEMs over $\L(\S)$. The proofs are
%exactly the same as in \cite{Hal20}, minus the unnecessary reasoning
%using D11.
%joe24
%This additional axiom is used in Halpern's completeness proof only to allow
D11 is used in Halpern's completeness proof only to
reduce consideration from formulas that mention multiple contexts
to formulas that mention only one
%joe24
%context.
\fullv{context.} \shortv{context, which are easily seen to be
  equivalent to formulas in $\L(\S)$.}
%joe24
\fullv{
As we observed above,
formulas in $\L_H(\S)$ that mention only one context are equivalent to
formulas in $\L(\S)$.   Indeed, we}
\shortv{We} can show that the axioms without
%joe34
%D11 are sound and complete for the language $\L(\S)$ using exactly the
D11 are sound and complete for \fullv{the language} $\L(\S)$ using exactly the
same proof as used by Halpern to show that the axioms with D11 are
sound and complete for $\L_H(\S)$, just skipping the step that uses
D11 to reduce to formulas involving just one context.  This is
%joe24: line shaving
%formalized in the following theorem, where we take a signature $\S =
%	(\U,\V,\R,\I)$ to be \emph{universal} if $\I = \I_{univ}$, the set of
formalized in the following theorem, where a signature $\S =
	(\U,\V,\R,\I)$ is \emph{universal} if $\I = \I_{univ}$, the set of
all interventions.

%joe6: stating formal theorem
\begin{theorem}\label{thm:completeness-for-SEMs}
%joe24
  %  \cite{Hal20}
   {\rm  \cite{Hal20}}
%joe24: We don't want a paragraph break here; it looks strange
   If $\S$ is a universal signature, then
   %spencer26:
   % $AX^+$ (resp., $AX^+_{rec}$) is a sound and compete axiomatization for
     $AX^+$ (resp., $AX^+_{rec}$) is a sound and complete axiomatization for
	%spencer23: switching this back again
	%joe6
	the language $\L(\S)$ for SEMs (resp., acyclic SEMs) with a
	% the language $\L(\S)$ for SEMs (resp., recursive SEMs) with a
	universal signature
	$\S$.
\end{theorem}

%joe6: added
As we shall see (Theorem~\ref{theorem:axioms-not-complete-for-SEMs}),
the assumption that $\S$ is universal is critical here;
Theorem~\ref{thm:completeness-for-SEMs} is not true in general without it.

%Joe1*: Now discuss how I defined AX^+ in my earlier paper, the
%theorems I proved in hte earlier paper.
%joe6
%section{GSEMs: a review}
\section{GSEMs}\label{sec:GSEMreview}
In this section, we briefly review the definition of
%joe6
%GSEM from \cite{HP20}; we encourage the reader to consult \cite{HP20}
GSEMs \cite{PH20}; we encourage the reader to consult \cite{PH20}
for more details and intuition.  We also
prove some results regarding the extent to which the language
%spencer34*:
% $\L(\S)$ characterizes GSEMs.
$\L(\S)$ characterizes GSEMs, and introduce the class of \emph{acyclic} GSEMs.
%joe34
%joe36
%We include a few results from (\cite{PH20}) to help set the scene
We include a few results from \cite{PH20} to help set the scene.
%spencer40: Love this sentence!
%joe35: :-)

The main purpose of causal modeling is to reason about a system's behavior
under intervention. A SEM can be viewed as a
function that
takes a context $\uu$ and an intervention $\YY \gets \yy$ and returns
a set of assignments to the endogenous variables
%joe26
(i.e., a set of outcomes),
%joe26: left solution here
%spencer28: totally agree.
namely, the set of all solutions to the structural equations after replacing
the equations for the variables in $\YY$ with
$\YY = \yy$.
%joe6: removed paragraph break
%
Viewed in this way, generalized structural-equations models (GSEMs) are a
%joe26: there's more to a GSEM than just the function
%generalization of SEMs. In a GSEM  is \emph{defined} as
generalization of SEMs. In a GSEM, there is
a function that takes a context $\uu$ and an intervention $\YY \gets \yy$ and returns a
%joe26
%set of solutions to the endogenous
%variables (along with a signature and set of allowed interventions).
%However, the solutions need not be determined by solving a set of
set of outcomes.
However, the outcomes need not be obtained by solving a set of
suitably modified
%spencer42:
%equations as they are for SEMs.
equations as they are for SEMs--they may be specified arbitrarily.
%spencer28: I like the wording in the other paper better.
% This relaxation gives GSEMs the
% ability to concisely represent situations with infinitely many
% variables, such as dynamical systems, and
% more flexibility than SEMs for handling finite-variable situations.
This relaxation gives GSEMs the
ability to concisely represent dynamical systems and other systems
with infinitely many variables, and
the flexibility to handle situations involving finitely many variables that
cannot be modeled by SEMs.

%spencer42: Moved para below as suggested by AAAI reviewer 2
% Because GSEMs don't have the structure that SEMs have by virtue
% of being defined in terms of structural equations, we may want to rule out certain unintuitive possibilities.
% In particular, we require that after intervening to set
% %joe6*: I think we should strengthen this, so it looks like the axiom
% %(and what we say below):
% %$Y \gets y$, all solutions satisfy $Y = y$.
% %spencer7*: Yes, we need this! This is actually what I did in PH20 in
% %the formal definition of GSEMs, but in the informal description I
% %used Y \gets y--I'll make a note to correct this in PH20.
% %joe26
% %$\XX \gets \xx$, all solutions satisfy $\XX = \xx$.
% $\XX \gets \xx$, all outcomes satisfy $\XX = \xx$.
%spencer6:
%This is in fact one of the
%axioms in a sound and complete axiomatization for
%SEMs given by Halpern~\nciteyear{Hal20}; see also \cite{GallesPearl98}.
% This is in fact the axiom D4 (effectiveness) given above.

%Requiring that all of these axioms hold
%guarantees that
%for
%all GSEMs with finitely many endogenous variables and finite ranges
%for each endogenous variable, there exists
%an equivalent SEM,
%that is, one that satisfies the same formulas,
%or equivalently for finite causal models,
%has the same solutions in all interventions and contexts (see Theorem~\ref{theorem:satisfies-same-formulas-equiv-has-same-solutions}).

Formally, a \emph{generalized structural-equations
	model (GSEM)} $M$ is a pair $(\S, \FF)$, where $\S$ is a signature,
and $\FF$ is a mapping from contexts and interventions to sets of
%joe26
%solutions.
outcomes.
%joe6
%More precisely, a signature $\S$ is a quadruple $(\U, \V,
%\R, \I)$ where, as before, $\U$ is a set of exogenous variables, $\V$
%is a set of endogenous variables, and $\R$ associates with every
%variable $Y$ in $\U \cup \V$ a nonempty, finite set $\R(Y)$ of
%possible values for $Y$; we extend $\R$ to subsets of $\V$ in the same
%way as before. However, we no longer require that $\U$, $\V$ or the sets
%$\R(Y)$ for $Y \in \U \cup\V$ be finite.  The mapping $\F$ is a function
As before, a signature $\S$ is a quadruple $(\U, \V, \R, \I)$, except
that we no longer require $\U$ and $\V$ to be finite, nor $\R(Y)$ to
be finite for all $Y \in \U \cup \V$.   The big difference is that
%spencer31: Cutting, doesn't make sense with the new notation
% now
$\FF$ is a function
$\FF: \I \times \R(\U) \to \P(\R(\V))$, where $\P$ denotes the powerset operation.
That is, it maps a context $\uu \in \R(U)$ and an allowed
%joe26
%intervention $I \in \I$ to a set of \emph{solutions} $\F(\uu, I)
%	\subseteq \P(\R(\V))$. As with SEMs, we denote these solutions
intervention $I \in \I$ to a set of \emph{outcomes} $\FF(\uu, I)
%joe34: the referee was right; this is a typo
%spencer40: Thanks for correcting it! :)
%\subseteq \P(\R(\V))$. As with SEMs, we denote these outcomes
\in \P(\R(\V))$. As with SEMs, we denote these outcomes
by $M(\uu, I)$.
%joe26
%As we said above, we require that solutions $\vv \in \F(\uu, \XX
%$
%spencer42: now we say it after.
%As we said, we require that each outcome $\vv \in \FF(\uu, \XX
We require that each outcome $\vv \in \FF(\uu, \XX
%spencer42: I don't think this needs a paragraph in this review
	%\gets \xx)$ satisfy $\vv[\XX] = \xx$.
\gets \xx)$ satisfy $\vv[\XX] = \xx$, since at a minimum, after intervening to set $\XX$ to $\xx$, the variables $\XX$ should actually have the values $\xx$.
%joe6: already said this
%  In the special case where all interventions are allowed, we take $\I =
%\I_{univ}$, the set of all interventions.
%joe6: added
Since the semantics of $\models$ as we have given it is defined in
terms of $M(\uu,I)$, we can define $\models$ for GSEMs in the
identical way.

%spencer6:
%We now make precise the sense in which GSEMs generalize SEMs.
%joe6
%  In \cite{PH20}, we showed that GSEMs generalize SEMs, in the following
%precise sense.
%
%joe24
%Peters and Halpern \nciteyear{PH20} showed that GSEMs generalize SEMs,
\fullv{Peters and Halpern \nciteyear{PH20} showed}
\shortv{It is shown in \cite{PH20}} that GSEMs generalize SEMs
in the following sense:
Two causal models $M$ and $M'$, which may either be SEMs or GSEMs,
%joe24: I odn't this we discuss them in this paper
%spencer26: Good catch!
%GSEMs (or causal constraints models, introduced later),
are \emph{equivalent}, denoted $M\equiv M'$, if they have the same
%joe24: line shaving
%signature and they have the same solutions, that is, if for all sets
%spencer27: This reads clearer to me
%signature and the same solutions, that is, if for all sets
%joe26
%signature, and they have the same solutions; that is, if for all
signature, and they have the same outcomes; that is, if for all
%spencer27: shaving
% sets of variables
$\XX \subseteq \V$, all
values $\xx \in \R(\XX)$ such that $\XX \gets \xx \in \I$, and all contexts $\uu
	\in \R(\U)$, we have
$M(\uu, \XX \gets \xx) = M'(\uu, \XX \gets \xx)$.

%joe6
%begin{theorem}\label{thm:GSEM-generalizes-SEM}
%spencer23*: should we say this is Theorem 3.1?
%joe21: I'm not sure I understand the question.  Did you mean should
%we say [Peters and Halpern, 2020; Theorem 3.1], I think that is
%marginally better.  (Use \cite[Theorem 3.1]{PH20})
%spencer24: Yes, perfect.
%spencer24*: But then, we should replace the bibliography entry for
%PH20 with one that does not reveal the authors, and just says
%'companion paper'. What's the standard way of writing that entry?
%joe22: I'm not sure that there is a standard way.  You can just
%replace the author field by ``Anonymous'', and say submitted to IJCAI
%(giving the submission number)
\begin{theorem}
  \label{thm:GSEM-generalizes-SEM}
	%spencer24:
	% \cite{PH20}
	\cite[Theorem 3.1]{PH20}
	For all SEMs $M$, there is a GSEM $M'$ such that $M \equiv M'$.
\end{theorem}
Recall that in the introduction we defined two models with signature
$\S$ to be  \emph{$\L(\S)$-equivalent} if they agree on
all formulas in $\L(\S)$.  Call a GSEM $M$ \emph{finitary} if, for all
%spencer26: added
contexts and
%joe26
%interventions, the set of solutions is finite.  Of course, a GSEM with
%joe34: line shaving
%interventions, the set of outcomes is finite.  Of course, a GSEM with
interventions, the set of outcomes is finite.  A GSEM with
a finite signature
%spencer7: added
(a finite GSEM)
is bound to be finitary, but even
%spencer7:
%GSEMs with
%non-finite signatures
infinite GSEMs
may be finitary.
%spencer7:
%The following theorem shows that
%joe7
%Peters and Halpern \cite{PH20} showed that
%joe24
%Peters and Halpern \nciteyear{PH20} showed that
%spencer44:
%\fullv{Peters and Halpern \nciteyear{PH20} showed}
\fullv{Peters and Halpern \nciteyear{PH20} showed that}
%joe34
%\shortv{It is shown in \cite{PH20}} that
\shortv{As shown in \cite{PH20},}
%spencer7:
%equivalence and $\S$-equivalence are coincide in finitary GSEMs.
equivalence and
%spencer20*:
% $\S$-equivalence
$\L(\S)$-equivalence
%spencer27:
%coincide in finitary GSEMs.
coincide in SEMs and finitary GSEMs.

%joe6*: moved here from appendix, rewrote to use new notation.
%spencer23*: Should we say this is theorem 2.1 of PH20?
%joe21: As aove; use
%spencer25*: As I noted in the introduction, this theorem is true, but
%\L(\S) is not usually the language we use for reasoning about
%infinite models.
\begin{theorem}
	\label{theorem:formulas-equiv-solutions}
	%spencer23: added
	%spencer24:
	% \cite{PH20}
	\cite[Theorem 2.1]{PH20}
	%joe21: remove the paragraph break on the next line
	%spencer24*: are you sure? It's harder to read and doesn't save a line.
%joe22: it's more standard ...
	If $M$ and $M'$ are finitary causal models over the same signature
	%spencer20:
	% $\S$, then $M \equivM'$ iff $M$ and $M'$ are $\S$-eqvivalent.
	$\S$, then $M \equiv M'$ iff $M$ and $M'$ are
	%spencer20*: replaced this everywhere below
	% $\S$-equivalent.
	$\L(\S)$-equivalent.
\end{theorem}
%spencer23*: moved proof to appendix and commented out because it's in PH20.

%joe6: moved example here
As the following example shows, the assumption that $M$ and $M'$ are
finitary is critical.
\begin{example}
	\label{example:infinite-solutions-not-characterized}
	Consider two GSEMs $M, M'$ with the same signature $\S = (\U, \V,
		\R, \I)$.
	$\V$ consists of countably many binary endogenous variables,
%joe26
%                that is, $\V = X_1, X_2, \dots$ and $\R(X_i) = \{0,
  %$
                that is, $\V = \{X_1, X_2, \dots\}$ and $\R(X_i) = \{0,
                1\}$ for
                %spencer28:
                % $i = 1, 2, \dots$.
                all $i$.
	%joe6: there should always be a comma after i.e. and e.g.
	%The models have only one context $\uu$ (i.e. $\U$ consists of one
	The models have only one context $\uu$ (i.e., $\U$ consists of one
%joe26
        %	trivial endogenous variable with a single value).
        	exogenous variable with a single value).
	There is only one allowed intervention, the null intervention $\emptyset$.
  %joe26
        %spencer28: wow, this reads a lot better.
        %joe27: :-)
%	The solutions to the models are as follows.
        	The outcomes of this intervention are as follows.
	%joe6
	%$\F(\uu, \emptyset)$ consists of all assignments to the $X_i$ where
	%only finitely many $X_i$ take the value 0.
          $\FF(\uu, \emptyset)$ consists of all assignments to the variables $X_i$ where
                %spencer28:
                %joe21: grammatically it's more correct with X_is, bu
                %it sound better this way, so let's leave it
	%only finitely many of the $X_i$s take the value 0.
only finitely many of the $X_i$ take the value 0.
	$\FF'(\uu, \emptyset)$ consists of all assignments to the $X_i$ where
	only finitely many $X_i$ take the value 1.
	%joe6
	%Note that for any finite subset of the variables, restricting the
	Note that for a finite subset of the variables, restricting the
%joe26
        %	solutions of either model
        	outcomes of either model
	to that subset yields all assignments to that subset.
	%joe6
	% Hence, any formula of the form
	Hence, a formula of the form
	%joe6
	%  $\<\emptyset\>(\varphi)$ will be false in both models if $\varphi$ is
	%  tautologically false,
	$\<\emptyset\>\varphi$ is false in both models if $\neg \varphi$ is
	valid,
	and true in both models otherwise, because $\varphi$ is a finite
	%joe6
	% formula and as such can only depend on finitely many variables.
%joe25: rewrite as ``and thus can depend on only finitely many variables''.
	formula and as such can depend only on finitely many variables.
	%joe6
	%  Hence the distinct models $M$ and $M'$ satisfy the same set of
	Hence, the distinct models $M$ and $M'$ satisfy the same set of
	causal formulas over $\L(\S)$.
	%joe6: I like to mark the end of examples.
	\bbox
\end{example}

%spencer34*: added
\subsection{Acyclic GSEMs}
\newcommand{\ancestor}{}

In this subsection, we introduce a class of GSEMs analogous to acyclic
SEMs. Just as many SEMs used in practice are acyclic, we expect that
%joe30
%many GSEMs of practical interest will also be acyclic. All the GSEMs
many GSEMs of practical interest will also be acyclic. For example, the GSEMs
constructed in \cite{PH20} to model dynamical systems are acyclic
according to our definition.
%spencer39: cutting this, consistency with below
% (after trivial modifications, discussed
%joe30
%later).
% below).

%spencer34*: Moved from below.
%joe30
%In SEMs, acyclicity is defined via the notion of
In SEMs, acyclicity is defined using the notion of
%joe29
%\emph{independence}. Recall from Section \label{sec:review} that given
\emph{independence}. Recall from Section~\ref{sec:review} that given
%joe30
%a SEM $M$ with endogenous variables $X$, $Y$, we say that \emph{$Y$ is
a SEM $M$ and endogenous variables $X$ and $Y$, we say that \emph{$Y$ is
  independent of $X$} (in context $\uu$) if the structural equation
$F_Y(\uu, \dots)$ for $Y$ does not depend on $X$.
An acyclic SEM is a SEM whose endogenous variables $\V$ can be totally ordered (for all contexts $\uu$) such that if $X \preceq_\uu Y$, then $X$ is independent of $Y$ in context $\uu$.
% (In the acyclic GSEMs constructed in \cite{PH20}, the ordering corresponds to time ordering; see below.)
% ($t < t \rimp X_t \preceq_\uu Y_{t'}$).
% The time ordering is really a partial ordering since t = t' is a possibility. We can then take any total ordering consistent with that partial ordering, as with SEMs.

%joe30: Seems somewhat circular.  YOu've just defined acyclic SEMs
%Also removed paragraph break
%This definition has a very natural interpretation in acyclic SEMs.
%joe34: added
We cannot use this definition for GSEMs, since there are no
%spencer40: typo
%equations.  But we an generalize an alternate characterization of
equations.  But we can generalize an alternate characterization of
acyclicity.
In acyclic SEMs, there is always a
unique outcome, and you cannot change the outcome for variables
preceding $X$ by intervening on $X$. More precisely, let
%joe30
%$\V_{\prec_\uu X} = \{Y \in \V | Y \prec_\uu X\}$. Then the outcome
%spencer35: Oops, used | instead of \mid
%joe31: I actually slightly prefer : anyway
$\V_{\prec_\uu X} = \{Y \in \V : Y \prec_\uu X\}$. Then the outcome
%$\V_{\prec_\uu X} = \{Y \in \V \mid Y \prec_\uu X\}$. Then the outcome
$\vv$ of doing $I; X \gets x$ in
context $\uu$ and the outcome $\vv'$ of doing $I; X \gets x'$ in context $\uu$ agree on $\V_{\prec_\uu X}$
($\vv[\V_{\prec_\uu X}] = \vv'[\V_{\prec_\uu X}]$).
%joe30: remoeed paragraph break
%
In words, in acyclic SEMs, changing variables later in the ordering does not affect variables earlier in the ordering. In fact, a SEM is acyclic if and only if there are orderings $\preceq_\uu$ such that this condition holds.
%   % Proof: If there is a cycle in the SEM, by using complete interventions, we can show that there is a cycle of variables such that changing one affects the next, which contradicts the existence of a satisfactory ordering.

%joe30
%This gives a very natural way to extend the definition of acyclicity
This gives a natural way to extend the definition of acyclicity
to GSEMs.
%joe34: said this already
%The definition of acyclicity for SEMs does not extend to
%GSEMs, because GSEMs do not have structural equations $F_X$. A GSEM
%$M$ only has outcomes $M(\uu, I) = \FF(\uu, I)$. But
Since the condition
$\vv[\V_{\prec_\uu X}] = \vv'[\V_{\prec_\uu X}]$ is a condition on
outcomes, it makes sense for GSEMs. Acyclic GSEMs may have multiple
%joe30
%solutions in general (see Example \ref{example:infinite-sem} below),
solutions,
%spencer35:
%so we need to generalize the condition slightly.
so we need to strengthen the condition slightly.
%joe30
%Given a set of outcomes $S$ and a subset $\YY$ of $\V$, define the
Given a set $S$ of outcomes and a subset $\YY$ of $\V$, define the
\emph{restriction
  of $S$ to $\YY$}, denoted $S[\YY]$, as $S[\YY] = \{\vv[\YY] \mid \vv
\in S\}$.

%joe30: This example seems out of place.  What's it an example of?
%What point does it illustrate?  It also breaks the flow
%spencer35: I intended it to illustrate that acyclic GSEMs may have
%multiple solutions (which isn't obvious, since acyclic SEMs
%don't). The point is to motivate the stronger condition (the two
%conditions are equivalent in the case of unique solutions). But now
%that I think about it, I think explaining how the GSEMs in GSEM.tex
%are acyclic will address that point, so we can leave this example
%out.
\commentout{
\begin{example}
    \cite{PH20}
    \label{example:infinite-sem}
    Let $M_{chain}$ be the GSEM
    with binary endogenous variables
    $X_1, X_2, \dots$
    (and a single context $\uu$)
    whose outcomes under any
  intervention are the solutions to the structural equations
  $X_1 = X_2$, $X_2 = X_3$,
  and so on. This GSEM is intuitively acyclic (the dependencies
  encoded by the equations only go from higher-numbered variables to
  lower-numbered ones,
  so there is no cycle of dependencies). However, it has two solutions
  under the empty
  intervention; all zeros, and all ones.
\end{example}

We now define acyclicity for GSEMs.
}
%joe30: \end{commentout}

\begin{definition}
  A GSEM $M$ is acyclic if, for all contexts $\uu$, there is a total
%joe34
  %  ordering $\prec_\uu$ of $\V$ such that
    ordering $\prec_\uu$ of $\V$ such that:
  %spencer35: Makes more sense below to include the quantifiers in the
  %labeld eqn
  % for all $X \in \V$, for all $x, x' \in \R(X)$, for all $I \in \I$, we have
%joe34: line shaving
%  the following holds.
      %   %spencer35:
  %     \begin{equation}
  %   \label{eq:2}
  %     M(\uu, I; X \gets x)[\V_{\prec_\uu X}] = M(\uu, I, X \gets x')[\V_{\prec_\uu X}].
  % \end{equation}
%joe31:
%  \begin{condition}
 %   \label{condition:stronger}
  \begin{description}
    \item[Acyc1.]
  For all $X \in \V$, for all $x, x' \in \R(X)$, for all $I \in \I$, we have
      $M(\uu, I; X \gets x)[\V_{\prec_\uu X}] = M(\uu, I, X \gets x')[\V_{\prec_\uu X}].$
%  \end{condition}
\end{description}
\end{definition}

%joe30: this is true, and we can put the example here, but I'm not
%sure what point it illustrates.  It also breaks the flow here
%It is easy to see that under this definition, Example
%\ref{example:infinite-sem} is acyclic.

It is natural to wonder whether this condition needs to involve all variables preceding $X$. After all, in SEMs, acyclicity is defined in terms of independence, and independence is defined pairwise. Indeed, the pairwise version of this condition is sufficient for SEMs; a SEM $M$ is acyclic if and only if for all contexts $\uu$, there is a total ordering $\prec_\uu$ such that
%spencer35:
the following holds.
%spencer35:
% if $Y \prec_\uu X$, then (for all $I, x, x'$)
% \begin{equation}
%   \label{eq:3}
%   M(\uu, I; X \gets x)[Y] = M(\uu, I; X \gets x')[Y].
% \end{equation}
%joe31
%\begin{condition}
\begin{description}
\label{condition:weaker}
\item[Acyc2.]
If $Y \prec_\uu X$, then for all $I, x, x'$, we have
   $M(\uu, I; X \gets x)[Y] = M(\uu, I; X \gets x')[Y].$
%\end{condition}
\end{description}

%joe30*: Added
%spencer35:
%Clearly (\ref{eq:2}) implies (\ref{eq:3}).
%joe31: I don't like ``Condition''
%Clearly Condition \ref{condition:stronger} implies Condition
%\ref{condition:weaker}.
Clearly Acyc1 implies Acyc2.
In SEMs, they are equivalent.

\begin{proposition} If $M$ is a SEM, then $M$ satisfies
  \label{proposition:acyc-conditions}
  %spencer35:
  %(\ref{eq:2}) iff $M$ satisfies (\ref{eq:3}).
%joe31
  %  Condition \ref{condition:stronger} iff $M$ satisfies Condition
%\ref{condition:weaker} (for a fixed context $\uu$).
 Acyc1 iff $M$ satisfies
Acyc2 (for a fixed context $\uu$).
\end{proposition}
%spencer35*: should I write a proof?
%joe31: Yes, if you have time. DONE tonight (trivial!)

%joe3
%However, it is not sufficient for GSEMs, in the sense that acyclicity
%it is not sufficient for GSEMs, in the sense that acyclicity
%defined with the pairwise version \eqref{eq:3} does not imply
%acyclicity defined with the full version \eqref{eq:2} for GSEMs. The
%weaker pairwise condition allows for ``acyclic'' GSEMs where changing
%variables later in the order affects variables earlier in the
%order. Consider the following example:
However, in GSEMs, the two conditions are not equivalent;
we claim that
%spencer35:
%the stronger condition (\ref{eq:2})
%joe31
%the stronger Condition \ref{condition:stronger}
the stronger condition Acyc1
is more appropriate
for characterizing acyclicity.
The following example illustrates why.

\begin{example}
	\label{example:switching-values}
  Define a GSEM $M$ with binary variables $A, B, C$, a single context $\uu$, allowed interventions $\I = \{A \gets 0, A \gets 1, B \gets 0, B \gets 1, C \gets 0, C \gets 1\}$, and the outcomes
  %spencer39: This looks bad.
%     \[M(\uu, C \gets 0) = \{(0, 0, 0), (1, 1, 0)\} \mbox{ and }\]
% \[M(\uu, C \gets 1) = \{(0, 1, 1), (1, 0, 1)\},\]
  \begin{align*}
    &M(\uu, C \gets 0) = \{(0, 0, 0), (1, 1, 0)\} \text{ and} \\
    &M(\uu, C \gets 1) = \{(0, 1, 1), (1, 0, 1)\}
  \end{align*}
where $(a, b, c)$ is short for $(A = a, B = b, C = c)$.
%joe34: Is this what you mean.
%spencer40: Yes, exactly!
%The outcomes for $A \gets a$ and $B \gets b$ are symmetric.
The outcomes for $A \gets a$ and $B \gets b$ are similar: for example,
after the intervention $A \gets a$, $A=a$ and $B = C \oplus a$.
%joe30: not the end
%\end{example}

%spencer35:
% $M$ is not acyclic when acyclicity is defined using \eqref{eq:2}. To
%joe31
%$M$ is not acyclic when acyclicity is defined using Condition
%\ref{condition:stronger}. To
%spencer46*: typo!
%$M$ is not acyclic when acyclicity is defined using Acyc2.
$M$ is not acyclic when acyclicity is defined using Acyc1.
To
%joe30
%see this, fix any ordering of the variables (the model is symmetric,
%we can take the ordering $A, B, C$ without loss of
%generality). Then intervening on the last variable (in this case $C$)
see this, fix an ordering of the variables; since the model is symmetric,
we take the ordering $A, B, C$ without loss of
generality. Then intervening on $C$, the last variable in the ordering,
changes the outcomes for the other two; $M(\uu, C \gets 0)[\{A, B\}] =
\{(0, 0), (1, 1)\}$, but $M(\uu, C \gets 1)[\{A, B\}] = \{(0, 1), (1,
%joe30
%0)\}$. We feel that this is the correct classification for $M$: $M$
%should not be acyclic. The situation that intervening on $C$ changes
%$
%spencer35:
%0)\}$.$
%joe31
%0)\}$, violating Condition \ref{condition:stronger}.
0)\}$, violating Acyc1.
%spencer35: para break $
%
This seems to us the correct classification: $M$
should not be acyclic. The fact that intervening on $C$ changes
the possible values for $(A, B)$, but both $A$ and $B$ precede $C$ in
%joe30
%$\prec_\uu$, is counterintuitive and never occurs in SEMs. However,
%spencer35:
%$\prec_\uu$, cannot occur in SEMs. However,
$\prec_\uu$, cannot occur in acyclic SEMs. However,
$M$ is acyclic when acyclicity is defined using
%joe31
%Condition \ref{condition:weaker}. This is
Acyc2. This is
because intervening on $C$ does not affect the possible values for $A$
%joe30
%(they are $A = 0, A = 1$ in both cases) or for $B$ (also $B = 0, B =
%1$ in both cases).
($A = 0$ and $A = 1$ in the two outcomes for each intervention) or for
$B$ ($B = 0$ and $B = 1$ in the two outcomes for each intervention).
%joe30
\bbox
\end{example}

%spencer35*: added
%spencer46*:
%All the GSEMs introduced in \cite{PH20}
As we said above, all the GSEMs introduced in \cite{PH20}
for modeling of dynamical systems, namely,
%spencer42*: Reviewer 2 wants us to define these. That seems unnecessary; what do you think? Maybe they wanted us to do that because the wording was a little different? Changing for consistency.
%ODE GSEMs (Section 4),
%joe36: I think that the reader will find the pointers to sections in
%the other paper confusing (and they mnight change in any case); I
%would just drop them.  That said, I don't think we should give
%details or define them here.  We don't use them in this paper, so
%I see no point.
%spencer43: I completely agree. Removing these.
% GSEMs for systems of ordinary differential equations (Section 4),
% GSEMs for hybrid automata
% (Appendix D.1), and GSEMs for rule-based models (Appendix D.2) are
GSEMs for systems of ordinary differential equations,
GSEMs for hybrid automata, and GSEMs for rule-based models are
%joe31*: we definitely do not want to get into the weeds here.  WE
%should change GSEM to make them acyclic.  I'm happy to tell a white
%lie here.
%acyclic, given suitable technical modifications. The order $\prec_\uu$
%acyclic, given suitable technical modifications. The order $\prec_\uu$
%spencer36*: Ok. Should I change the version of GSEM that I put in the appendix? This might take some time. If not, shouldn't we at least put a footnote?
%joe32*: If you don't have time, I would just leave it.  I can live
%with the white lie.  I think that we should ultimately change it.
acyclic.
The order $\prec_\uu$ in each case corresponds to
%spencer42:
%time;
the natural notion of time in the dynamical system;
intervening on variables at a given
time cannot affect variables earlier in time (or at the same time).
%joe31*: I think we should just cut all this.  It will just confuse
%the reader, and definitely not help our case
%spencer36*: I don't disagree... should I put it in the appendix?
%joe32: No; I would cut it altogether (and change GSEM when you have
%the change, so that it's true)
\commentout{
Here we present the case of ODE GSEMs; the other two cases are
similar.
%spencer35*: Is this clear? If so, is it too much detail? We could
%move it to the appendix.
In ODE GSEMs, each variable $X_i^t$ corresponds to the value of a dynamical variable $\mathcal{X}_i$ at the time $t \geq 0$. We can order the variables by taking $X_i^t \prec_\uu X_j^{t'}$ if $t < t'$ and breaking ties ($t = t'$) arbitrarily. For a given intervention and context, the outcomes of an ODE GSEM consist of all valid executions of an algorithm SOLVE-ODE-GSEM (Theorem 4.1 of \cite{PH20}). Intuitively, SOLVE-ODE-GSEM solves for an outcome step by step, proceeding forward in time. At each step, it either solves an initial value problem (steps 2b and 3) or sets up initial conditions (steps 1 and 2c). At each of these steps, when the value of a variable $X_i^t$ is defined, its possible values depend only on variables $X_j^{t'}$ with $t' < t$. If SOLVE-ODE-GSEM terminates in step 4, it outputs the outcome that assigns to each variable the value it was defined as during execution. If SOLVE-ODE-GSEM always terminated in step 4, then the set of outcomes output by some execution of SOLVE-ODE-GSEM, restricted to variables $X_i^s$ with $s \leq t$, would not change when a variable $X_j^{t'}$ with $t' \geq t$ is intervened on. Only the outcomes restricted to variables with $s > t'$ would change. This would imply that ODE GSEMs are acyclic. However, ODE GSEMs are not acyclic as defined in PH20, because SOLVE-ODE-GSEM does not always terminate in step 4. If SOLVE-ODE-GSEM encounters an unsolvable initial value problem in steps 2b or 3, it terminates immediately and outputs ``No solution''. Thus, intervening on a variable $X_j^t$ in such a way that an initial value problem starting at time $s \geq t$ is unsolvable means that some outcomes are not output by SOLVE-ODE-GSEM, thus affecting the possible outcomes restricted to variables with times less than $t$. To fix this, we modify SOLVE-ODE-GSEM to output partial solutions rather than reporting that it cannot find a solution. Specifically, instead of outputting ``No solution'' in steps 2b and 4 when an insoluble problem is encountered at time $t_i$, we introduce a special error value, set all variables $X_i^t$ with $t \geq t_i$ that are not intervened on to that value, and output the resulting outcome. Now the same logic as before implies that intervening on $X_j^{t'}$ does not change the outcomes restricted to $t < t'$. This change to SOLVE-ODE-GSEM corresponds to a modification in the ODE GSEM conditions (ODE1, ODE2, ODE3$'$) to allow partial solutions.
}
%joe35: \end{commentout}

%spencer35*:
%joe34: line hsaving
%Next we discuss our approach to axiomatizing this important class of GSEMs.
%joe30
%In SEMs, acyclicity was captured by the axiom D6. The semantics of D6
%correspond to the pairwise condition \eqref{eq:3}. To capture the
%definition of acyclicity for GSEMs (which uses the stronger condition
%\eqref{eq:2}), we need a modification of D6, which we call
In SEMs, acyclicity corresponds to the axiom D6, which
%spencer35:
%captures the pairwise condition \eqref{eq:3}.
%joe31
%captures the weaker Condition \ref{condition:weaker}.
captures the weaker Acyc2.
To get an axiom for
acyclicity for GSEMs,  we need a modification of D6 that captures the
%spencer35:
%stronger condition (\ref{eq:2}).
%joe31
%stronger Condition \ref{condition:stronger}.
stronger Acyc1.
%joe30
%which we call
%D6$^+$. D6$^+$ will be discussed in detail later. Over finite
%signatures, D6$^+$ is the following.
% For $X \in \V$, $x, x' \in \R'(X)$, $\YY$ a finite subset of $\V$,
% $\yy \in \R'(\YY)$, and $I \in \I'$, we write $X \affects_{x, x',
% \ww, I} \YY$ as an abbreviation for the formula
% \[[I; X \gets x](\YY = \yy) \wedge [I; X \gets x'](\YY \neq \yy).\]
%joe30: removed paragraph break
%joe30*: affects looks to me the same as the symbol used in D6.  It
%would be good to have a symbol that was different.
%spencer35: on it!
%spencer35*: You're right, we should just introduce D6^+ in full right
%away. Moved from below.
%joe34*: different abbreviations.  Note that we need to consider a set
%of interventions.
%spencer40: Thanks for catching that. I think the intuition is great.
But we cannot express the full Acyc1, because variables may have
infinite ranges, the set $\V_{\prec_\uu} X$ may be infinite, and the
set of interventions may be infinite.
Thus, we consider a finitary version of Acyc1.
%joe36: cut next sentence; it seems redundant
%It suffices that it hold for all finite choices.
%spencer42: para break

%joe36
%Given a finite set of variables $X_1,
%\ldots, X_k$, finite sets $U_i \subseteq \R(X_i)$, for $i=1,\ldots,
Given a finite set $\{X_1,
\ldots, X_k\}$ of variables, finite sets $U_i \subseteq \R(X_i)$, for
$i=1,\ldots,
k$, and a finite set $\I'$ of interventions, let $X_i \leadsto_{\I',U_1,
  \ldots, U_k,
  %spencer42: typo
  %X_1, \ldots, X_k} X_{-i}$
X_1, \ldots, X_k} \XX_{-i}$
%spencer42:
%say that
describe the following conditions:
%spencer42: rewrite
%intervening on $X_i$ after performing some interventions in $\I'$ affect
after performing some intervention in $\I'$, intervening on $X_i$ affects
%spencer42: clarification
%the values of
the joint values of the variables
%spencer42: typo
%$X_{-i} = (X_1, \ldots, X_{i-1}, X_{i+1}, \ldots, X_k)$,
$\XX_{-i} = (X_1, \ldots, X_{i-1}, X_{i+1}, \ldots, X_k)$.
%spencer42:
% which
Moreover, for some intervention on $X_i$, the joint value
is in $U_{-i} = U_1
  \times \cdots \times U_{i-1} \times U_{i+1} \times \cdots \times U_k$.
  That is, $X_i \leadsto_{\I,U_1, \ldots, U_k,
    %spencer42: typo
  %X_1, \ldots, X_k} X_{-i}$ is an abbreviation of $
X_1, \ldots, X_k} \XX_{-i}$ is an abbreviation of
%spencer40: Overfull hbox, so switching to align
\begin{equation} \nonumber
\begin{split}
&\lor_{x,x' \in U_i, \vec{y} \in U_{-i}, I \in \I'} \\ &[I; X\gets
    x](\vec{X}_{i-1} = \vec{y}) \land [I; X \gets x'](\vec{X}_{i-1}
    \ne \vec{y}).
  \end{split}
  \end{equation}
  %joe34*: cut all this:
  \commentout{
  For $X \in \V$, $x, x' \in \R'(X)$, $\YY$ a finite subset of $\V$,
$\yy \in \R'(\YY)$, and $I \in \I'$, we write $X \affects_{x, x',\ww,
  I} \YY$ as an abbreviation for the formula
%spencer35: I moved this from below, but it was wrong.
%\[[I; X \gets x](\YY = \yy) \wedge [I; X \gets x'](\YY \neq \yy).\]
%joe31: put this inline.  I think it looks better
%\[[I; X \gets x](\YY \neq \yy) \wedge \<I; X \gets x'\>(\YY = \yy).\]
$[I; X \gets x](\YY \neq \yy) \wedge \<I; X \gets x'\>(\YY = \yy).$
%joe34*: we need to add intuition here, both for the abbreviation and
%the axiom.  I agree with the reviewer that it's extremely hard to
%parse!  You hae some intuition later, but it comes too late.  I
%suspect that we might be better off with a different abbreviation
%altogether. Suppose you take $X_i \succ_{X_1, \ldots, X_k, T, U_1,
%\ldots, U_k} X_{-i}$ to mean that changing the value of X has no
%effect on the value of X_1, \ldots, X_{i-1}, X_{i+1}, ..., X_k,
%provided that (a) we restrict to interventions in T followed by an
%intervention X_i gets x_i for x_i \in U_i, and X_j is restricted to
%taking values in U_j.  Then you can take the axiom to just say
%$\/_{i \in k} X_i \succ_{X_1, \ldots, X_k, T, U_1,
%\ldots, U_k} X_{-i}$, which is easier to follow.
%spencer40: I think this is what you changed it to. I think it's
%great. I'm ambivalent between \leadsto and \succ.
  }

%joe31:
%spencer39: changed n to k for consistency with below and with D6
Consider the following axiom:
\begin{itemize}[leftmargin=\parindent + 5pt, align=left, labelwidth=\parindent, labelsep=5pt, itemsep=2pt]
\item[D6$^+$.]
  %spencer35: I think this is clearer.
  %$\bor_{i = 1}^n \band_{x, x' \in U_i, \yy \in S_i, I \in T} \neg X_i
  %$
%joe34*: another attempt
%  $\bor_{i = 1}^k \neg \bor_{x, x' \in U_i, \yy \in S_i, I \in T} X_i
%   \affects_{x, x', \yy, I} \{X_1, \dots, X_{i - 1}, X_{i+1}, \dots, X_k\}$,
%   where $T$ is a finite subset of $\I'$, $U_i$ is a finite subset of
%   $\R(X)$ for $1 \leq i \leq k$, and $S_i = U_1 \times \dots \times
  %   U_{i-1} \times U_{i + 1} \times \dots \times U_k$.
  $\bor_{i = 1}^k \neg (X_i \leadsto_{\I',U_1, \ldots, U_k,
  X_1, \ldots, X_k} X_{-i})$.
\end{itemize}
(There is an instance of this axiom for all choices of $\I', U_1, \ldots,
%joe36
%U_k,$ and $X_1, \ldots, X_k$.
U_k,$ and $X_1, \ldots, X_k$.)
%
%spencer35:
%joe34*: cut
\commentout{
Over finite signatures, it is straightforward to verify that D6$^+$ is
equivalent to
%joe30
%\begin{itemize}
% \begin{itemize}[leftmargin=\parindent + 5pt, align=left, labelwidth=\parindent, labelsep=5pt, itemsep=2pt]
% \item[D6$^+$ (finite signatures)]
%spencer38: put in eqn for reference later
\begin{equation}
  %joe34*: why is it obvious that you can get rid of the intervention
  %I?  We use it in D6
  %spencer40: Good check! I'd forgotten my original argument, but I
  %think the following is correct: for finite signatures, the instance
  %of the axiom with $\I' = \I_{univ}$ is stronger than all other
  %instances, so we can drop the other instances (this is what getting
  %rid of the intervention does).
  \label{eq:d6-finite}
  \bor_{i = 1}^k \neg (X_i \affects
  \{X_1, \dots, X_{i - 1}, X_{i+1}, \dots, X_k\})
\end{equation}
% \end{itemize}
where $X \affects \YY$ abbreviates the formula
$\bor_{x, x' \in \R(X), \yy \in \R(\YY), I \in \I}
%spencer35: clearer
%\<I; X \gets x\>(\YY = \yy) \wedge [I; X \gets x'](\YY \neq \yy)$.$
X \affects_{x, x', \yy, I} \YY$.
%joe31: removed paragraph break
%
%spencer35:
%Intuitively, D6 states that every subset of the variables has an
Over finite signatures, D6$^+$ states that every (finite) subset of the variables has an
%joe30
%element (the maximum element under $\preceq_uu$) which does not
%affect
element that does not
%spencer35: moved below
% (the maximum element under $\preceq_\uu$) that does
%spencer35:
% not affect the others.
affect the others, in the sense that intervening on that element does not change the outcomes restricted to the others. Over infinite signatures, D6$^+$ is weaker: it makes the same statement, but only for interventions of the form $I; X_i \gets x$ and $I; X_i \gets x'$ where $I \in T$ and $x, x' \in U_i$, and only if the changed value is in $S_i$.
%spencer38: removed para break
%joe34: \end{commentout}
%
}
%spencer35:
D6$^+$ is clearly sound in acyclic GSEMs, since in an acyclic GSEM, every finite subset of variables has a maximum element under $\preceq_\uu$, and this element does not affect the others.
%spencer35: DONE make sure we do this
%spencer38*:
Moreover, in SEMs, D6$^+$ implies D6, which Halpern used
%spencer44: shaving
%to axiomatize acyclic SEMs.
for acyclic SEMs.
This is because D6$^+$ holds iff there is an ordering $\prec_\uu$ of the endogenous variables such that Acyc1 holds, and D6 holds iff there is an ordering such that the weaker condition Acyc2 holds.
%spencer38*:
% In Section 6 we show that D6$^+$ is in fact complete for acyclic GSEMs.
%joe34*: the arguments works.
%spencer40: Amazing!!
%In Section \ref{sec:finiteGSEM} we show that an axiom system
%containing D6$^+$ is in fact complete for finite acyclic GSEMs.
%We believe this is also true for infinite GSEMs, but we have not yet
%finished working out the details.
%
%joe36: isn't it just Section 6?
%spencer43: You're absolutely right!
%As the results of the following sections show, D6$^+$ captures acyclicity.
As Theorem \ref{theorem:GSEMs} shows, D6$^+$ captures acyclicity.
%spencer35*: Is it important to explain how it implies D6?
%joe31: I think it's useful to say a few sentences
% DONE

% DONE put GSEM paper in appendix

%joe6: added label
\section{Axiomatizing finite GSEMs}\label{sec:finiteGSEM}
%spencer6:
%The language for SEMs makes sense for finite GSEMs.
%joe6: added next two sentences
%The language $\L(\S)$ given above for SEMs makes perfectly good sense
Our goal is to provide a sound and complete axiomatization of GSEMs.
We start with finite GSEMs, that is, GSEMs over a finite signature;
in the next section, we consider arbitrary GSEMs.  Note that
the language $\L(\S)$ given above for SEMs makes perfectly good sense
for finite GSEMs; the semantics of the language for GSEMs is identical
to the semantics for SEMs.
%joe6
%GSEMs are more flexible than SEMs, and indeed, there is a sound and
%complete axiomatization for finite GSEMs which is a strict subset of
%$AX^+$.
Because GSEMs are more flexible than SEMs, they do not satisfy all the
axioms in $AX^+$.  As we now show, a strict subset of $AX^+$ provides a sound
and complete axiomatization of finite GSEMs.

\begin{definition}
	%joe6: I prefer ``basic'' to ``primitive''; it also has the merit of
	%being shorter, so I changed it globally.  It's easy enough to change back.
	%    $AX^+_{primitive}$ consists of axiom schema D0, D1, D2, D4, D7,
	%spencer7: I like basic as well.
	$AX^+_{basic}$ consists of axiom schema D0, D1, D2, D4, D7,
	D8, and inference rule MP.
\end{definition}

%spencer1: should this be a special case of the result for infinite GSEMs?
%spencer42*: spencer1 had a point! I think it is a special case, and I
%think it's worth saying so. Do you agree?
%joe36.  Soundness should follow from Proposition 6.1 and Theorem
%6.2.  But I don't see why completeness follows.  It may be that there
%are extra axioms in finite GSEMs.
%spencer43: I see. I think you're right.
\begin{theorem}
	\label{theorem:finite-GSEMs}
	$AX^+_{basic}$ is sound and complete for finite GSEMs.
\end{theorem}

%spencer23*: Moved proof to appendix.
%spencer23*: added
%joe34
%The proof of this and all other results not in the main text can be
The proof of this and all other results can be%
%spencer39:
%found in the appendix.
%joe36: again, no more supplementary material
%spencer44*: Should we move more proofs to main text?
%spencer44:
%found in the supplementary material.
%joe38: do you mean for the full paper?  If so, then we certainly
%should move the proof sketch for Theorem 6.2 to the main text.  I'm
%OK with leaving everything else as is.
%By the way, I had labeled the
%reference as PH21, not HP21.  I changed it to HP21 in joe.bib
%spencer45: Yes, I meant the full paper. Thanks!
\shortv{found in the full paper \cite{HP21}.}
\fullv{found in Appendix \ref{appendix:additional-proofs}.}
%spencer35*: added. I hope I have time to prove this!
Let $AX^+_{basic, rec}$ consist of the axioms in $AX^+_{basic}$ along with axiom schema D6$^+$. Then

\begin{theorem}
  \label{theorem:acyclic-finite-GSEMs}
  $AX^+_{basic, rec}$ is sound and complete for acyclic finite GSEMs.
\end{theorem}

\fullv{As shown by Peters and Halpern \nciteyear{PH20},}
\shortv{As shown in \cite{PH20},}
if $\S$ is a universal
signature, then SEMs over $\S$ are equivalent in expressive power to
%spencer28:
%GSEMs that satisfy $AX^+$.
%joe27
%finite GSEMs that satisfy $AX^+$.
%spencer30: typo
%finite GSEMs that where all the axioms in $AX^+$ are valid.
%joe34: line shaving
%finite GSEMs where all the axioms in $AX^+$ are valid.
finite GSEMs where the axioms in $AX^+$ hold.

%joe1: I added this to joe.bib.  We should have a general discussion
%of order of
%authors at some point.
%spencer2: For sure! I'm completely unaware of the conventions here.
%\cite{UAIsub}.
%joe1*: We should formally state the result we proved in our earlier paper.
%spencer2: check!
\begin{theorem}
	\label{theorem:SEMs-equiv-finite-GSEMs-univ}
	%joe6
	%Fix a signature $\S$ such that $\I = \I_{univ}$. For all finite GSEMs
	%with signature $\S$ satisfying $AX^+(\S)$, there is an equivalent
	%SEM, and vice versa.
	\cite{PH20} If $\S$ is a universal signature, then for every finite GSEM
	over $\S$ that satisfies the axioms of $AX^+(\S)$ there is an
	equivalent SEM over $\S$, and for every SEM over $\S$ there is an
	equivalent
	%spencer20:
	finite
	GSEM over $\S$ that satisfies the axioms of $AX^+(\S)$.
\end{theorem}
Since equivalence is the same as $\L(\S)$-equivalence for finitary
models (Theorem \ref{theorem:formulas-equiv-solutions}),
this immediately implies the following.
%joe6
%\begin{theorem}
\begin{corollary}\label{corollary:finite-GSEMs-satisfying-axioms}
	%joe6
	%Fix a signature $\S$ such that $\I = \I_{univ}$. $AX^+(\S)$ is
	%complete for the class of finite GSEMs satisfying $AX^+(\S)$.
	%\end{theorem}
	If $\S$ is a universal signature, then $AX^+(\S)$ is a sound and
	complete axiomatization for $\L(\S)$ for finite GSEMs
        %joe36: added
        over $\S$
        satisfying $AX^+(\S)$.
\end{corollary}
%spencer23: moved this proof to the appendix

%spencer7: this is a trivial statement--it's the completeness part
%that is nontrivial.
%Saying that $AX^+$ is sound for the set of GSEMs satisfying $AX^+(\S)$
%seems like an almost trivial statement.
%spencer28:
%Saying that $AX^+$ is sound and complete for the set of GSEMS
%joe27: why not just cut this
%Saying that $AX^+$ is sound and complete for the set of finite GSEMs
%satisfying $AX^+(\S)$ seems like an almost trivial statement.
%joe26
%There are two points to be made regarding this. First, we give a much
%spencer28*: This isn't really true; we don't consider GSEMs satisfying all the axioms in AX^+ in the next section. I think we can just cut this.
% We give a much
% more natural
% characterization of the set of GSEMs satisfying these axioms in the
% %joe26
% %next section.  Second,
% next section.
%spencer28:
% Moreover,
%joe21
%However, despite its apparently trivial nature,
%joe37
%Despite its apparently trivial nature,
Although it may seem trivial,
%spencer42: typo
%Theorem~\ref{corollary:finite-GSEMs-satisfying-axioms} does not hold in
Corollary~\ref{corollary:finite-GSEMs-satisfying-axioms} does not hold in
general for non-universal signatures,%
%spencer31:
%spencer44*: Moving shell game example to main text of full version
\shortv{
as \cite[Example 3.6]{PH20} shows.
% as the following example,
% %joe24
% %due to Peters and Halpern \nciteyear{PH20} shows.
% \fullv{due to Peters and Halpern \nciteyear{PH20},}
% %joe26
% %\shortv{given in \cite{PH20},}
% \shortv{taken from \cite{PH20},}
% shows.
%spencer39:
%This example, included in the appendix for completeness
%joe36: again, supplementary material
%spencer44:
%This example, included in the supplementary material
%joe38: it's still in the appendix; I definitely would move to the
%main part of the paper.
%spencer45: I can't fit it in the main text in the AAAI version. It's
%moved to the main text in the Arxiv version.
%joe39: makes sense
%spencer45*:
% This example, included in Appendix \ref{appendix:shell-game}
% for completeness
% %joe34
% %\ref{example:shell-game-SEM}), is a GSEM over a finite signature $\S$
% (Example \ref{example:shell-game-SEM}),
This example
is a GSEM over a finite signature $\S$
that satisfies the axioms of $AX^+(S)$ but is not equivalent to a
SEM.
}
\newcommand{\shellGameSEMExample}{
  \begin{example}\label{example:shell-game-SEM}
	Suppose that Suzy is playing a shell game with two shells.
  %spencer28: for consistency with line shaving in other paper
  %One of the shells conceals a dollar, and the other shell is empty.
  One of the shells conceals a dollar;  the other shell is empty.
  %spencer28:
	% Suzy can choose to flip over one or the other of the two shells.
  Suzy can choose to flip over a shell.
  If
	she does, the house flips over the other shell. If Suzy picks shell 1,
  %spencer28:
	%which hides the dollar, she wins the dollar, otherwise she wins
which hides the dollar, she wins the dollar; otherwise she wins
	%joe6
	%nothing. This example can be modeled by a GSEM $M_{shell}$ with
	nothing. This example can be modeled by a GSEM $M_{shell} = (\S_{shell},\FF_{shell})$ with
	two binary endogenous variables $S_1, S_2$
	describing whether shell 1 is flipped over and shell 2 is flipped
	over, respectively, and a binary endogenous variable $Z$ describing
	the change in Suzy's winnings.
	(The GSEM also has a trivial exogenous variable whose range has size
	1, so that there is only one context $\uu$.)
	%joe6
	%That defines $\U, \V$, and $\R$;
	%we set $\I = \{S_1 \gets 1, \S_2 \gets 1\}$; and $\F$ is
	%defined as follows:
	%solutions $M'(\uu, S_1 \gets 1) = \{(S_1 = 1, S_2 = 1, Z = 1)\}$ and
	%$M'(\uu, S_1 \gets 2) = \{(S_1 = 1, S_2 = 1, Z = 0)\}$.
	That defines $\U_{shell}, \V_{shell}$, and $\R_{shell}$;
	we set $\I_{shell} = \{S_1 \gets 1, \S_2 \gets 1\}$; finally, $\FF_{shell}$ is
	defined as follows:
	%joe6*: something got garbled here.  There seems to be a whole chunk.
	%added all the material below from our other paper
	%missing
	%spencer7: Thanks!
	\begin{align*}
		\FF(\uu, S_1 \gets 1) & = M_{shell}(\uu, S_1 \gets 1)    \\
		                     & = \{(S_1 = 1, S_2 = 1, Z = 1)\}  \\
		\FF(\uu, S_2 \gets 1) & = M_{shell}(\uu, S_2 \gets 1)    \\
		                     & = \{(S_1 = 1, S_2 = 1, Z = 0)\}.
	\end{align*}
%joe24
%	As shown by Peters and Halpern \nciteyear{PH20},
        \fullv{As shown by Peters and Halpern \nciteyear{PH20},}
%joe26
        %        \shortv{As shown in \citeyear{PH20},}
                \shortv{As shown in \cite{PH20},}
        $M_{shell}$
        is a GSEM
	where all the axioms in $AX^+(\S_{shell})$ are valid (note that in the
	case of D9, validity is vacuous, since none of the relevant
%joe26
        %	interventions are in $\L(\S_{shell})$.
        	interventions is in $\L(\S_{shell})$.
	However, $M_{shell}$ is not equivalent to a SEM;
%joe26
        %	no SEM $M'$ over $\S_{shell}$  can have the solutions
        	no SEM $M'$ over $\S_{shell}$  can have the outcomes
	$M_{shell}(\uu, S_1 \gets 1) = \{(S_1 = 1, S_2 = 1, Z = 1)\}$ and
	$M_{shell}(\uu, S_1 \gets 2) = \{(S_1 = 1, S_2 = 1, Z = 0)\}$.
	This is because in a SEM,
	the value of
	$Z$ would be specified by a structural equation $Z = \F_Z(\U, S_1, S_2)$. This
%joe26
%	cannot be the case here, since there are two solutions having
	cannot be the case here, since there are two outcomes having
	$S_1 = S_2 =
%joe26
%	1$, but with different values of $Z$.\bbox
  %$
        	1$, but with different values of $Z$.  \bbox
        \end{example}
}
\fullv{
  as the following example, taken from \cite{PH20}, shows.
  \shellGameSEMExample
}

%spencer28*: With all the proofs in the appendix, we don't really need
%more space, but if we do, we could probably move this example to the
%appendix, since it is not new.
%joe27: I agree; you may want to think about whether there's something
%useful that can be added in its place.  If not, just leave it.
%spencer31*: Cutting the example
\commentout{
%joe10: stray reference
%\nciteyear{PH20}.
\begin{example}\label{example:shell-game-SEM}
	Suppose that Suzy is playing a shell game with two shells.
  %spencer28: for consistency with line shaving in other paper
  %One of the shells conceals a dollar, and the other shell is empty.
  One of the shells conceals a dollar;  the other shell is empty.
  %spencer28:
	% Suzy can choose to flip over one or the other of the two shells.
  Suzy can choose to flip over a shell.
  If
	she does, the house flips over the other shell. If Suzy picks shell 1,
  %spencer28:
	%which hides the dollar, she wins the dollar, otherwise she wins
which hides the dollar, she wins the dollar; otherwise she wins
	%joe6
	%nothing. This example can be modeled by a GSEM $M_{shell}$ with
	nothing. This example can be modeled by a GSEM $M_{shell} = (\S_{shell},\FF_{shell})$ with
	two binary endogenous variables $S_1, S_2$
	describing whether shell 1 is flipped over and shell 2 is flipped
	over, respectively, and a binary endogenous variable $Z$ describing
	the change in Suzy's winnings.
	(The GSEM also has a trivial exogenous variable whose range has size
	1, so that there is only one context $\uu$.)
	%joe6
	%That defines $\U, \V$, and $\R$;
	%we set $\I = \{S_1 \gets 1, \S_2 \gets 1\}$; and $\F$ is
	%defined as follows:
	%solutions $M'(\uu, S_1 \gets 1) = \{(S_1 = 1, S_2 = 1, Z = 1)\}$ and
	%$M'(\uu, S_1 \gets 2) = \{(S_1 = 1, S_2 = 1, Z = 0)\}$.
	That defines $\U_{shell}, \V_{shell}$, and $\R_{shell}$;
	we set $\I_{shell} = \{S_1 \gets 1, \S_2 \gets 1\}$; finally, $\FF_{shell}$ is
	defined as follows:
	%joe6*: something got garbled here.  There seems to be a whole chunk.
	%added all the material below from our other paper
	%missing
	%spencer7: Thanks!
	\begin{align*}
		\FF(\uu, S_1 \gets 1) & = M_{shell}(\uu, S_1 \gets 1)    \\
		                     & = \{(S_1 = 1, S_2 = 1, Z = 1)\}  \\
		\FF(\uu, S_2 \gets 1) & = M_{shell}(\uu, S_2 \gets 1)    \\
		                     & = \{(S_1 = 1, S_2 = 1, Z = 0)\}.
	\end{align*}
%joe24
%	As shown by Peters and Halpern \nciteyear{PH20},
        \fullv{As shown by Peters and Halpern \nciteyear{PH20},}
%joe26
        %        \shortv{As shown in \citeyear{PH20},}
                \shortv{As shown in \cite{PH20},}
        $M_{shell}$
        is a GSEM
	where all the axioms in $AX^+(\S_{shell})$ are valid (note that in the
	case of D9, validity is vacuous, since none of the relevant
%joe26
        %	interventions are in $\L(\S_{shell})$.
        	interventions is in $\L(\S_{shell})$.
	However, $M_{shell}$ is not equivalent to a SEM;
%joe26
        %	no SEM $M'$ over $\S_{shell}$  can have the solutions
        	no SEM $M'$ over $\S_{shell}$  can have the outcomes
	$M_{shell}(\uu, S_1 \gets 1) = \{(S_1 = 1, S_2 = 1, Z = 1)\}$ and
	$M_{shell}(\uu, S_1 \gets 2) = \{(S_1 = 1, S_2 = 1, Z = 0)\}$.
	This is because in a SEM,
	the value of
	$Z$ would be specified by a structural equation $Z = \F_Z(\U, S_1, S_2)$. This
%joe26
%	cannot be the case here, since there are two solutions having
	cannot be the case here, since there are two outcomes having
	$S_1 = S_2 =
%joe26
%	1$, but with different values of $Z$.\bbox
  %$
        	1$, but with different values of $Z$.  \bbox
\end{example}
} %spencer31: end commentout
%joe3
%The existence of a finite GSEM satisfying $AX^+$ which is not
%equivalent to any SEM has rich implications.
The existence of a finite GSEM satisfying $AX^+$ that is not
equivalent to an SEM has a significant implication.
%spencer42*: On reflection, this theorem is kind of silly because
%$AX^+(\S)$ is just a weaker version of $AX^+(\S')$, where \S' is \S
%but with all interventions allowed. The axioms are still sound if
%we're able to talk about disallowed interventions. Do you agree?
%joe36: I'm not sure what talking about disallowed interventions has
%to do with it.  If \S' is \S with all interventions allowed, thne
%it's certainly true that all the axioms in \AX(S) are valid in
%\AX(S').  However, the theorem is not at all trivial (and actually
%requires proof).  To show that AX^+ is not complete, you have to show
%that there's a formula in the language that is valid in the set of
%SEMs with signature \S that is not provable in AX^+.  This does not
%follow immediately from the fact that there is a GSEM that's not
%equivalent to any SEM.  You actually prove it in the proof of Theorem 5.6.
%spencer43: Sorry, I meant that the axioms are still complete (not
%``still sound'') over \S'. That is, I just meant Corollary 5.5.) So
%I'm not saying that the theorem is trivial. Just that it's not
%surprising that removing statements from the set of axiom instances
%makes the axioms weaker.
%joe37: I'm not following your point, Spencer.  We're *not* removing
%axioms.  We're saying that the axioms we had before aren't enough.
%We need to add more axioms if we remove the universality requirement
%spencer44: Sorry. We're not removing axioms. We're removing instances of axioms (those corresponding to disallowed interventions).

\begin{theorem}
	\label{theorem:axioms-not-complete-for-SEMs}
	%joe3: It's either complete or it's not
	%$AX^+(\S)$ is not complete for SEMs of signature $\S$ in general
	%spencer3*: I mean, we know that it is not complete for a particular
	% signature $\S$, that of the shell game.
	%joe4: It should follow that it's not complete for any ``larger''
	%signature (in which you can express the same story), so the only
	%technical question is what happens for ``small'' signatures
        %joe35: no paragraph break
        %
        % How should I best say "there exists a signature $\S$ where some interventions are not allowed in $\S$ and $AX^+(\S)$ is not complete for SEMs of signature \S?
  %spencer42*: rewrite
	% $AX^+(\S)$ is not complete for SEMs of signature $\S$
	% when some interventions are not allowed in $\S$, although it is
	% sound.
  There is a (non-universal) signature $\S$ for which $AX^+(\S)$, although sound, is not complete for SEMs of signature $\S$.
\end{theorem}

\section{Axiomatizing infinite GSEMs}\label{sec:non-finite}
%joe1: I don't like the term ``types of infiniteies in this context''
%There are two types of infinities that distinguish general GSEMs from
%joe6
%In standard SEMs, there are only finitely many variables and the range
%of each variable is finite.
%In GSEMs, we allow both the number of variables to be infinite and
%their ranges to be infinite.
Things change significantly in infinite GSEMs.  To see just one of the
problems, note that if $X$ is a variable with infinite range, then
%The language $\L(\S)$ still makes sense for infinite signatures $\S$
%(i.e. signatures with infinitely many variables or with infinite
%ranges). However, there are a couple of serious problems with the
%axioms $AX^+$. The first is that if any variable $X$ has an infinite
%range, the
instances of D2 corresponding to $X$, namely
%joe24: saving space
%$$[\YY \gets \yy](\bor_{x \in \R(X)} X = x),$$
\fullv{$$[\YY \gets \yy](\bor_{x \in \R(X)} X = x),$$}
\shortv{$[\YY \gets \yy](\bor_{x \in \R(X)} X = x),$}
are no longer in the language, since the disjunction is infinitary.
%joe6*: added rest of paragraph and following paragraph
Moreover, if $\R(X)$ is uncountable and the language includes all formulas
of the form $X=x$ for $x \in \R(X)$, then the language will be
uncountable.  While there is no difficulty giving semantics to this
uncountable language, there seem to be nontrivial technical problems
when it comes to axiomatizations.

On the other hand, suppose that, for example, the range of $X$ is the
real numbers.  In practice, we do not want to make statements like $X=
	\pi^3 - e$.  It should certainly suffice in practice to be able to
mention explicitly only countably many real numbers.  (Indeed, we
expect that, in practice, it will suffice to talk explicitly about only finitely
%joe34
%many real numbers.)  Similarly, we expect that it will suffice to talk
many real numbers.)  Similarly, it should suffice to talk
explicitly about only countably many variables and interventions.  To
get a countable language, we thus proceed as follows.

%joe6*: cut; no longer true
\commentout{
	To address this problem, we augment $\L(\S)$ with a new type of atomic event, namely $X \in A$, where $X \in \V$ and $A \subseteq \R(X)$. This atomic event is satisfied by $\vv \in \R(\V)$ if $\vv[X] \in A$. We denote the augmented language by $\L_\in(\S)$.

	However, with no restrictions on the cardinality of $\V$ and the cardinality of $\R(X)$ for $X \in \V$, the language $\L_\in(\S)$ has unbounded cardinality. This is undesirable from multiple standpoints. From a pragmatic standpoint, standard techniques for proving completeness are difficult to apply to such languages. Similarly, this language is far more expressive than necessary for modeling applications, and it is difficult to argue that the complexity adds anything useful. For these reasons, we would like to restrict $\L_\in(\S)$ to a countable sublanguage.

	%spencer6: rewrite
	To obtain a countable language, we need the set of interventions $\YY
		\gets \yy$ appearing in formulas to be countable, or equivalently, the
	sets of $\YY$ and $\yy$ appearing in interventions must each be
	countable. In addition, we need the set of atomic statements $X = x$
	to be countable, or again equivalently, the set of variables $X$ and
	the set of values $x$ for each variable must each be
	countable. Finally, we need the set of atomic statements $X \in A$ to
	be countable, so the set of $A$ that can appear in these statements
	must be countable. Following this high-level overview, we define the
	class of restricted languages $\L_{\WW, \R', \C, \A}$ as follows.
}
%joe6: \end{commentout}

%joe
%Let $\WW$ be an arbitrary countable subset of $\V$; we say that the
%spencer42: egregious typo
%Given a signature $\S = (\U,\V,\F,\I)$ or $\S = (\U, \V, \FF, \I)$,
Given a signature $\S = (\U,\V,\R,\I)$,
%joe18: ``arbitrary'' is implicit
%let $\WW$ be an arbitrary countable subset of $\V$; we call the
let $\WW$ be a countable subset of $\V$; we call the
%joe26
%elements of $\WW$ are \emph{named variables}.
elements of $\WW$ \emph{named variables}.
%joe18
%For each named variable $X$, let $\R'(X)$ be an arbitrary
For each named variable $X$, let $\R'(X)$ be a
%joe6*
%countable subset of $\R(X)$;
%spencer7*: I think you mean for \R'(X) to be countable, right?
%joe7: yes I did, but isn't that what I wrote?
%spencer8: You had "let \R'(X) be an arbitrary subset of \R(X), except that..."
% maybe I'm missing something.
% otherwise the set of formulas $[I](X = x)$ for fixed I, X is uncountable.
%subset of $\R(X)$,
countable subset of $\R(X)$,
except that we require that (a) if $\R(X)$ is
finite, then $\R'(X) = \R(X)$ and (b) if $\R(X)$ is infinite, then so
is $\R'(X)$.
The elements of $\R'(X)$ are called \emph{named values}.
%joe6*: no need
%For each named variable $X$, let $\C(X)$ be a countable subset of
%$\P(\R'(X))$; the elements of $C$ are called \emph{named value
%  subsets}.
%Next, let $\A$ be an arbitrary countable subset of $\P(\WW)$; the
%elements of $\A$ are called \emph{named variable subsets}.
%Finally, for $\XX \in \A$, let $\R'(\XX)$ be an arbitrary countable
%subset of $\times_{X \in \XX} \R'(X)$.
%sublanguage of $\L(\S)$. Intuitively, it consists of all
%formulas that mention only named entities; named variables, named
%values, (infinite) named subsets of the named values, and (infinite)
%named subsets of the named variables. It is easy to see that
%restricted languages are countable.
%spencer7: Rewording so we don't have to write $\I' \inter \I$ below.
%Finally, let $\I$ be a countable set of interventions $\XX \gets \xx$ such that
Finally, let $\I'$ be an arbitrary countable subset of $\I$, except
that we require that if $\XX \gets \xx \in \I'$, then
%joe7:
%$\XX \subseteq \WW$ and $\xx \subseteq \R'(X)$.
$\XX \subseteq \WW$ and $\xx \subseteq \R'(X)$ and
%joe6*:  This turns out to be necessary to deal with D3 and D6 in the
%comleteness proof.
%spencer7: interesting!
%joe7
%We assume that \emph{$\I'$ is closed under finite differences
we assume that \emph{$\I'$ is closed under finite differences
	with $\I$}, so that if $I_1 \in \I'$, $I_2 \in \I$, $(I_1 - I_2)
	\union (I_2 - I_1)$ is finite, and $I_2 = \XX \gets \xx$, where
$\XX \subseteq \WW$ and $\xx \in \R'(\XX)$, then $I_2 \in \I'$.
That is, if we are willing to talk about the intervention $I_1$, and
$I_2$ is an allowable intervention that differs from $I_1$ only in
how it sets a finite number of variables, all of which we are willing
to talk about, as well as the values that they are set to, then
we should be willing to talk about $I_2$ as well.
%spencer46* LATER we should point out that this assumption excludes the shell game in the finite case.
%joe6*: shortened
%Now we can define $\L_{\WW, \R', \C, \A}$.  Say that an atomic event
%$X = x$ is \emph{named} if $X \in \WW$ and $x \in \R'(X)$, and an
%atomic event $X \in A$ is \emph{named} if $X \in \WW$ and $A \in \C$.
%Say that an intervention $\YY \gets \yy$ is \emph{named} if $\YY \in
%\A$, $\yy \in \R'(\YY)$. Say that a basic causal formula $[\YY \gets
%  \yy]\varphi$ is \emph{named} if $\YY \gets \yy$ is named, and
%$\varphi$ is a Boolean combination of named atomic events. $\L_{\WW,
%  \R', \C, \A}(\S)$ consists of all Boolean combinations of named
%basic causal formulas.
The language $\L_{\WW, \R', \I'}(\S)$ consists of Boolean combinations
of basic causal formulas $[\YY \gets \yy]\phi$ where $\YY \gets \yy
	%spencer7: changed to requiring that \I' subseteq \I
	% \in \I \inter \I'
	\in \I'$
%joe18
%and $\phi$ is a Boolean combination of event $X =
% x$ where $X \in \WW$ and
and $\phi$ is a Boolean combination of events of the form  $X =
	x$, where $X \in \WW$ and $
	%spencer18:
	% s
	x
	\in \R'(X)$. $\L_{\WW, \R', \I'}(\S)$ is clearly a
sublanguage of $\L(\S)$. Intuitively, it consists only of entities
(variables, values, and interventions) that can be named.  Since there
are only countably many entities that can be named, it easily follows
that $\L_{\WW, \R', \I'}(\S)$ is countable.

This sublanguage, although not as expressive as the full language, is
still extraordinarily expressive.
%spencer18: moved above
For example, if the exogenous variables are $X_t$ for $t$
ranging over the real numbers, we could choose $\WW$
to be the subset
of $\{X_t \mid t \in \mathbb{R}\}$ for which $t$ is rational.
Likewise, if
%spencer18:
% the variables
each variable
$X_t$
ranges over the real numbers, we could choose
%spencer18:
% $\R'(X)$
$\R(X_t)$
to be the
rationals.

We are interested in axiomatizing classes of GSEMs
%spencer7:
%using essentially using
%joe7
%using essentially
essentially using
subsets of the axioms in $AX^+$, but it seems that we need one
new inference rule.  While we keep axiom D2, it applies only to
variables $X$ such that $\R(X)$ is finite.  However, even if $\R(X)$
is infinite, we still want to be able to conclude something like
%joe26
%$[\YY \gets \yy](\exists x (X=x)$: after setting $\YY$ to $\yy$, $X$
$[\YY \gets \yy](\exists x (X=x))$: after setting $\YY$ to $\yy$, $X$
takes on \emph{some} value.  Of course, we cannot say this, since we
have no existential quantification in the language.%
%joe24: cut for abstracts
\fullv{
\footnote{Extending the language to allow for such quantification may
	be of interest, but we do not explore that in this paper.  In any
	case, we   believe that the language that we are proposing should
	suffice for many applications of interest.}
}
Although it is far from obvious,
%joe24
\fullv{
as we shall see in our completeness proof,}
the following rule of
inference plays the same role as D2 for variables $X$ with infinite
%spencer7*:
%domains:
%joe7: this feels a bit like overkill to me.  I think we can get away
%with saying ``\phi mentions x if x appears as a symbol in \phi''.
%spencer8: Fair enough. I think I just had to clarify the meaning to myself--I wasn't immediately clear about the intervention part of the formula.
%spencer28: consistency
% domains.
ranges.
%spencer18: cutting this, as discussed above.
% Before we can state it, we define formally what it means for
% a causal formula to mention a value.
% Say that an event $\phi$ \emph{mentions} a value $x \in \R(X)$
% if some atomic event in the Boolean combination $\phi$ is $X = x$; a basic formula $[\YY \gets \yy]\phi$ \emph{mentions} $x$
% if either $\phi$ mentions $x$, or $X \in \YY$ and $\YY[X] = x$; and a
% causal formula $\psi$ \emph{mentions} $x$ if any basic formula in the
% Boolean combination $\psi$ mentions $x$. Now we can state D2$^+$.
%spencer7*: should this be here?
%Suppose that $X$ is a variable such that $\R(X)$ is infinite,
%spencer15: updating D2^+
\begin{itemize}
	\item[D2$^+$.]
	      %joe14: removing paragraph break
	      %
	      %spencer7:
	      % %that mention
	      % spencer15*: updating D2^+
	      %spencer7*: I think this is a little clearer (reverted)
	      % Suppose that $\phi \rimp [\YY \gets \yy]\psi$ is a formula in
	      % $\L_{\WW,\C,\I'}(\S)$,
	      % that mentions
	      % only the elements
	      % $x_1, \ldots, x_k \in \R'(X)$,
	      %spencer7*: to avoid ambiguity
	      %and $x_{k+1} \in \R'(X) - \{x_1, \ldots, x_k\}$.
	      % and
	      % either $\R(X) = \{x_1, x_2, \dots, x_k\}$, or
	      % there is a value $x_{k+1} \in \R'(X) - \{x_1, \ldots, x_k\}$.
	      Suppose that $S \subseteq \R'(X)$ is a finite subset of values of $X$ that contains
	      all the values of $X$ mentioned in the formula $\phi \rimp [\YY
			      \gets \yy]\psi$,
	      %joe14: you can't say S \neq \R(X), since you haven't specified S yet
	      %    and if $S \neq \R(X)$, contains an additional value not mentioned
	      %    in the formula.
	      %spencer15: Hmm, that is a bit confusing.
	      %spencer15:
	      % and if
	      and
	      some value in $\R'(X)$ not in the formula if there is such a value.
	      %spencer7: going back to old version now that "mention" is defined
	      %Suppose that $\phi$ is a formula in $\L_{\WW, \C, \I'}(\S)$,
	      %$\psi$ is an event of $\L_{\WW, \C, \I'}(\S)$,
	      %$\phi$ and $\psi$ only mention the elements $x_1, \ldots, x_k \in \R'(X)$,
	      %there is $x_{k+1} \in \R'(X) - \{x_1, \ldots, x_k\}$,
	      %spencer7*: We need to say that the intervention does not set X here
	      %or it makes no sense.
	      % To see this, suppose YY \gets yy sets X to x, where x is distinct from x_1, ... x_{k + 1}.
	      % Then the premise is true independent of the model and context
	      % even if \psi = true, so the inference clearly can't follow.
	      % Wait a minute--I'm an idiot.
	      % We know that YY <- yy can set X only to one of the x_1 ... x_k.
	      % But no issues can arise in this case,
	      % because then it can't be that \psi can rule out all of the X = x_i.
	      %because there we could never have
	      % and_i [\YY \gets \yy](\psi \rimp (X \ne x_i)) unless \psi was false.
	      %spencer7: going back to old version
	      %and $\YY \gets \yy$ is an intervention that does not set $X$ (that is, $X \notin \YY$).
	      %joe7
	      %Then from \phi \rimp \land_{i=1}^{x_{k+1}} [\YY \gets \yy](\psi
	      %spencer15*: updating D2^+
	      % Then from $\phi \rimp \land_{i=1}^{k+1} [\YY \gets \yy](\psi
	      % \rimp (X \ne x_i))$  infer $\phi \rimp [\YY \gets \yy]\neg \psi$.
	      Then from $\phi \rimp \land_{x \in S} [\YY \gets \yy](\psi
		      \rimp (X \ne x))$  infer $\phi \rimp [\YY \gets \yy]\neg \psi$.
\end{itemize}
% spencer15: need extra condition now
% Note that since we have assumed that $\R'(X)$ is infinite if $\R(X)$
% is infinite, there will always be an element $x_{k+1}$ in $\R'(X)$
Note that if $\R(X)$ is infinite, since we have assumed that $\R'(X)$
is infinite if $\R(X)$
%joe14:
%is, there will always be an element $x_{k+1}$ in $\R'(X)$
is, there will always be an element in $\R'(X)$
that is not mentioned in $\phi$ or $\psi$.

While D2$^+$ may not look anything like D2, we can show that in the
case of variables $X$ with finite
%spencer28: consistency
% domain,
range,
it is equivalent to D2
%joe34a
\fullv{in the sense made precise in the following result.}
\shortv{in the following precise sense:}

\begin{proposition}
	%spencer23: added
	\label{prop:d2-plus}
	%joe14
	%  If we replace D2 by D2$^+$ in $AX_{basic}$, then we can
	%derive D2 for variables with finite domains. Moreover,
  %spencer28: changing $AX_{basic}^\dag$ to $AX_{basic}^*$ everywhere, since that is consistent with the introduction and with the statement of Theorem 6.2.
	If $AX_{basic}^*$ is the result of replacing D2 with D2$^+$ in
	$AX_{basic}^+$, then we can
	derive D2 for variables with finite
  %spencer28:
  % domains
  ranges
  in $AX_{basic}^*$. Moreover,
	%joe7
	%D2$^+$ is derivable in $AX_{basic}$, in the sense that if
	%joe14
	%D2$^+$ is derivable in $AX_{basic}$, in the sense that
%joe26
%	D2$^+$ is derivable in $AX_{basic}^+$, in the sense that
	D2$^+$ is derivable in $AX_{basic}^+$ for variables $X$ with
        finite range, in the sense that
	%spencer15: updated D2^+
	% if $\phi \rimp \land_{i=1}^{S}$
	%joe14: might as well use notation
	%if $\phi \rimp \land_{x \in S}
	if $AX_{basic}^+ \vdash \phi \rimp \land_{x \in S}
		%spencer15: I think this is a typo
		% \<\YY \gets \yy\>
		[\YY \gets \yy]
		(\psi
		%spencer15: updated D2^+
		% $\rimp (X \ne x_i))$ is provable in $AX_{basic}$
		%joe14
		%\rimp (X \ne x))$ is provable in $AX_{basic}$
		\rimp (X \ne x))$
	and $\R(X)$ is finite,
	%spencer16: typo
	% then so is $\phi \rimp [\YY \gets \yy]\neg \phi$.
	%joe14
	%then so is $\phi \rimp [\YY \gets \yy]\neg \psi$.
	then  $AX_{basic}^+ \vdash \phi \rimp [\YY \gets \yy]\neg \psi$.
\end{proposition}

While D2$^+$ is unnecessary for finite GSEMs, it is necessary for
%joe18
%infinite GSEMs.  Let $AX_{basic}^*(\S,\W,\R',\I')$ consist of all the
infinite GSEMs.
Let $AX_{basic}^*(\S,\WW,\R',\I')$ consist of all the
axioms and inference rules in $AX_{basic}^+(\S)$ together with D2$^+$,
%joe18
%restricted to formulas  in $\L_{\W,\R',\I'}(\S)$.
restricted to formulas  in $\L_{\WW,\R',\I'}(\S)$.
%spencer28*: Added.
Then $AX_{basic}^*(\S, \WW, \R', \I')$ is sound and complete for GSEMs
%joe34: line shaving
%over $\S$ (this is a special case of Theorem \ref{theorem:GSEMs}
over $\S$ (see Theorem \ref{theorem:GSEMs}
below).%
%joe34*: please check!
%spencer40: This example looks great!
\footnote{We remark that the soundness of D2$^+$ depends on the fact
that we have assumed no structure on the domain, so the only way
we have of comparing variable values is by equality.  If we assumed an
ordering on the domain, so that, for example, we could write
$X \ge x$
in addition to $X=x$ and $X \ne x$, then D2$^+$ would no longer be sound.
For example, taking $\phi = \true$, $\psi = (X > 2)$,
%spencer40: Notation typo
%$\S = \{1,2\}$,
$S = \{1,2\}$,
and $\vec{Y} = \emptyset$,  from $X > 2 \rimp (X \ne 2 \land
X\ne 1)$, we would not want to infer $X \le 2$!  While we can extend
D2$^+$ to deal with $>$ and other ``nice'' relations,
%joe34: decided not to include this,
%Roughly speaking, the idea is to
%define, for every formula $\sigma$, an
%equivalence relation $\sim_\sigma$ on values of $X$ such that $x
%\sim_\sigma x'$ iff, in all outcomes $v$, $\sigma[v[x/X]]$ has the same
%truth value as $\sigma[v[x'/X]]$ (i.e., $\sigma$ cannot distinguish $x$
%and $x'$). If the only predicate is equality, as in our setting, all
%values not explicitly mentioned in a formula are in the same
%equivalence class. But this is not true with, say, $>$.  As long as
%there only finitely many equivalences classes for each formula, we can
%get an analogue to D2$^+$.  However,
pursuing this topic would take us too far afield here.}

Considering GSEMs also helps explain the role of some of the other
axioms.
%joe24: moved up from below
A GSEM $\S$ is \emph{coherent}
if for all interventions $\XX \gets \xx$ and $\XX
	\gets \xx; \YY \gets \yy$ in $\I$, if
%spencer19: added
$\YY$ is finite,
%spencer19:
% $\vv \in \F(\uu,\XX \gets \xx)$
$\vv \in \FF(\uu,\XX \gets \xx)$,
and $\vv[\YY] =\yy$, then $\vv \in \FF(\uu,\XX \gets \xx; \YY \gets
	\yy)$.  The intuition for coherence is straightforward: if we think of
  the assignments in $\FF(\uu,\XX \gets \xx)$ as representing the
  %spencer46: LATER this is a little confusing. Assignments can't be close to contexts.
``closest'' assignments to $\uu$ for which $\XX = \xx$ holds, and $\YY
	= \yy$ already holds in some assignment $\vv \in \FF(\uu,\XX \gets
	\xx)$, then $\vv$ is one of the assignments closest to $\uu$ where
both $\XX = \xx$ and $\YY = \yy$ hold, so $\vv$ should be in
$\FF(\uu,\XX \gets \xx; \YY \gets \yy)$.
%joe19*: We may want to say something about ``strong coherence'',
%where \YY is infinite, and leave it to future work to try to
%characterize that axiomatically.
%spencer20*:
%joe24
%spencer28: We have room; putting it back
% \fullv{
Note that this intuition applies equally to the case where $\YY$ is infinite (``strong coherence''). However, this seems to be harder to axiomatize, because the intervention $\XX \gets \xx; \YY \gets \yy$ is not guaranteed to be in $\I'$. We leave the problem of axiomatizing ``strong coherence'' to future work.
% }
%joe24
%joe34
%spencer40: Awesome!!
%As we show, D3 correponds to coherence.
As we show, D3 correponds to coherence and D6$^+$ corresponds to acyclicity.
%spencer26:
% As the name suggests, D10(a) corresponds to there being at least one solution
%joe26
%As their names suggest, D10(a) corresponds to there being at least one
%least one solution to
%each intervention, and D10(b) corresponds to there being at most one
%%spencer26:
%% solution, so D10 (i.e., the combination of 10(a) and 10(b)) correspond
%solution, so D10 (i.e., the combination of 10(a) and 10(b)) corresponds
%spencer28:
%As their names suggest, D10(a) corresponds each intervention having at
%spencer44: shaving
% As their names suggest,
D10(a) corresponds to each intervention having at
%spencer28:
%least one outcome,  and D10(b) corresponds to each intervention having
least one outcome (in any given context),  and D10(b) corresponds to each intervention having
at most one outcome,
so D10 (i.e., the combination of 10(a) and 10(b)) corresponds
%joe26
%to there being a unique solution to each intervention.
to each intervention having a unique outcome.
%spencer26: moved from below
%joe25: I'm not sure it's worth saying ``which is one of the main
%results of this paper''
This is made precise in Theorem \ref{theorem:GSEMs}
%below, which is one of the main results of this paper.
below.

%joe24
%spencer26:
% On the other hand, D5, D6, and D9 do not seems meaningful in GSEMs.
%spencer35:
%On the other hand, D5, D6, and D9 do not seem meaningful in GSEMs.
On the other hand, D5 and D9 do not seem meaningful in GSEMs.
%spencer26*: rewriting
% D5 and D9 do not have analogues if we have infinitely many variables
% with infinite ranges
% (since we cannot express $\ZZ = \zz$ in that case there are infinitely
% many variables, and there are uncountably many interventions of the
% form $\YY \gets \yy$ if $\YY = \V - \{X\})$).
%joe34
%D5 and D9 do not have analogues if we have infinitely many variables,
They do not have analogues if we have infinitely many variables,
since we cannot express $\ZZ = \zz$, and there are uncountably many
%joe25: I would remove the parens here.
%spencer27: Hmm. How about
%complete interventions (of the form $\YY \gets \yy$ for $\YY = \V - \{X\}$).
complete interventions (interventions of the form $\YY \gets \yy$ for $\YY = \V - \{X\}$).
%joe34: I think that we can live without this for now.
\fullv{
They also do not seem to represent particularly interesting properties,
%spencer26: added
since their scope is very limited (D5 always mentions all variables,
%joe26
%whereas D9 only applies to complete interventions).
whereas D9 applies only to complete interventions).
}

Let $\G^\emptyset(\S)$ denote the class of GSEMs over $\S$.
%joe35: we never use =1
%Let $\G^{\ge 1}(\S)$, $\G^{\le 1}(\S)$, and $\G^{=1}(\S)$ denote the
Let $\G^{\ge 1}(\S)$ and $\G^{\le 1}(\S)$ denote the
class of GSEMs over
$\S$ where
%spencer28: cut
% there
%joe26
%is at least one, at most one, and exactly one solution to each
%intervention; be the class of coherent GSEMs over $\S$.
%joe35
%each intervention has at least one, at most one, and exactly one
%outcome; let $\G^{coh}$ denote the class of coherent GSEMs over $\S$;
each intervention has at least one and at most one
outcome, respectively; let $\G^{coh}$ denote the class of coherent
GSEMs over $\S$;
%joe34
let $\G^{acyc}$ denote the class of acyclic GSEMs over $\S$.
%intervention; be the class of coherent GSEMs over $\S$.
%spencer18: typo
% Give
Given
%joe18
%a subset $A$ of $\{D3, D6, D10(a), D10(b)\}$, let $\A$ be the
%spencer22*:
% a subset $A$ of $\{$D3, D6, D10(a), D10(b)$\}$, let $\A$ be the
%joe34
%a subset $A$ of $\{$D3, D10(a), D10(b)$\}$, let $\A$ be the
a subset $A$ of $\{$D3, D6$^+$, D10(a), D10(b)$\}$, let $\A$ be the
%spencer22*:
% corresponding subset of $\{coh, acyc, \ge 1, \le 1\}$.  Let
%spencer28:
%corresponding subset of $\{coh, \ge 1, \le 1\}$.  Let
%joe34
%corresponding subset of $\{coh, \propertyge, \propertyle\}$.
corresponding subset of $\{coh, acyc, \propertyge, \propertyle\}$.
Let
%spencer24:
%$AX_{basic,A}^*(\S,\WW,\R',\I')$ be the axioms system consisting of
$AX_{basic,A}^*(\S,\WW,\R',\I')$ be the axiom system consisting of
the axioms and rules of inference of $AX_{basic}^*$ together with the
axioms in $A$, restricted to the language $\L_{\WW,\R',\I'}(\S)$.
%Again, we often omit the ``$(\S,\WW,\R',\I')$'' if it is clear from context.
%spencer18*: added
Let $\G^\A$ be the class of GSEMs satisfying
%spencer40: shaving
% all of
the properties in
$\A$; that is,
$\G^\A = \bigcap_{P \in \A} \G^P$.
%joe24: line shaving
%spencer26: Undone since this doesn't fall on a line boundary anymore.
Then

%\commentout{
%We said above the D3 corresponds to coherence.  This is true, but if
%we want to satisfy D3 and D10 simultaenously, the
%the set of interventions expressible in the language (i.e., $\I
%\inter \I'$) is sufficiently rich.   The problem already arises in
%Example~\ref{example:shell-game-SEM}).   In that example, both $\<S_1
%  \gets 1\>(S_2 = 1 \land Z=1)$,  and $\<S_2 \gets 1\>(S_1 = 1 \land
%Z=0)$ held.  If $S_1\gets 1; S_2 \gets 1$ were an allowed intervention, then
%D3 would require both $\<S_1 \gets 1, S_2 \gets 1\>Z=0$ and
%$\<S_1 \gets 1, S_2 \gets 1\>Z=1$ to hold.  There is no coherent
%The only reason that the
%example satisfied both D3 and D10 is that $S_1 \gets 1; S_2 \gets 1$
%was not an allowed intervention.
%}

%Using the
%spencer6:
%restricted sublanguages
%restricted languages
%defined above, we can now state the
%joe6
%main results of this paper. Let $\L(\S)_{\WW, \R', \C}$ be an
%main results of this paper. Let $\L_{\WW, \R', \I'}(\S)$ be an
%arbitrary
%spencer6:
%restricted sublanguage
%restricted language
%with signature $\S$.
%We now define several subclasses of GSEMs over signature $\S$

\begin{theorem}
	\label{theorem:GSEMs}
	%joe6
	%$AX'^+_{basic}$ is sound and complete for the class of GSEMs with
	%signature $\S$ over language $\L(\S)_{\WW, \R', \C}$.
  %joe26
  %spencer28*: This sentence is redundant.
%   If $A \subseteq \{$D3, D10(a),D10(b)$\}$ and $\A$ is the
%   %spencer28:
%         %corresponding subset of $\{coh, \ge 1, \le 1\}$, then
% corresponding subset of $\{coh, \propertyge, \propertyle\}$, then
        $AX^*_{basic,A}(\S,\WW,\R',\I')$ is sound and complete for the class $\G^{\A}$
	of GSEMs with
	signature $\S$ over language $\L_{\WW, \R', \I'}(\S)$.
\end{theorem}
%spencer37*: Proof sketch. Some material moved from below.
%spencer40: I don't think we need this sentence anymore.
% For reasons of space, we defer the full
% proof of Theorem \ref{theorem:GSEMs} to the
% %spencer39:
% % appendix.
% supplementary material.
%joe34
%However, since the completeness proof requires several nontrival ideas
%spencer45*: Saving this for shortv, and moving proof sketch for fullv.
\shortv{
We remark that  the completeness proof requires several
%spencer45: typo
%nontrival ideas
nontrivial ideas
beyond what is needed for the analogous results for SEMs (and
%joe35: shaving even more
%other
%spencer40: shaving
% soundness and completeness
%analogous results in modal logic).
%joe39
%modal logics).
modal logics); see the full paper for details.
}
\fullv{
  %spencer45:
  %\section{Proof Sketch for Theorem \ref{theorem:GSEMs}}
%spencer40:
% \prf (Sketch):
% \prf (of Theorem \ref{theorem:GSEMs}, sketch)

  %spencer37: Rewrote this, since we don't present the proof of Theorem 6.2 in the main text.
%   For completeness, we proceed in the same spirit as the proof of
%   Theorem~\ref{theorem:finite-GSEMs}.
%   But there are a number of new
% subtleties.   In Theorem~\ref{theorem:finite-GSEMs}, a maximal consistent
% %joe10
% %$C$ set was used to construct a causal model $M$.  We defined the
% set $C$ was used to construct a causal model $M$.  We defined the
% %spencer31: Changed \F(\YY \gets \yy, \uu) to \F(\uu, \YY \gets \yy) everywhere.
% function $\FF$ in $M$ by taking $\vv \in \FF(\uu, \YY \gets \yy)$ if
% $\<\YY \gets \yy\>(\V = \vv) \in C$.  But if there are infinitely many
% variables, $\V = \vv$ is an infinitary formula, so is not in the
% language.
%spencer40:
  %spencer45:
%In this section, we sketch the completeness portion of Theorem \ref{theorem:GSEMs}.
  Here we sketch the completeness portion of Theorem \ref{theorem:GSEMs}. We defer the full proof
  to%
  \shortv{the full paper.}
  \fullv{Appendix \ref{appendix:additional-proofs}.}
%spencer40:
% To prove completeness, we
We proceed in the same spirit as for finite GSEMs (Theorem \ref{theorem:finite-GSEMs}). The proof of that theorem
  %spencer39:
%(in the appendix)
%joe35: we're in the supplementary material, I assume
%spencer41: Yes, you're right.
%(in the supplementary material)
  uses essentially the same technique as that used by
Halpern \nciteyear{Hal20}. We briefly sketch the details here:
Suppose that $\varphi \in \L(\S)$ is \emph{consistent
with $AX^+_{basic}(\S)$}
%joe6
(i.e., we cannot prove $\neg \phi$ in $AX^+_{basic}(\S)$).
%joe6: don't need to talk about the axioms
%Then (by definition) the set of causal
%formulas $S$ containing just $\varphi$
%and $AX^+_{basic}(\S)$ is
%consistent. Extend this set to a maximal consistent set $C \supseteq
%joe6: this is not the definition of maximal consistent set
%S$ (i.e. for every causal formula in $\L(\S)$ this set contains either
%this formula or its negation).  Then $\phi$ can be extended to a
%maximal consistent set $C$ of
%spencer13: I think the statement above is not supposed to be commented, adding
Then  $\phi$ can be extended to a maximal consistent set $C$ of
formulas; that is, $\phi \in C$, every finite subset $C'$ of $C$ is
consistent with $AX^+_{basic}(\S)$ (i.e., the conjunction of the
formulas in $C'$ is consistent with $AX^+_{basic}(\S)$), and no
strict superset $C^*$ of $C$ has the property that every finite subset
of $C^*$ is consistent with $AX^+_{basic}(\S)$.  Standard arguments
show that, for every formula $\psi \in \L(\S)$, either $\psi$ or
$\neg \psi$ must be in $C$; moreover, every instance of the axioms in
%spencer13: Added more facts about C.
%% $AX^+_{basic}(\S)$ is in $C$
$AX^+_{basic}(\S)$ is in $C$.
%joe11*: there is a subtlety about what ``closure under provability''
%means if you have rules of inference like D2.  Once you have all
%axioms in C (which we already said), the only property you need is
%closure under implication: if \phi and \phi -> \psi are both in C,
%then so is \psi.  We could add this if you want, although it follows
%easily from the maximal consistency
%%Finally, $C$ is closed under provability--if a formula $\psi$ is
%provable from formulas in $C$, then $\psi \in C$.
%this formula or its negation).
%spencer14: Makes sense. I agree that all we need is closure under
%implication, and that it's not worth stating here.
Define a finite GSEM $M^C$ with signature $\S$ as follows. For all
contexts $\uu$ and allowed interventions $\YY \gets \yy \in \I$,
define
$\FF(\uu, \YY \gets \yy) = \{\vv \in \R(\V) \mid \<\YY \gets \yy\>(\V =
	\vv) \in C\}$.
%joe6: added to help the reader
Thus, the formulas in $C$ tell us how interventions work in $M^C$.
%spencer13*: added, refactoring.
%joe18
%Fix an arbitrary context $\uu$. To finish the proof, we need to show
%spencer37: Finally
To finish the proof, we need to show that $M^C$ is a GSEM, and that it models $\phi$. That is, given a fixed context $\uu$, $(M^C, \uu) \models \phi$.

However, in the case of infinite signatures, there are a number of new
subtleties. We can no longer define the function $\FF$ in $M$ by taking $\vv \in \FF(\uu, \YY \gets \yy)$ if
$\<\YY \gets \yy\>(\V = \vv) \in C$, because $\V = \vv$ is an infinitary formula, so is not in the language.
We deal with this by saying that the outcome $\vv$ is in $\FF(\uu, \YY
\gets \yy)$ if $\<\YY \gets \yy\>(\XX = \vv[\XX]) \in C$  for all
finite  $\XX \subseteq \WW$.

%joe10: added paragraph breakx
But there is a more serious problem.
In the proof of Theorem~\ref{theorem:finite-GSEMs}, we showed that
(for the model $M$ constructed from the maximal consistent set $C$),
we have that $(M,\uu) \sat \phi$ iff $\phi \in C$.  But suppose that
%joe10
%$X$ is a variable such that $\R(X) = \{x_1, x_2, x_3, \ldots$ is
%countable, $\R'(X) = \R(X)$, and $\Y gets y \in \I' \inter \I$.
$X$ is a variable such that $\R(X) = \{x_1, x_2, x_3, \ldots\}$ is
countable, $\R'(X) = \R(X)$, and $\Y \gets y \in \I' \inter \I$.
It is not hard to see that the set $C' = \{\<Y \gets y\>true, [Y \gets y](X
	\ne x_1), [Y \gets y](X \ne x_2), \ldots\}$ is consistent with
$AX_{basic}^*$ (since every finite subset of this set is obviously
consistent).  Hence $C'$ can be extended to a maximal consistent set
$C$.  But there is no model $M$ such that $(M,u) \sat \phi$ for all
formulas $\phi \in C'$.  Thus, we have to restrict the set of maximal
consistent sets that we consider.

%spencer18*: Can we call these 'conjunctive events'? I'd like to keep
%a clear distinction between events and formulas.
%joe18: Why not call it a conjunctive formula?  Events are usually
%taken to be sets of worlds (i.e., semantic objects), whereas formulas
%as syntactic objects.
%spencer19: That makes sense. My issue with this is that we're already
%using the term 'formula' for causal formulas [\YY \gets
%\yy]\phi. Here, I would prefer not to call \phi a formula, since it
%isn't in fact a causal formula. Do you think 'event' has confusing
%connotations? It does have the virtue of being consistent with your
%book.
%joe19: Unfortunately, ``event'' does have other connotations,
%although there are certainly papers that conflate events and formulas
%out there.  Here's a suggestion, that I think would result in minimal
%changes to the paper.   I think it's reasonable to call [\YY
%\gets \yy]\phi an atomic causal formula and say that we sometimes
%drop ``causal'' and just say ``atomic formula''.  We can also call
%Boolean combinations of atomic causal formulas causal formulas, and
%again, occasionally drop causal.  We then call X=x a primitive
%(propositional) formula, and a conjunction of primitive formlas can
%be an conjunctive formula.  That way, we reserve ``formula'' for a
%syntactic object.
\dfn
% \begin{restatable}{definition}{conjunctiveFormula}
  \label{definition:conjunctive-formula}
A \emph{conjunctive formula} is a conjunction of formulas of the
form  $X = x$ and $X \ne x$.  (The formula $\true$ is viewed as
conjunctive, since it is an empty conjunction.)
A set $C$ of formulas in $\L_{\WW,\R',\I'}(\S)$
is \emph{acceptable for $\<\YY \gets \yy\>\phi$
	with respect to $Z \in \WW$}, where $\phi$ is a conjunctive formula,
if there is some $z \in \R'(Z)$ and a conjunctive formula $\psi$ such that every
conjunct of $\phi$ is a conjunct of $\psi$, $Z=z$ is a conjunct of
%joe18
%$\psi$, and $\<\YY \gets \yy\>\psi \in C$.  $C$ is \emph{acceptable} if
$\psi$ for some $z \in \R'(Z)$, and $\<\YY \gets \yy\>\psi \in C$.  $C$ is \emph{acceptable} if
$C$ is acceptable for every formula of the form $\<\YY \gets \yy\>\phi \in
	C$ such that $\phi$ is conjunctive formula and $Z \in \WW$.
%spencer18: Note to self: intuitively, this means that any formula in
%C can be extended to another formula in C which contains X = x, so
%we're not allowed to rule out all values of X.
\edfn
% \end{restatable}

%joe10
%The set $C'$ above is not acceptable for $\<Y \gets y>true$ and $X$, and
The set $C'$ above is not acceptable for $\<Y \gets y\>true$ and $X$, and
cannot be extended to an acceptable consistent
set.  Our proof technique involves constructing a model from an
acceptable maximal consistent set.  So we must show that every consistent
formula
%spencer18: typo
% in
%joe26
%
%spencer18*: Would it make more sense to say 'there exists an
%acceptable maximal consistent set'?
%joe18: there does exist an acceptable maximal consistent set, but
%we're making a stronger statement here: every consistent formula is
%contained in some acceptable maximal consistent set.
%spencer19: Oh, ok. Where does this stronger claim get used later in the proof?
%joe19 We certainly use it for the main reult: if \phi is consistent,
%then its satisfiable.  The construction shows that every consistent
%formula is satisfiable in the canonical model that we construct,
%since every consistent formula is a maximal consistent set, and thus
%true in the state on the canonical model corresopnding to that set.
is
included in an acceptable maximal consistent set.
%We actually prove a stronger result, which will be needed for our
%induction hypothesis. Say that a set $C$ of formulas is \emph{safe}
%if, for each variable $X \in \WW$, if $\R'(X)$ is infinite, then there is a
%countable subset $F_X$ of  $\R'(X)$ such
%that none of the values in $F_X$ appear in a formula in $C$.  The
%result we need is summarized in the following lemma.
The next lemma gives the key step for doing this.
%spencer38:
%spencer39:
%Its proof can be found in the appendix.
%joe36: supplementary material
%spencer44: DONE shortv cite; fullv move to main text
Its proof can be found in
%spencer44:
%the supplementary material.
%spencer45:
%Appendix \ref{appendix:additional-proofs}.
\fullv{Appendix \ref{appendix:additional-proofs}.}
\shortv{the full paper.}
\begin{lemma}\label{lem:acceptable} If $C$ is a finite subset of
	$\L_{\WW,\R',\I'}(\S)$
	consistent with $AX^*_{basic,A}(\S,\WW,\R',\I')$,
	$\<\YY \gets \yy\>\phi \in C$, and $X \in
		%joe18
		%  \WW$, then we can add a formula $\psi \in \L_{\WW,\R',\I'}(\S)$ such
		\WW$, then there exists a formula $\psi \in \L_{\WW,\R',\I'}(\S)-C$ such
	that $
		%spencer18:
		% \C
		%$
		C
		\union \{\psi\}$ is consistent with $AX^*_{basic,A}(\S,\WW,\R',\I')$
	and acceptable with respect to $\<\YY \gets \yy\>\phi$ and $X$.
\end{lemma}
We can now prove completeness.   Given a formula $\phi$ consistent
%joe18
%with $AX^*_{basic,A}(\S,\WW,\R',\I')$ we
with $AX^*_{basic,A}(\S,\WW,\R',\I')$, we
construct a maximal acceptable set $C$ consistent with
$AX^*_{basic,A}(\S,\WW,\R',\I')$ containing $\phi$ as follows.    Let
$\sigma_0, \sigma_1, \sigma_2, \ldots$ be an enumeration of the
formulas in $\L_{\WW,\R',\I'}(\S)$ such that $\sigma_0 = \phi$ and let $X_0, X_1, X_2, \ldots $ be
an enumeration of the variables in $\WW$.  It is well known that
there is a bijection $b$ from $\IN$ to $\IN \times \IN$ such that if
$b(n) = (n_1, n_2)$, then $n_1 \le n$.  We construct a sequence of
sets $C_0, C_1,
	C_2, \ldots$ such that $C_0 = \{\phi\}$, $C_k \subseteq C_{k+1}$, and
(a) either $\sigma_k \in C_k$ or $C_k \cup \{\sigma_k\}$ is
inconsistent with $AX^*_{basic,A}(\S,\WW,\R',\I')$, (b) $C_k$ is
consistent, (c) if $b(k) = (k_1,k_2)$ and $\sigma_{k_1}$ has the form
$\<\YY \gets \yy\>\phi$, then $C_{k+1}$ is acceptable with respect to
$\sigma_{k_1}$ and $X_{k_2}$.  We construct the sequence inductively.
%joe10
%Given $C_k$, then we add $\sigma_{k+1}$ to $C_k$ if C_k \union
Given $C_k$, then we add
%spencer18*: I'm pretty sure you meant
% $\sigma_{k_1}$
$\sigma_{k + 1}$
to $C_k$ if $C_k \union
	\{\sigma_{k+1}\}$ is consistent.  In addition, if $\sigma_{k_1}$ has
the form $\<\YY \gets \yy\>\phi$, then we apply
Lemma~\ref{lem:acceptable} to add a formula if necessary to make
$C_{k+1}$ acceptable with respect to $\sigma_{k_1}$ and $X_{k_2}$.

Let $C= \union_{k=0}^\infty C_k$.  Clearly $C$ contains $\phi$.  It is
consistent, since if not, some finite subset of $C$ must be
inconsistent.  But this finite subset must be contained in $C_k$ for
some $k$, and $C_k$ is consistent, by construction.  Finally, $C$ is
%spencer18: typo
% accsptable.
acceptable.
For suppose that $\<\YY \gets \yy\> \phi \in C$ and $X
	\in \WW$.    There must exist $k_1$ and $k_2$ such that
$\<\YY \gets \yy\> \phi = \sigma_{k_1}$ and $X = X_{k_2}$.  Let $k =
	b^{-1}(k_1,k_2)$. Since $\sigma_{k_1} = \<\YY \gets \yy\> \phi \in C$,
it must already be in $C_{k_1}$ (since it would be added in the
construction of $C_{k_1}$ if it was not already in $C_{k_1-1}$). By
the choice of $b$, $k_1 \le k$, so $\sigma_{k_1} \in C_k$.  By
construction, $C_{k+1}$ is acceptable with respect to $\<\YY \gets
	%joe18
	%\yy\> \phi$ and $X_{k_2}$, hence is $C$.
	\yy\> \phi$ and $X_{k_2}$, hence so is $C$.
%$

We now construct a model $M^C$ with signature $\S$.  For
interventions
%spencer7: changed defn above so that \I' subseteq \I
%$\YY \gets \yy in \I \inter \I'$,
%joe10
%$\YY \gets \yy in \I'$,
$\YY \gets \yy \in \I'$,
we take $\FF(\YY \gets \yy, \uu) =
	%spencer18: for consistency
	% \{\vv:$ for all finite subsets $\XX \subseteq \WW$,
	\{\vv \mid$ for all finite subsets $\XX \subseteq \WW$,
%spencer18:
%$\< \yy
$\<\YY
	\gets \yy\>(\XX = \vv[\XX]) \in C\}$.
Now we still need to
%joe10
%define $\F$ on interventions in $\I - \I'$.  If
define $\FF$ on interventions in $\I - \I'$.
Let $\vv^*$ be a fixed assignment. For $I \in \I - \I'$, define
$\FF(I,\uu) = \{\vv^*\}$.
%joe10: removed paragraph break
%
We claim that $M^C \in \G^\A$ and that $\psi \in C$ iff, for all
contexts $\uu$, we have $(M^C,\uu) \sat \psi$.
%spencer18*: added
If these claims hold, then we have produced a model $M^C$ of $\phi$ that satisfies the properties in $\A$, completing the consistency proof.
%spencer39:
%We refer the interested reader to the appendix
%joe36: supplementary material
%spencer44*: Should we move the full proof to the main text in the
%full paper?
%joe38: I'm OK with just moving the sketch
We refer the interested reader to
%spencer45:
%the supplementary material
\shortv{the full paper}
\fullv{Appendix \ref{appendix:additional-proofs}}
for the proofs of these claims. Both proofs use the definition of $M^C$ to relate the formulas in $C$ to the properties of $M^C$. We show the first claim separately for each property in $\A$ given its counterpart axiom in $A$. The second claim is proved by structural induction on $\psi$.
\bbox
}
% DONE remark that this result clarifies the role played by each of Halpern's axioms.

%spencer42*: added. Maybe I even want a corollary spelling out the
%implications for finite GSEMs.
Theorem \ref{theorem:GSEMs} shows that each of the axioms D3, D6$^+$,
D10(a), and D10(b) independently enforces a corresponding property in
GSEMs; namely, coherence, acyclicity, having at most one outcome, and
having at least one outcome. Since in finite GSEMs, D6$^+$ is
equivalent to D6, and acyclicity is equivalent to the usual acyclicity
in SEMs, Theorem \ref{theorem:GSEMs} also implies that each of
Halpern's axioms D3, D6, and D10 independently enforce coherence,
acyclicity, and unique outcomes in SEMs.
%joe37*: I think that we can cut this for AAAI; although I would
%reinstate it for the full paper
%spencer44: That makes sense.
\fullv{The result that D6 enforces
acyclicity in SEMs in the presence of the other axioms was shown in
\cite{Hal20}, and the fact that D10 enforces unique solutions in SEMs
was essentially known. However, the fact that the axioms have
independent effects and the characterization of the effect of D3 are
both novel.}
%joe36: it doesn't make sense to talk about individual axioms being
%complete.  I would just cut this sentence (saying something accurate
%here would largely repeat the theorem statement).
%Moreover, Theorem \ref{theorem:finite-GSEMs} shows that
%Halpern's other axioms (e.g., $AX^+_{basic}$) are complete only for
%the class of all finite GSEMs.
%spencer43*: I want to say  that the axiom system consisting of the
%remaining axioms (without D3, D6, and D10) is an axiomatization for
%the class of all finite GSEMs (and thus, don't imply any statements
%that aren't satisfied by all finite GSEMs, which is a very general
%class).
%joe37: I realize that; that's Theorem 5.2.  I'm not sure it's worth
%repeating (but I could be talked into it).
%spencer44: I think what we say before is probably enough.
In  \cite{PH20}, we make the case that
GSEMs are the most general class of causal models that have the same
input and output as SEMs (and satisfy
%spencer44: shaving
%the basic sanity check of
effectiveness). Putting the pieces together gives a full picture of
how each of Halpern's original axioms relates to the properties of
SEMs. The axioms of $AX^+_{basic}$ are just enough to prove statements
that hold in \emph{all} causal models with the same input and output
%joe37
%as SEMs (and satisfying effectiveness). Then, each of the remaining
as SEMs (and satisfying effectiveness). Each of the remaining
axioms simply independently enforces a natural property of SEMs.
%joe36: I don't mind leaving this in, but it feels somewhat repetitive
%spencer43: I agree, this is repetitive, cutting.
% D3 enforces coherence, D6 enforces acyclicity, and (breaking D10 into two
% natural components), D10(a) enforces having at least one solution
% while D10(b) enforces having at most one solution.
%joe36
%We feel this
This
may be of interest completely independently of GSEMs and their
applications.
\newpage
\section{Acknowledgments}
Work supported in part by
NSF grants IIS-178108 and IIS-1703846, a grant from the Open
Philanthropy Foundation, ARO grant W911NF-17-1-0592, and MURI grant
W911NF-19-1-0217.

%joe39
%\bibliography{joe}
\shortv{\bibliography{joe}}

%spencer21: added for proofs
\appendix
%spencer28*: Added so I can split the pdf into paper and supplementary material.
\clearpage
\fullv{
\section{Proofs}
%spencer44:
\label{appendix:additional-proofs}
%spencer21: moved macros from GSEM paper
\newenvironment{oldthm}[1]{\par\noindent{\bf Theorem #1:} \em \noindent}{\par}
\newenvironment{oldlem}[1]{\par\noindent{\bf Lemma #1:}
	\em \noindent}{\par}
\newenvironment{oldcor}[1]{\par\noindent{\bf Corollary #1:} \em \noindent}{\par}
\newenvironment{oldpro}[1]{\par\noindent{\bf Proposition #1:} \em \noindent}{\par}
\newenvironment{olddefn}[1]{\par\noindent{\bf Definition #1:} \em \noindent}{\par}
\newcommand{\othm}[1]{\begin{oldthm}{\ref{#1}}}
		\newcommand{\eothm}{\end{oldthm} \medskip}
\newcommand{\olem}[1]{\begin{oldlem}{\ref{#1}}}
		\newcommand{\eolem}{\end{oldlem} \medskip}
\newcommand{\ocor}[1]{\begin{oldcor}{\ref{#1}}}
		\newcommand{\eocor}{\end{oldcor} \medskip}
\newcommand{\opro}[1]{\begin{oldpro}{\ref{#1}}}
  \newcommand{\eopro}{\end{oldpro} \medskip}
\newcommand{\odefn}[1]{\begin{olddefn}{\ref{#1}}}
  \newcommand{\eodefn}{\end{olddefn} \medskip}

\opro{proposition:acyc-conditions} If $M$ is a SEM, then $M$ satisfies
  %spencer35:
  %(\ref{eq:2}) iff $M$ satisfies (\ref{eq:3}).
%joe31
  %  Condition \ref{condition:stronger} iff $M$ satisfies Condition
%\ref{condition:weaker} (for a fixed context $\uu$).
 Acyc1 iff $M$ satisfies
Acyc2 (for a fixed context $\uu$).
\eopro
%spencer38*: added proof
\prf
%spencer39:
%We need only prove that Acyc2 implies Acyc1 in SEMs.
We need only prove that Acyc2 implies Acyc1 in SEMs, since the other direction is obvious.
Fix a context $\uu$. We first claim that if $M$ satisfies Acyc2, then $M$ has unique solutions. Fix an intervention $I$; it suffices to show that for all variables $X$ there is $x$ such that for all solutions $\vv \in M(\uu, I)$, $\vv[X] = x$. We show this by induction on $\prec_\uu$. For the base case, let $X$ be the minimum variable with respect to $\preceq_\uu$. Acyc2 implies that the structural equation for $X$ does not depend on any variables; that is, it is a constant $\F_X = x$. Hence $\vv[X] = x$ for all solutions $\vv \in \M(\uu, I)$. For the inductive step, Acyc2 implies that the structural equation for $Z$ does not depend on variables $Y \succeq_\uu Z$. Hence $\F_Z$ only depends on variables preceding $Z$. But by inductive hypothesis, those variables have the same values in all solutions. Plugging those values into $\F_Z$, we find that $\F_Z$ is also a constant.
Now that we have shown that $M$ has unique solutions, we can finish the proof by observing that Acyc2 states that the unique outcome of $I; X \gets x$ and the unique outcome of $I; X \gets x'$ agree on all variables $Y \prec_\uu X$. But this implies Acyc1.
\eprf

\othm{theorem:finite-GSEMs}
$AX^+_{basic}$ is sound and complete for finite GSEMs.
\eothm
\prf
First we prove soundness. Fix a finite GSEM $M$ and context $\uu$.
D0 is trivially sound. D1 is sound since for all $\vv \in M(\uu, \YY
	%joe6: added some words to make it easier to parse
	%\gets \yy)$, $\vv[X] = x \Rightarrow \vv[X] \neq x'$ for $x \neq
	%x$. D2 is sound since for all $\vv \in M(\uu, \YY \gets \yy$, $\vv[X]
	\gets \yy)$, we have that $\vv[X] = x \Rightarrow \vv[X] \neq x'$ for $x \neq
	x'$. D2 is sound since for all $\vv \in M(\uu, \YY \gets \yy)$, we
have that $\vv[X]
	\in \R(X)$. D4 is sound by definition of GSEMs. D7, D8 and MP are
trivially sound.
%joe6: added paagraph break.

Next we prove completeness.
%spencer7*: Now we only use F1 (in its if and only if form)
%The proof uses the fact that various
%formulas (some of which are already mentioned in Footnote 1)
%are derivable from the axioms using well-known techniques of
%modal logic, in particular:
%\begin{itemize}
%\item[F1.] $[\YY \gets \yy]\phi_1
%    \land [\YY \gets \yy]\phi_2 \rimp [\YY \gets \yy](\phi_1 \land
%    \phi_2)$ and its dual
%  $\<\YY \gets \yy\>(\phi_1 \lor \phi_2) \rimp \<\YY \gets
%    \yy\>\phi_1 \lor \<\YY \gets \yy\>\phi_2$;
%    \item[F2.]  $[\YY \gets \yy]\phi_1 \land \<\YY \gets \yy\>\phi_2
%      \rimp \<\YY \gets \yy\>(\phi_1 \land \phi_2)$;
%\item[F3.]        $\<\YY \gets \yy\>(\phi_1 \wedge \phi_2) \Rightarrow
%  \<\YY \gets \yy\> \phi_1$.
%\end{itemize}
%spencer8: moved below
%The proof uses the following formulas, which are derivable from the
%axioms using well-known techniques of modal logic.
%\begin{description}
%    \item[F1.] $[\YY \gets \yy]\phi_1
%    \land [\YY \gets \yy]\phi_2 \Leftrightarrow [\YY \gets \yy](\phi_1 \land
%    \phi_2)$ and its dual
%  $\<\YY \gets \yy\>(\phi_1 \lor \phi_2) \Leftrightarrow \<\YY \gets
%    \yy\>\phi_1 \lor \<\YY \gets \yy\>\phi_2$.
%    \item[F2.] $[\YY \gets \yy]\phi_1
%    \land \<\YY \gets \yy\>\phi_2 \Leftrightarrow \<\YY \gets \yy\>(\phi_1 \land
%    \phi_2)$
%\end{description}
%joe7: this needs to be corrected if you make this a lemma someplace.
%spencer8: check!
%(Note that F1 and a special case of F2 already appeared in Footnote 1.)
Fix a finite signature $\S$. It suffices
to show that for any causal formula $\varphi \in \L(\S)$ that is
consistent with $AX^+_{basic}(\S)$, there is a finite GSEM $M$ and
context $\uu$ such that $(M, \uu) \models \varphi$.

%spencer13: do we need to explain this technique?
%joe11: We now do explain it, and if space is not an issue, I'd prefer
%to leave in the explanation.  If we're short of space, let's discuss
%where to make cuts.  I' not sure that this is the first place I'd choose.
%spencer14: No, I mean the 'it suffices' claim'. (Unless citing Hal20
%is enough explanation.)
%joe12: My guess is that anyone reading this will know basic modal
%logic, so will accept this.  Otherwise, we should define ``consistent
%with AX'' as well.  All this may not be a bad thing to do in the full paper.

%spencer15: Makes sense.
%joe6
%Let us prove this. Suppose that $\varphi \in \L(\S)$ is consistent
To prove this, we use essentially the same technique as that used by
Halpern \nciteyear{Hal20}.  We briefly sketch the details here:
Suppose that $\varphi \in \L(\S)$ is \emph{consistent
with $AX^+_{basic}(\S)$}
%joe6
(i.e., we cannot prove $\neg \phi$ in $AX^+_{basic}(\S)$).
%joe6: don't need to talk about the axioms
%Then (by definition) the set of causal
%formulas $S$ containing just $\varphi$
%and $AX^+_{basic}(\S)$ is
%consistent. Extend this set to a maximal consistent set $C \supseteq
%joe6: this is not the definition of maximal consistent set
%S$ (i.e. for every causal formula in $\L(\S)$ this set contains either
%this formula or its negation).  Then $\phi$ can be extended to a
%maximal consistent set $C$ of
%spencer13: I think the statement above is not supposed to be commented, adding
Then  $\phi$ can be extended to a maximal consistent set $C$ of
formulas; that is, $\phi \in C$, every finite subset $C'$ of $C$ is
consistent with $AX^+_{basic}(\S)$ (i.e., the conjunction of the
formulas in $C'$ is consistent with $AX^+_{basic}(\S)$), and no
strict superset $C^*$ of $C$ has the property that every finite subset
of $C^*$ is consistent with $AX^+_{basic}(\S)$.  Standard arguments
show that, for every formula $\psi \in \L(\S)$, either $\psi$ or
$\neg \psi$ must be in $C$; moreover, every instance of the axioms in
%spencer13: Added more facts about C.
%% $AX^+_{basic}(\S)$ is in $C$
$AX^+_{basic}(\S)$ is in $C$.
%joe11*: there is a subtlety about what ``closure under provability''
%means if you have rules of inference like D2.  Once you have all
%axioms in C (which we already said), the only property you need is
%closure under implication: if \phi and \phi -> \psi are both in C,
%then so is \psi.  We could add this if you want, although it follows
%easily from the maximal consistency
%%Finally, $C$ is closed under provability--if a formula $\psi$ is
%provable from formulas in $C$, then $\psi \in C$.
%this formula or its negation).
%spencer14: Makes sense. I agree that all we need is closure under
%implication, and that it's not worth stating here.
Define a finite GSEM $M$ with signature $\S$ as follows. For all
contexts $\uu$ and allowed interventions $\YY \gets \yy \in \I$,
define
$\FF(\uu, \YY \gets \yy) = \{\vv \in \R(\V) \mid \<\YY \gets \yy\>(\V =
	\vv) \in C\}$.
%joe6: added to help the reader
Thus, the formulas in $C$ tell us how interventions work in $M$.
%spencer13*: added, refactoring.
%joe18
%Fix an arbitrary context $\uu$. To finish the proof, we need to show
Fix a context $\uu$. To finish the proof, we need to show
two properties of $M$: first, that $M$ is in fact a GSEM (that is,
it satisfies effectiveness) and second, that $(M, \uu) \models
	\varphi$.
%joe11
%We will prove instead the following stronger claim.
These both follow from a standard \emph{truth lemma}:
\begin{lemma}\label{lemma:c-characterizes-m}
	For all $\psi \in \L(\S)$,
	\begin{equation} \label{psi-in-c-equiv-m-models-psi}
		%joe18: I prefer to reserve <=>  for formulas
		%  \psi \in C \Leftrightarrow (M, \uu) \models \psi.
		\psi \in C \mbox{ iff } (M, \uu) \models \psi.
	\end{equation}
\end{lemma}
%joe11: I found the parenthetical comment confusing.  I don't think it helpss.
%That is, $C$ contains exactly the formulas true of $M$ (in any
%context; the solution sets $M(\uu, \YY \gets \yy)$ of $M$ are
%independent of context).
%spencer14: I agree.
That is, $C$ contains exactly the formulas true of $M$.
%joe26
%First, effectiveness follows from Lemma~\ref{lemma:c-characterizes-m}
Effectiveness follows from Lemma~\ref{lemma:c-characterizes-m}
%joe11: it seems strange to say that effectiveness follows .. because
%C contains all instances of effectiveness.  It also seems like overkill.
%because $C$ contains all instances of effectiveness (D4), thus $M$
%satisfies all instances of effectiveness (so by definition, $M$
%satisfies effectiveness).
%spencer14: I also agree here.
because $C$ contains all instances of D4.
%joe26
%Second, the desired result $(M, \uu) \models \varphi$) follows trivially from
The desired result $(M, \uu)
	\models \varphi$) also follows easily from
%spencer18: typo
Lemma~\ref{lemma:c-characterizes-m} using the fact $\varphi \in
	%spencer18: typo
	% \C
	C
$.

To prove Lemma~\ref{lemma:c-characterizes-m}, the key step is
to prove
(\ref{psi-in-c-equiv-m-models-psi})
for the basic causal formulas
%joe14
%$[\<\YY \gets \yy] \rho$.
$[\YY \gets \yy] \rho$.
Since $C$ is maximal,
%spencer15:
% and every causal formula can be written as a Boolean combination of
% dual basic causal formulas,
%joe14
%and causal formulas are by definition Boolean combinations of basic
and causal formulas are, by definition, Boolean combinations of basic
causal formulas,
the result then follows by a
straightforward structural induction on $\psi$.

%joe14
%It further suffices to check (\ref{psi-in-c-equiv-m-models-psi})
%for formulas of the form $\psi = [\YY \gets \yy](\band_{\vv \models
%$
It suffices to check (\ref{psi-in-c-equiv-m-models-psi})
for basic causal formulas of the form $[\YY \gets \yy](\band_{\vv \models
		\neg \rho} \V \neq \vv)$,
since for all events $\rho$,
\begin{equation} \label{eq:equivalence-improved}
	AX_{basic}^+ \vdash [\YY \gets \yy]\rho \dimp [\YY \gets
		%joe13
		%    \yy](\band_{\vv \models \neg \rho} \V \neq \vv).
		\yy]\band_{\vv \models \neg \rho} \neg( \V = \vv),
\end{equation}
%joe14
where, as usual, we write $ AX \vdash \phi$ to denote that
$\phi$ is provable in the axiom system AX.
% Wait, is this semantically correct? Recall that [\YY \gets \yy]\rho
% means that all solutions under intervention \YY \gets \yy satisfy
% rho. A solution either satisfies \rho, or it doesn't. So another way
% of saying this is that none of the solutions under \YY \gets \yy are
% solutions which don't satisfy \rho. Rephrasing one more time, all of
% the solutions under \YY \gets \yy are not equal to any of the
% solutions that don't satisfy rho. Yeah, so this makes sense.
%joe13: removed paragraph break
%
%joe26
%Viewing $\rho$ as a predicate on solutions,
Viewing the formula $\rho$ as a predicate on outcomes,
(\ref{eq:equivalence-improved}) expresses the intuitively obvious
claim that $\rho$ is equivalent to the predicate that checks that its
%joe26
%input is not equal to any solution not satisfying $\rho$.
input is not equal to any outcome not satisfying $\rho$.

%joe13
%To prove this, we need one lemma, which we state here without proof,
To prove (\ref{eq:equivalence-improved}), we need one lemma, which we
state here without proof,
but which follows from just D0, D7 and D8 using standard modal logic
arguments.
\begin{lemma}
	\label{lemma:and-distrib-box}
  %joe26
  %joe33:
        %\begin{itemize}
\begin{itemize}[leftmargin=\parindent + 5pt, align=left,
  labelwidth=\parindent, labelsep=5pt, itemsep=2pt]
\item[(a)]
	$AX_{basic}^+ \vdash [\YY \gets \yy]\phi_1
		\land [\YY \gets \yy]\phi_2 \dimp [\YY \gets \yy](\phi_1 \land
		\phi_2)$ \\
	%spencer14: we don't need this.
	%joe12: we don't need it here, but we do need it for the
        %earlier argument.
%joe26
%	        and its dual
    % DONE change refs to these lemmas to (a) and (b)
                \item[(b)]
	%spencer15: OK, putting it back.
	$AX_{basic}^+ \vdash \<\YY \gets \yy\>(\phi_1 \lor \phi_2) \dimp_{AX_{basic}^+}  \<\YY \gets
		\yy\>\phi_1 \lor \<\YY \gets \yy\>\phi_2$.
                %joe26
                \end{itemize}
        \end{lemma}

Let $\phi_{D1} = (\band_{X \in \V, x \in \R(X), x' \neq x} (X = x
	\rimp X \neq x')$ and $\phi_{D2} = (\bor_{X \in \V, x \in \R(X)} (X =
	x))$. These formulas are the conjunctions of the formulas after the
$[\YY \gets \yy]$ in D1 and D2 respectively. Let $\phi_{EQ} =
	\phi_{D1} \wedge \phi_{D2}$.
%joe13: added
$\phi_{EQ}$ says that, for each variable $X \in \V$, there is exactly
one value $x \in \R(X)$ such that $X = x$.

% denote the conjunction of the formulas after the $[\YY \gets \yy]$ in D1 and D2.
It follows from D1, D2, and Lemma~\ref{lemma:and-distrib-box} that
%joe14
%$[\YY \gets \yy]\phi_{EQ}$.
%spencer17: typo
% $AX^+_{basic} [\YY \gets \yy]\phi_{EQ}$.
$AX^+_{basic} \vdash [\YY \gets \yy]\phi_{EQ}$.
% By D1, D2, and Lemma~(\ref{lemma:and-distrib-box}),
% $$AX_{basic}^+ \vdash [\YY \gets \yy] \rho \dimp [\YY \gets \yy](\rho \wedge \phi_{EQ})$$.
% Suppose that
It is straightforward to show that $\phi_{EQ} \rimp (\rho \dimp
	\band_{\vv \models \neg \rho} \V \neq \vv)$ is a propositional
tautology.
%joe14: I'm not sure this is necessary, but in any case, it's not
%quite right.  Specifically, it's not true that \V\ne v' -> \V = v
%(so it's not true that \V-v <=> \V \ne v')
%spencer15: Oops, you're right. Ok, let's leave this out for now.
%We sketch a brief proof below. First observe
%$\phi_{D1} \dimp (\band_{\vv \neq \vv'} (\V = \vv \dimp \V \neq
%\vv'))$ is a propositional tautology.
%So is $\phi_{D2} \dimp (\bor_{\vv \in \R(V)} \V = \vv)$.
%Hence $\phi_{D2} \wedge \rho \rimp \bor_{\vv \models \rho} \V = \vv$.
%But then $\phi_{D1} \wedge \phi_{D2} \wedge \rho \rimp \band_{\vv
%\models \neg \rho} \V \neq \vv$.
%joe13: a little more detai
%Thus it follows from D8 that $AX_{basic}^+ \vdash [\YY \gets \yy]
%$
%joe14
By D8, we have that $AX_{basic}^+ \vdash [\YY \gets \yy]
	(\phi_{EQ} \rimp (\rho \dimp \band_{\vv \models \neg \rho} \V \neq
	\vv))$. Hence, by D7, $AX_{basic}^+ \vdash [\YY \gets \yy](\rho \dimp
	\band_{\vv \models \neg \rho} \V \neq \vv)$.
%joe13
%But then from D7 it easily follows  $AX_{basic}^+ \vdash [\YY \gets
%$
Using D7 again, it follows that $AX_{basic}^+ \vdash [\YY \gets
		\yy]\rho \dimp [\YY
		\gets \yy]\band_{\vv \models \neg \rho} \V \neq \vv)$, which is
exactly \eqref{eq:equivalence-improved}.

%joe13
%Thus it suffices to show \eqref{psi-in-c-equiv-m-models-psi} for
%$\psi$ of the form $\psi = [\YY \gets \yy](\band_{\vv \models \neg
%$
%joe14
%It remains to show \eqref{psi-in-c-equiv-m-models-psi} for $\psi$
%of the form $[\YY \gets \yy](\band_{\vv \models \neg
%  \rho} \V \neq \vv)$.
It remains to prove \eqref{psi-in-c-equiv-m-models-psi} for formulas
of the form \mbox{$[\YY \gets \yy](\band_{\vv \models \neg
			\rho} \V \neq \vv)$}.
%joe13:
%But since $C$ is maximal and consistent with $AX_{basic}^+$,
Since $C$ is maximal and consistent with $AX_{basic}^+$,
%joe13
%Lemma~(\ref{lemma:and-distrib-box}) implies that $[\YY \gets
%joe14
%Lemma~\ref{lemma:and-distrib-box} implies that $[\YY \gets
% \yy](\band_{\vv \models \neg \rho} \V \neq \vv) \in C$ if and only

%$
it follows from Lemma~\ref{lemma:and-distrib-box} that \mbox{$[\YY \gets
				\yy](\band_{\vv \models \neg \rho} \V \neq \vv) \in C$} if and only
if for all $\vv \models \neg \rho$, $[\YY \gets \yy](\V \neq \vv) \in
	%joe14
	%C$. Clearly we also have $(M, \uu) \models [\YY \gets \yy](\band_{\vv
	C$. Clearly, we also have that $(M, \uu) \models [\YY \gets \yy](\band_{\vv
		\models \neg \rho} \V \neq \vv)$ if and only if for all $\vv \models
	\neg \rho$, $(M, \uu) \models [\YY \gets \yy](\V \neq \vv)$. Finally,
%joe14
%by definition, $\neg [\YY \gets \yy](\V \neq \vv) \dimp \<\YY \gets
%$
by definition, $\neg [\YY \gets \yy](\V \neq \vv)$ is $\<\YY \gets
	\yy\>(\V = \vv)$.
%joe14
%Therefore, it suffices to show \eqref{psi-in-c-equiv-m-models-psi} for
%$\psi$ of the form $\<\YY \gets \yy\>(\V = \vv)$.
Therefore, it suffices to prove \eqref{psi-in-c-equiv-m-models-psi} for
formulas $\psi$ of the form $\<\YY \gets \yy\>(\V = \vv)$.

%joe13
%We now show this, completing the proof. Suppose $\psi \in C$. Then by
%joe14: just minor changes
%We now show this, completing the proof. Suppose that $\psi \in C$. Then by
%definition of $M$, $\vv \in \F(\uu, \YY \gets \yy)$, so $(M, \uu)
%\models \psi$. Conversely, suppose $(M, \uu) \models \psi$. Then $\vv
%\in \F(\uu, \YY \gets \yy)$. Again by definition of $M$, this implies
Suppose that $\psi \in C$. Then, by
definition of $M$, we have that $\vv \in \FF(\uu, \YY \gets \yy)$, so $(M, \uu)
	\models \psi$. Conversely, suppose that $(M, \uu) \models \psi$. Then $\vv
	\in \FF(\uu, \YY \gets \yy)$. Again, by definition of $M$, this
implies that
$\psi \in C$.{}
%joe13
\eprf

\othm{theorem:acyclic-finite-GSEMs}
  $AX^+_{basic, rec}$ is sound and complete for acyclic finite GSEMs.
\eothm
%spencer38*:
\prf
%spencer39:
%Consider the version of D6$^+$ (\ref{eq:d6-finite}), which is
%spencer40: Moving from above, since we cut this discussion.
% Consider (\ref{eq:d6-finite}), which is
% %spencer39:
% %equivalent over finite signatures.
% equivalent to D6$^+$ over finite signatures.
Over finite signatures, it is straightforward to verify that D6$^+$ is
equivalent to
%joe30
%\begin{itemize}
% \begin{itemize}[leftmargin=\parindent + 5pt, align=left, labelwidth=\parindent, labelsep=5pt, itemsep=2pt]
% \item[D6$^+$ (finite signatures)]
%spencer38: put in eqn for reference later
\begin{equation}
  %joe34*: why is it obvious that you can get rid of the intervention
  %I?  We use it in D6
  \label{eq:d6-finite}
  \bor_{i = 1}^k \neg (X_i \affects
  \{X_1, \dots, X_{i - 1}, X_{i+1}, \dots, X_k\})
\end{equation}
% \end{itemize}
where $X \affects \YY$ abbreviates the formula
$\bor_{x, x' \in \R(X), \yy \in \R(\YY), I \in \I}
%spencer35: clearer
%\<I; X \gets x\>(\YY = \yy) \wedge [I; X \gets x'](\YY \neq \yy)$.$
X \affects_{x, x', \yy, I} \YY$.
Soundness is clear; given an acyclic GSEM $M$ and context $\uu$, for all instances of D6$^+$, the definition of acyclicity implies that the disjunct corresponding to the largest element of $\{X_1, \dots, X_k\}$
%spencer39:
%in $\prec_\uu$ is true.
with respect to $\prec_\uu$ is satisfied by $M$.
For completeness, we build off the proof of Theorem \ref{theorem:finite-GSEMs}. It suffices to show that the model $M$ constructed in that proof is acyclic. $M$ has only one context $\uu$, so it suffices to show the acyclicity condition Acyc1 holds for that $\uu$. Lemma \ref{lemma:c-characterizes-m} implies that $M$ satisfies all instances of D6$^+$ in context $\uu$. Hence, for every instance $\phi$ of D6$^+$, $M$ satisfies at least one disjunct $1 \leq i \leq k$ of $\phi$ (in context $\uu$).  We use this to construct an order $\prec_\uu$ over $\V$ as follows. Consider the unique instance $\phi_k$ of D6$^+$ where $\{X_1, \dots, X_k\} = \V$. Let $i$ be any disjunct of $\phi_k$ satisfied by $M$, and
%spencer39:
%let $X_i$ be the maximum element of $\prec_\uu$.
define the maximum element of $\prec_\uu$ to be $X_i$.
Next consider the unique instance $\phi_{k-1}$ of D6$^+$ where $\{X_1, \dots, X_k\} = \V - \{X_i\}$. Let $j$ be any disjunct of $\phi_{k-1}$ satisfied by $M$, and
%spencer39:
%let $X_j$ be the second largest element of $\prec_\uu$.
define the second largest element of $\prec_\uu$ to be $X_j$. Repeating this process gives an order $\prec_\uu$ of all the variables, such that for all $X \in \V$, $M$ satisfies $\neg (X \affects \V_{\prec_\uu X})$. But this implies that $M$ satisfies Acyc1.
\eprf

%spencer23*: Moved from main text
\ocor{corollary:finite-GSEMs-satisfying-axioms}
%joe6
%Fix a signature $\S$ such that $\I = \I_{univ}$. $AX^+(\S)$ is
%complete for the class of finite GSEMs satisfying $AX^+(\S)$.
%\end{theorem}
If $\S$ is a universal signature, then $AX^+(\S)$ is a sound and
complete axiomatization for $\L(\S)$ for finite GSEMs satisfying
$AX^+(\S)$.
\eocor
\prf
Since for every finite GSEM satisfying $AX^+(\S)$ there is an
%spencer7: even though they're equivalent, I feel like the notion
%directly defined in terms of causal formulas is better.
%equivalent SEM
$\S$-equivalent SEM
and vice versa, the same set of causal formulas is
valid in each class of models.
%joe6
%Since all formulas valid in SEMs are
%provable in $AX^+(\S)$, it follows that every causal formula valid in
%GSEMs is provable in $AX^+(\S)$.
Since, as shown by Halpern~\nciteyear{Hal20}, $AX^+(\S)$ is a sound and
complete axiomatization of SEMs, the result follows.
\eprf

\othm{theorem:axioms-not-complete-for-SEMs}
$AX^+(\S)$ is not complete for SEMs of signature $\S$
when some interventions are not allowed in $\S$, although it is
sound.
\eothm

\prf
For incompleteness, consider the causal formula $\phi$ that
%joe26
%characterizes the shell game and its solutions, that is,
characterizes the shell game and its outcomes, that is,
$\phi = [S_1 \gets 1](S_1 = 1\land S_2 = 1 \land Z = 1) \wedge [S_2 \gets
		1](S_1 = 1 \land S_2 = 1 \land Z = 0)$.
This formula is false in all SEMs with signature $\S_{shell}$ (the
signature of $M_{shell}$). Hence, $\neg
	\phi$ is true in all SEMs with signature $\S_{shell}$. However, $\neg
	\phi$ is not provable from $AX^+(\S_{shell})$, because $M_{shell}$
satisfies all axioms of $AX^+(\S_{shell})$, but $\phi$ is true in
$M_{shell}$.
%spencer42*: Reading this now, it seems a little odd.
% I think I should say that for all signatures \S, \AX^+(\S) is always sound for SEMs over signatures \S, because it is strictly weaker than \AX^+(\S'), where
% \S' is \S but with all interventions allowed, and \AX^+(\S') is
% sound for SEMs over signatures \S. Do you agree?
%joe36: I'm actually OK with what's written.
%spencer43*: But I'm confused. Why does this observation imply AX^+(S)
%is sound for SEMs of signature S? Or, is that not what the theorem is
%stating?
%joe37*: As we say in Apprenfix D.1, we observed in Example 3.6 of the
%GSEM paper, all the axioms in AX^+(S_{shell}) are valid in
%M_{shell}.  But we can't say "as we observed", because now we're
%observing it later.
%spencer44: I'm still confused, but I'm not very worried about
%it. I'll just rewrite this.
%For soundness, as we observed, all the axioms of $AX^+(\S_{shell})$
For soundness, as shown in \cite{PH20}, all the axioms of $AX^+(\S_{shell})$
are valid in $M_{shell}$.
\eprf

\opro{prop:d2-plus}
%joe14
%  If we replace D2 by D2$^+$ in $AX_{basic}$, then we can
%derive D2 for variables with finite domains. Moreover,
If $AX_{basic}^*$ is the result of replacing D2 with D2$^+$ in
$AX_{basic}^+$, then we can
derive D2 for variables with finite domains in $AX_{basic}^*$. Moreover,
%joe7
%D2$^+$ is derivable in $AX_{basic}$, in the sense that if
%joe14
%D2$^+$ is derivable in $AX_{basic}$, in the sense that
%jo26
%D2$^+$ is derivable in $AX_{basic}^+$, in the sense that
D2$^+$ is derivable in $AX_{basic}^+$ for variables $X$ with finite range, in the sense that
%spencer15: updated D2^+
% if $\phi \rimp \land_{i=1}^{S}$
%joe14: might as well use notation
%if $\phi \rimp \land_{x \in S}
if $AX_{basic}^+ \vdash \phi \rimp \land_{x \in S}
	%spencer15: I think this is a typo
	% \<\YY \gets \yy\>
	[\YY \gets \yy]
	(\psi
	%spencer15: updated D2^+
	% $\rimp (X \ne x_i))$ is provable in $AX_{basic}$
	%joe14
	%\rimp (X \ne x))$ is provable in $AX_{basic}$
	\rimp (X \ne x))$
and $\R(X)$ is finite,
%spencer16: typo
% then so is $\phi \rimp [\YY \gets \yy]\neg \phi$.
%joe14
%then so is $\phi \rimp [\YY \gets \yy]\neg \psi$.
then  $AX_{basic}^+ \vdash \phi \rimp [\YY \gets \yy]\neg \psi$.
\eopro
% joe7*: We still need to do the proof here ..
\prf
%joe14
%To prove the first claim, assume $\R(X)$ is finite, and define $S =
%$
To prove the first claim, suppose that $\R(X)$ is finite. Define $S =
	\R(X)$, $\phi = true$, and $\psi = \band_{x \in S} X \neq x$.
%joe14
%
%It is obvious that $AX_{basic} \vdash \phi \rimp [\YY \gets \yy](\psi
%\rimp \band{x \in S} X \neq x)$.
%Applying $D2^+$ gives
Clearly, $AX_{basic}^+ \vdash \phi \rimp [\YY \gets \yy](\psi
	\rimp \band_{x \in S} X \neq x)$,
so
%joe14
%$AX_{basic} \cup \{D2^+\} \vdash \phi \rimp [\YY \gets \yy](\neg \psi)$.
$AX_{basic}^* \vdash \phi \rimp [\YY \gets \yy](\neg \psi)$.
%joe14: not quite
%But this formula is propositionally equivalent to $D2$, namely $[\YY
%\gets \yy](\bor_{x \in \R(X)} X = x)$.
%joe16
%But $\neg \psi$ is equivalent to $\bor_{x \in \R(X)} X = x)$, so using
But $\neg \psi$ is equivalent to $\bor_{x \in \R(X)} X = x$, so using
D7 and D8, we can conclude that $AX_{basic}^* \vdash [\YY \gets \yy]
	(\bor_{x \in \R(X)} X = x)$, as desired.
% The conclusion of D2$^+$ is $\phi $

%joe14: added paragraph break
%To prove the second claim, assume that $\phi \rimp
%\land_{i=1}^{x_{k+1}} [\YY \gets \yy](\psi \rimp (X \ne x_i))$ is
%provable in $AX_{basic}^+$. Rewriting this assumption, there exists a
For the second claim, suppose that $AX_{basic}^+ \vdash \phi \rimp
	%joe17: added a sentence, which will come in handy later, and helps
	%make clear what $S$ is
	%\land_{x \in S} [\YY \gets \yy](\psi \rimp (X \ne x))$.$
	\land_{x \in S} [\YY \gets \yy](\psi \rimp (X \ne x))$, where $S
	\subseteq \R'(X)$ and $S$ contains some value in $\R'(X)$ not in
$\phi \rimp [\YY \gets \yy]\psi$ if there is one.
%joe14*: Spencer, given the change in D2^+, this isn't quite right.
%We need to consider two cases: First S = \R(X).  Then you can use D2
%to get the conclusion.  Specifically, D2 gives you |- [...]\/_{x \in
%\R(X) X = x.  Now just use Lemma 5.4, D7, D8, and the observation
%that \land_{x \in \R(X)} (\psi = X \ne x) and \/_{x_ \in \R(X) X=x
%implies \neg \psi.
%Second, S \ne \R(X).  In that case, you can
%use your argument (except you should get rid of x_1, ..., x_k.  Just
%let x be the element of S that doesn't apear anywhere else, and show
%that you can replace it by any other element of \R(X).  Then you can
%finis off the proof as above, using D2, D7, D8, and Lemma 5.4.
%spencer17*: Adding the S = \R(X) case.
%
First we consider the case $S = \R(X)$. Applying
Lemma~\ref{lemma:and-distrib-box}, we have
$AX_{basic}^+ \shows \phi \rimp [\YY \gets \yy](\band_{x \in \R(X)} \psi \rimp (X \neq x))$.
%spencer17: moving from below.
It is easy to see that $\band_{x \in \R(X)} (
	%spencer17: typo
	% \phi
	\psi
	\rimp (X \ne x))$ is logically
%spencer17*: They are equivalent--if a -> AND_b b, then AND_b a -> b as well.
% implies
equivalent
%spencer17: typos
to $\psi \rimp \band_{x \in \R(X)} (X \ne x)$, which is
logically equivalent to $\lor_{x \in \R(X)} (X =x) \rimp \neg \psi$.
Thus, using D7 and D8, we get
$$
	AX_{basic}^+ \vdash \phi \rimp [\YY \gets \yy]( \lor_{x \in \R(X)}
	(X = x) \rimp \neg \psi).   $$
Now using D2, D7, and D8, we easily get
$$
	AX_{basic}^+ \vdash \phi \rimp [\YY \gets \yy]\neg \psi,$$
as desired.

Next we consider the case $S \neq \R(X)$.
Since we assumed that
$\R(X)$ is finite, $\R'(X) = \R(X)$. Hence there is a value
%joe17
%$x \in \R'(X)$ that is not mentioned in $\phi \rimp [\YY \gets \yy]\psi$.
%Suppose without loss of generality that $x \in S$.
in $\R'(X)$ that is not mentioned in $\phi \rimp [\YY \gets \yy]\psi$, so
there must be such an element in $S$, say $x$.
%spencer18: Thanks! I think this clarification was needed.
%joe17: removed paragraph break
%
%spencer17*: getting rid of X_1, \dots, X_k.
% Rewriting this assumption, there exists a
% proof of
% $\phi \rimp \land_{i=1}^{k} [\YY \gets \yy](\psi \rimp (X \ne x_i))
% \wedge [\YY \gets \yy](\psi \rimp (X \ne x_{k+1}))$ in
% $AX_{basic}^+$. Suppose that $x \notin \{x_1, \dots, x_k\}$, and
% substitute $x$ for $x_{k+1}$ everywhere in this proof. The result is
% a proof of $\phi \rimp \land_{i=1}^{k} [\YY \gets \yy](\psi \rimp (X
% \ne x_i)) \wedge [\YY \gets \yy](\psi \rimp (X \ne x))$ in
% $AX_{basic}^+$.
%joe17: let's use the notation
%%By assumption, there exists a proof of $\phi \rimp \band_{x \in S}[\YY
%  \gets \yy](\phi \rimp (X \neq x))$ in $AX_{basic}^+$. Suppose that
%$x' \notin S$, and replace $x$ with $x'$ everywhere in this proof. The
By assumption, $AX_{basic}^+ \shows \phi \rimp \band_{x \in S}[\YY
		\gets \yy](\phi \rimp (X \neq x))$.
%spencer18*: Do we need to say "This means there exists a proof..."? It seems to me the reviewers might read the next sentence and ask "what proof?"
%joe18: you can use ``derivation'' instead of proof, if you're uncomfortable
Thus, there exists a derivation of $\phi \rimp \band_{x \in S}[\YY
		\gets \yy](\phi \rimp (X \neq x))$ in $AX_{basic}^+$.
For all
%joe18
%$x' \notin S$, we can replace $x$ with $x'$ everywhere in this proof. The
$x' \notin S$, we can replace $x$ with $x'$ everywhere in this derivation. The
result is a
%spencer19:
% proof
derivation
of $\phi \rimp \band_{x \in S \cup \{x'\} \setminus
		\{x\}}[\YY \gets \yy](\phi \rimp (X \neq x))$ in $AX_{basic}^+$.
%joe17: moved this up one sentence, put it in parens, and slightly rewrote
%Notice that $\phi$ and $\psi$ do not undergo any substitution, because
(Notice that $\phi$ and $\psi$ remain unchanged, since
they do not mention
%spencer17:
% $x_{k+1}$.
%joe17
%$x$.
$x$.)
%spencer17: added.
%joe17
%In particular, this implies
%$AX_{basic} \rimp [\YY \gets \yy](\phi \rimp (X \neq x'))$.
Thus,
$AX_{basic}^+ \shows \phi \rimp [\YY \gets \yy](\phi \rimp (X \neq x'))$.
Using the assumption that $\R(X)$ is finite, it then follows that
$AX_{basic}^+ \vdash \phi \rimp \land_{x \in \R(X)} [\YY \gets \yy](\psi \rimp (X \neq x))$.
The result now follows from the argument given for the case $S = \R(X)$.
\eprf

\othm{theorem:GSEMs}
%joe6
%$AX'^+_{basic}$ is sound and complete for the class of GSEMs with
%joe34: removed paragraph break
%
%signature $\S$ over language $\L(\S)_{\WW, \R', \C}$.
%joe26
%joe34
%        If $A \subseteq \{$D3, D10(a),D10(b)$\}$ and $\A$ is the
        If $A \subseteq \{$D3, D6$^+$, D10(a),D10(b)$\}$ and $\A$ is the
corresponding subset of $\{coh, \ge 1, \le 1\}$, then
$AX^*_{basic,A}(\S,\WW,\R',\I')$ is sound and complete for the class $\G^{\A}$
of GSEMs with
signature $\S$ over language $\L_{\WW, \R', \I'}(\S)$.
\eothm
\prf
The soundness of all the axioms and inference rules with respect to
the appropriate class of structures is straightforward except for
D2$^+$.
To see that D2$^+$ is sound in $\G^{\A}(\S)$, suppose that
%To see that D2$^+$ is sound, suppose that
% $\phi$ is a formula in $\L_{\WW, \C, \I'}(\S)$,
%$\psi$ is an event of $\L_{\WW, \C, \I'}(\S)$,
%$\phi$ and $\psi$ only mention the elements $x_1, \ldots, x_k \in \R'(X)$,
%there is $x_{k+1} \in \R'(X) - \{x_1, \ldots, x_k\}$,
$\phi \rimp [\YY \gets \yy]\psi$ is a formula in $\L_{\WW,
		%spencer18:
		% \C,
		\R',
		\I'}(\S)$
that mentions only $x_1, \dots, x_k \in \R'(X)$,
and there is a value $x_{k + 1} \in \R'(X) - \{x_1, \dots, x_k\}$.
%and $\YY \gets \yy$ is an intervention that does not set $X$ (that is, $X \notin \YY$).
Suppose further that
the formula
%spencer8:
%\begin{equation}
%\label{eq:premise}
%%joe7
%%\phi \rimp \land_{i=1}^{x_{k+1}} [\YY \gets \yy](\psi \rimp (X \ne x_i))
%\rho = \phi \rimp \land_{i=1}^{k+1} [\YY \gets \yy](\psi \rimp (X \ne x_i))
%\end{equation}
%joe10
%$\theta$ defined by
$$\theta = \quad \phi \rimp \land_{i=1}^{k+1} [\YY \gets \yy](\psi \rimp (X \ne x_i))$$
is valid in $\G^{\A}(\S)$.
%joe10: removed paragraph break
%
We want to show that
$\phi \rimp [\YY \gets \yy]\neg \psi$ is valid in $\G^{\A}$.
%joe10: removed paragraph break
%
Suppose not.
Then there is a GSEM $M \in \G^{A}$ and context $\uu$ such that
$(M,\uu) \sat \phi  \land \<\YY \gets \yy\>\psi$.
%joe26
%Thus, there must be some solution $\vv^* \in \F(\uu,\YY \gets \yy)$
Thus, there must be some outcome $\vv^* \in \FF(\uu,\YY \gets \yy)$
such that $\vv^*$ satisfies $\psi$.
%joe18: removed paragraph break
%
Let $x^* := \vv^*[X]$.
%joe7
%Since (M,u) \sat \land_{i=1}^{x_{k+1}} [\YY \gets \yy](\psi \rimp (X
Since $(M,\uu) \sat \land_{i=1}^{k+1} [\YY \gets \yy](\psi \rimp (X
	\ne x_i))$,
it cannot be the case that
$x^* \in \{x_1, \ldots, x_{k+1}\}$.

%
%If we had a model $M' \in \G^{\A}$
%satisfying $\phi \land \<\YY \gets \yy\>\psi$,
%with a solution $\vv' \in M(\uu, \YY \gets \yy)$ satisfying $\psi$ such that
%$\vv[X] = x_{k + 1}$ instead of $\vv[X] = x$,
%then we would have a contradiction to the validity of Formula \ref{eq:premise}.
%joe7
%Now define a model $M'$ with signature $\S$ based on $M$ as follows.
%spencer8:
%Now define a model $M'$ with signature $\S$ as follows.
%spencer10: New M' construction.
Now define a model $M' = (\S, \FF')$ with the same signature $\S$ as
%joe26
%$M$, but modified solutions $\F'$, as follows. Let $\vv[x/X]$ denote
%the solution such that $\vv[x/X][X] = x$ and for all $Y \neq X$,
$M$, but a modified mapping $\FF'$, defined as follows. Let $\vv[x/X]$ denote
the outcome such that $\vv[x/X][X] = x$ and for all $Y \neq X$,
$\vv[x/X][Y] = \vv[Y]$.
%joe26
%Define the solution $\vv^r$ (the $r$ indicates ``replacing'' $x^*$ by
Define the outcome $\vv^r$ (the $r$ indicates ``replacing'' $x^*$ by
$x_{k+1}$) as $\vv^r = \vv[x_{k+1}/X]$ if $\vv[X] = x^*$, otherwise
$\vv^r = \vv$.

%Define $\vv^r$ as the solution resulting from replacing the assignment $\vv[X] = x$ with $\vv[X] = x_{k+1}$, in the case where $\vv[X] = x$ (the $r$ stands for ``replace''). More precisely,
%\begin{equation*}
%    \vv^r[Z] =
%    \begin{cases}
%        x_{k + 1}, & Z = X \wedge \vv[X] = x \\
%        \vv[Z], & \text{otherwise.} \\
%    \end{cases}
%\end{equation*}

%Define $\F'(\uu, \YY \gets \yy) = $
%    \begin{equation*}
%        \begin{cases}
%            \{\vv[x_{k+1}/X] \mid \vv \in M(\uu, \YY \gets \yy)\}
%                &\text{if } X \notin \YY \\
%            M(\uu, \YY \gets \yy)
%                &\text{if } X \in \YY \wedge \yy[X] \neq x_{k+1}\\
%            \{\vv[x_{k + 1}/X] \mid \vv \in \F(\uu, (\YY \gets \yy \setminus (X, x)))\}
%                &\text{if } X \in \YY \wedge \yy[X] = x_{k+1}.\\
%        \end{cases}
%    \end{equation*}

Suppose $X \notin \WW$. Define
$$\FF'(\uu, \WW \gets \ww) = \{
	\vv^r \mid \vv \in \FF(\uu, \WW \gets \ww)
	\}.$$
For $x \neq x_{k + 1}$, define
$$\FF'(\uu, \WW \gets \ww; X \gets x) = \FF(\uu, \WW \gets \ww; X \gets x).$$
Finally, define
\begin{align*}
	\FF'(\uu, & \WW \gets \ww; X \gets x_{k+1}) =                    \\
	\{
	         & \vv[x_{k + 1}/X] \mid \vv \in \FF(\uu, \WW \gets \ww)
	\}.
\end{align*}
We claim that $M'$ is a counterexample to the supposed validity of $\theta$ in $\G^A(\S)$. This claim consists of two subclaims: first, that $M'$ does not satisfy $\theta$, and second, that $M' \in \G^A(\S)$.

First we argue $M'$ does not satisfy $\theta$. It suffices to show
%joe10
%that a formula $\phi$ which does not mention the values $x, x_{k+1}$
%cannot distinguish $M$ and $M'$; that is, either $M \sat \phi$ and
%$M' \sat \phi$, or $M \not\sat \phi$ and $M' \not\sat \phi$. To
that a formula $\phi$ that does not mention the values $x, x_{k+1}$
cannot distinguish $M$ and $M'$; that is, $M \sat \phi$ iff
$M' \sat \phi$. To
show this, it suffices to consider atomic causal formulas, since all
other causal formulas are Boolean combinations of these. Let $\phi =
	%joe10
	%[\WW \gets \ww](\gamma) be an arbitrary atomic causal formula in $\I$
	%which does not mention $x$ or $x_{k + 1}$. The claim that
	%joe18
	%[\WW \gets \ww]\gamma$ be an arbitrary atomic causal formula in $\I$
	%$
	[\WW \gets \ww]\gamma$ be an atomic causal formula in $\I$
that does not mention $x$ or $x_{k + 1}$. The claim that
$M \sat \phi \Leftrightarrow M' \sat \phi$ follows from a
straightforward structural induction on $\gamma$ in the first two
cases defining $\FF'$, namely $X \notin \WW$ and
$X \in \WW, \ww[X] \neq x_{k + 1}$. The third case ($X \in \WW, \ww[X]
	%joe10
	%= x_{k + 1}) is not possible since by assumption,
	= x_{k + 1}$) is not possible since, by assumption,
$[\WW \gets \ww](\gamma)$ does not mention $x_{k + 1}$.

Next we argue that $M' \in \G^A(\S)$. $M'$ was defined to have signature $\S$, so it suffices to show that $M'$ satisfies the axioms of $A$.
%joe10: removed paragraph break
%
%It is evident from the definition of $\F'$ that
It is clear from the definition of $\FF'$ that
$|\FF'(\uu, \WW \gets \ww)| \leq |\FF(\uu, \WW \gets \ww)|$, where
$|\cdot|$ denotes set cardinality. To see this, note that in the
%joe26
%second case defining $\F'$, the two solution sets are identical,
second case defining $\FF'$, the two outcome sets are identical,
whereas in the first and third cases, $\FF'(\uu, \WW \gets \ww)$ is the
image of $\FF(\uu, \WW \gets \ww)$ under the maps $\vv \mapsto \vv^r$
%joe10: ``$\leq 1 \in A$'' is hard to parse
%and $\vv \mapsto \vv[x_{k+1}/X]$ respectively. Suppose $\leq 1 \in A$,
%joe10
%and $\vv \mapsto \vv[x_{k+1}/X]$ respectively. Suppose that $A$
and $\vv \mapsto \vv[x_{k+1}/X]$, respectively. Suppose that $A$
includes the property $\leq 1$,
so $M$ satisfies $\leq 1$. Then $|\FF'(\uu, \WW \gets \ww)| \leq
	|\FF(\uu, \WW \gets \ww)| \leq 1$, so $M'$ satisfies $\leq 1$. It is
%joe10
%also evident that if $|\F(\uu, \WW \gets \ww)| \geq 1$ then
also clear that if $|\FF(\uu, \WW \gets \ww)| \geq 1$, then
$|\FF'(\uu, \WW \gets \ww)| \geq 1$, since the image of a nonempty set
%joe10
%is a nonempty set. Hence if $\geq 1 \in A$, $M'$ satisfies $\geq 1$.
is a nonempty set. Hence, if $\geq 1 \in A$, $M'$ satisfies $\geq 1$.

It remains to show that if $coh \in A$, then $M'$ satisfies
%Joe10
%$coh$. That is,for all interventions $\ZZ \gets \zz$,
$coh$. That is, for all interventions $\ZZ \gets \zz$,
$\TT \gets \ttt$, all contexts $\uu$, and all assignments $\vv'$,
\begin{equation}\label{eq:coh}
	\begin{array}{ll}
		\mbox{if $\vv' \in \FF'(\uu, \ZZ \gets \zz)$ and $\vv'[\TT] = \ttt$,} \\
		\mbox{then $\vv' \in \FF'(\uu, \ZZ \gets \zz; \TT \gets \ttt)$.}
	\end{array}
\end{equation}
This turns out to
be the most involved part of the soundness proof.

%joe10
%It suffices to prove the above for the cases $\TT \gets \ttt = X \gets
%x$ and $\TT \gets \ttt = \RR \gets \rr$ with $X \notin \RR$. This
%follows from two facts. First, if the above holds for
%joe11
%It suffices to prove (\ref{eq:coh}) for two cases (a) $\TT \gets \ttt
%$
It suffices to prove (\ref{eq:coh}) for two cases: (a) $\TT \gets \ttt
	= X \gets x$ and (b) $X \notin \TT$. This
follows from two facts. First, if (\ref{eq:coh}) holds for
$\TT_1 \gets \ttt_1$ and $\TT_2 \gets \ttt_2$, it also holds for
%joe10
%$\TT_1 \gets \ttt_1; \TT_2 \gets \ttt_2$. To see this, assume
%$\vv' \in \F'(\uu, \ZZ \gets \zz)$. Using the above for
$\TT_1 \gets \ttt_1; \TT_2 \gets \ttt_2$. To see this, suppose that
$\vv' \in \FF'(\uu, \ZZ \gets \zz)$ and $\vv[\TT_1;\TT_2]
	= \ttt_1;\ttt_2$. Applying (\ref{eq:coh}) to
$\TT_1 \gets \ttt_1$ gives
$\vv' \in \FF'(\uu, \ZZ \gets \zz; \TT_1 \gets \ttt_1)$,
%joe10
% and using the above for $\TT_2 \gets \ttt_2$ with $\ZZ \gets \zz
%= \ZZ \gets \zz; \TT_1 \gets \ttt_1$ gives
then applying it to $\TT_2 \gets \ttt_2$  gives
$\vv' \in \FF'(\uu, \ZZ \gets \zz; \TT_1 \gets \ttt_1; \TT_2 \gets \ttt_2)$. Second,
any intervention $\TT \gets \ttt$ can be written as
% spencer13:
% either
$\TT_1 \gets \ttt_1; \TT_2 \gets \ttt_2$, where
%joe26: removed paragraph break
%
$X \notin \TT_1$, and either $\TT_2 = \emptyset$, if $X \notin \TT$; or $\TT_2 = \ttt_2 = X \gets x$, if $X \in \TT$.
We now prove (\ref{eq:coh}) for these two cases. Suppose that $\vv' \in
	\FF'(\uu, \ZZ \gets \zz)$ and
$\vv'[\TT] = \ttt$.
We want to show that $\vv' \in \FF'(\uu, \ZZ \gets \zz; \TT \gets \ttt)$.

%joe11
%We proceed by cases on $X \in \ZZ$. First suppose $X \notin \ZZ$. In
%this case, by definition of $\F'$, there is $\vv \in \F(\uu, \ZZ
%$
First suppose that $X \notin \ZZ$. Then,
by definition of $\FF'$, there is some $\vv \in \FF(\uu, \ZZ
	\gets \zz)$ with $\vv' = \vv^r$.
%joe11: overkill
%There are now two subcases,
%corresponding to $\TT \gets \ttt = X \gets x$, and $X \notin \TT$,
%respectively.
%
%Suppose $\TT \gets \ttt = X \gets x$. Suppose further $x \neq
%x_{k+1}$. We have $\vv[X] \neq x^*$; otherwise $\vv'[X] = \vv^r[X] =
%$
For case (a), if $x \neq
	x_{k+1}$, then $\vv[X] \neq x^*$; otherwise $\vv'[X] = \vv^r[X] =
	x_{k+1}$, but $\vv'[X] = x \neq x_{k+1}$, a contradiction. It follows
$\vv' = \vv$, since $\vv^r[X] = \vv[X]$ when $\vv[X] \neq x^*$. Thus
%joe11
%$\vv[X] = x$, and by coherence of $M$, $\vv \in \F(\uu, \ZZ \gets
%$
$\vv[X] = x$, and by the coherence of $M$, $\vv \in \FF(\uu, \ZZ \gets
	\zz; X \gets x)$. But by definition of $\FF'$, $\FF'(\uu, \ZZ \gets \zz; X
	\gets x) = \FF(\uu, \ZZ \gets \zz; X \gets x)$. Thus $\vv' = \vv \in
	\FF'(\uu, \ZZ \gets \zz; X \gets x)$, the desired conclusion.
%joe11
%Suppose instead $x = x_{k + 1}$. In this case $\vv' = \vv^r = \vv =
%$
And if $x = x_{k + 1}$, then $\vv' = \vv^r = \vv =
	\vv[x_{k+1}/X]$. But by definition of $\FF'$, $\FF'(\uu, \ZZ \gets \zz; X
	\gets x_{k + 1}) = \{\vv_1[x_{k+1}/X] \mid \vv_1 \in \FF(\uu, \ZZ \gets
	\zz)\}$. In particular, $\vv' = \vv[x_{k+1}/X] \in \FF(\uu, \ZZ \gets
	%\zz; X \gets x_{k+1})$ as desired.$
	\zz; X \gets x_{k+1})$ as desired.

%joe11
%Next suppose $X \notin \TT$. In this case, $\vv[\TT] = \vv^r[\TT] =
%\ttt$, so by coherence of $M$, $\vv \in \F(\uu, \ZZ \gets \zz; \TT
%$
For case (b), if $X \notin \TT$, then $\vv[\TT] = \vv^r[\TT] =
	\ttt$, so by the coherence of $M$, $\vv \in \FF(\uu, \ZZ \gets \zz; \TT
	\gets \ttt)$. Thus by definition of $\FF'$, $\vv^r \in \FF'(\uu, \ZZ
	\gets \zz; \TT \gets \ttt)$. But $\vv' = \vv^r$, so $\vv' \in \FF'(\uu,
	\ZZ \gets \zz; \TT \gets \ttt)$, as desired.

%joe11
%Now we consider the case $X \in \ZZ$. Let $x_\zz = \zz[X]$. Again,
%there are two subcases, corresponding to $\TT \gets \ttt = X \gets x$,
%and $X \notin \TT$.
%Suppose $\TT \gets \ttt = X \gets x$. We have $\vv'[X] = x_{\zz}$, by
%effectiveness, but $\vv'[X] = x$, by assumption $\vv'[\TT] =
%\ttt$. Thus $x_\zz = x$. Hence $\ZZ \gets \zz; \TT \gets \ttt = \ZZ
%$
Now suppose that $X \in \ZZ$ and $x_{\zz} = \zz[X]$.
For case (a), we have $\vv'[X] = x_{\zz}$, by effectiveness, and
$\vv'[X] = x$ by assumption, so
$x_{\zz} = x$. Hence, $\ZZ \gets \zz; \TT \gets \ttt = \ZZ
	% joe11*: I'm not sure = is right here; they are equivalent.
	% spencer14: I think = is correct. The two expressions are equal by
	% the definition of the composition operation ; that we had in the
	% GSEM paper.
	%joe12: OK; it may be worth at least saying this (and pointing to the
	%other paper for the formal definition)
	%spencer15: Makes sense.
	\gets \zz; X \gets x = \ZZ \gets \zz$. The desired conclusion $\vv'
	\in \FF'(\uu, \ZZ \gets \zz; \TT \gets \ttt)$ trivially follows from
the assumption $\vv' \in \FF'(\uu, \ZZ \gets \zz)$.

%joe11
%Next suppose $X \notin \TT$.
For case (b),
%joe11*: I would prefer not to proliferate letters for sets of
%variables; I've replaced \RR by \ZZ' globally.  More importantly, how
%do you know that you can write \ZZ like this?  Specifically, why does
%X <- x_{k+1} in Z?  I'm assuming that you meant x, not x_{k+1}.
%Write $\ZZ \gets \zz = \RR \gets \rr; X \gets x_{k+1}$, where $X \notin \RR$.
%spencer14: I agree with \RR -> \ZZ change.
% Also you are right--I meant x, not x_{k + 1}. Good catch.
write $\ZZ \gets \zz = \ZZ' \gets \zz'; X \gets x$, where $X \notin \ZZ'$.
%joe1
%Further suppose $x \neq x_{k+1}$. By definition of $\F'$, $\F'(\ZZ$
If $x \neq x_{k+1}$, then by definition of $\FF'$, we have that $\FF'(\uu, \ZZ
	\gets \zz) = \FF(\uu, \ZZ \gets \zz)$. Hence $\vv' \in \FF(\uu, \ZZ \gets \zz)$,
%joe11
%and by coherence of $M$, $\vv' \in \F(\uu, \ZZ \gets \zz; \TT \gets \ttt)
%$
and by the coherence of $M$, $\vv' \in \FF(\uu, \ZZ \gets \zz; \TT \gets \ttt)
	= \FF(\uu, \ZZ' \gets \zz'; X \gets x; \TT \gets \ttt) = \FF(\uu, \ZZ' \gets \zz'; \TT
	%joe11
	%\gets \ttt; X \gets x)$. But again by definition of $\F'$, $\F(\uu, \ZZ'
	\gets \ttt; X \gets x)$. But again, by definition of $\FF'$, $\FF(\uu, \ZZ'
	\gets \zz'; \TT \gets \ttt; X \gets x) = \FF'(\uu, \ZZ' \gets \zz'; \TT \gets
	\ttt; X \gets x) = \FF'(\uu, \ZZ' \gets \zz'; X \gets x; \TT \gets \ttt) =
	\FF'(\uu, \ZZ \gets \zz; \TT \gets \ttt)$. The desired conclusion $\vv' \in
	\FF(\uu, \ZZ \gets \zz; \TT \gets \ttt)$ follows.
%joe11
%Suppose instead $x = x_{k+1}$.
%By definition of $\F'$, $\F'(\uu, \ZZ \gets \zz) = \{\vv[x_{k+1}/X]
%$
On the other hand, if $x = x_{k+1}$, then
by definition of $\FF'$, $\FF'(\uu, \ZZ \gets \zz) = \{\vv[x_{k+1}/X]
	\mid \vv \in \FF(\uu, \ZZ' \gets \zz')\}$.
%joe11
%Thus there is $\vv \in \F(\uu, \ZZ' \gets \zz')$ with $\vv' = \vv[x_{k+1}/X]$
Thus, there is some $\vv \in \FF(\uu, \ZZ' \gets \zz')$ with $\vv' =
	\vv[x_{k+1}/X]$.
%joe26
%        Further, $\vv[T] = \vv'[T] = \ttt$, since $\vv[x_{k+1}/X]$ only
        %differs from $\vv'$ at $X$, if at all, and $X \notin \TT$. So, by the
Further, $\vv[T] = \vv'[T] = \ttt$, since $\vv[x_{k+1}/X]$ differs
from $\vv'$ only at $X$, if at all, and $X \notin \TT$. So, by the
coherence of $M$, $\vv \in \FF(\uu, \ZZ' \gets \zz'; \TT \gets
	\ttt)$. Thus again by definition of $\FF'$, $\vv' = \vv[x_{k+1}/X] \in
	\FF'(\uu, \ZZ' \gets \zz'; \TT \gets \ttt; X \gets x_{k+1})$. But $\ZZ'
	\gets \zz'; \TT \gets \ttt; X \gets x_{k+1} = \ZZ' \gets \zz'; X \gets
	x_{k+1}; \TT \gets \ttt = \ZZ \gets \zz; \TT \gets \ttt$, so we have
%joe11
%$\vv' \in \F'(\uu, \ZZ \gets \zz; \TT \gets \ttt)$ as desired.
$\vv' \in \FF'(\uu, \ZZ \gets \zz; \TT \gets \ttt)$, as desired.

Following the sketch in the main text, to prove completeness, we first prove Lemma \ref{lem:acceptable}. Then, after constructing an acceptable maximal consistent set $C$ as in the sketch, we construct a GSEM $M^C$ with signature $\S$ and show that it satisfies two properties.  First, it satisfies the appropriate axioms; that is, $M^C \in \G^\A$. Second, for all formulas $\psi \in \L(\WW, \R', \I')(\S)$, we have $\phi \in C$ if and only if, for all contexts $\uu$, $(M^C, \uu) \models \psi$.

%spencer18*: Can we call these 'conjunctive events'? I'd like to keep
%a clear distinction between events and formulas.
%joe18: Why not call it a conjunctive formula?  Events are usually
%taken to be sets of worlds (i.e., semantic objects), whereas formulas
%as syntactic objects.
%spencer19: That makes sense. My issue with this is that we're already
%using the term 'formula' for causal formulas [\YY \gets
%\yy]\phi. Here, I would prefer not to call \phi a formula, since it
%isn't in fact a causal formula. Do you think 'event' has confusing
%connotations? It does have the virtue of being consistent with your
%book.
%joe19: Unfortunately, ``event'' does have other connotations,
%although there are certainly papers that conflate events and formulas
%out there.  Here's a suggestion, that I think would result in minimal
%changes to the paper.   I think it's reasonable to call [\YY
%\gets \yy]\phi an atomic causal formula and say that we sometimes
%drop ``causal'' and just say ``atomic formula''.  We can also call
%Boolean combinations of atomic causal formulas causal formulas, and
%again, occasionally drop causal.  We then call X=x a primitive
%(propositional) formula, and a conjunction of primitive formlas can
%be an conjunctive formula.  That way, we reserve ``formula'' for a
%syntactic object.
%spencer37:
%spencer45: getting rid of redundant definition and lemma statement.
\commentout{
\odefn{definition:conjunctive-formula}
% \begin{restatable}{definition}{conjunctiveFormula}
A \emph{conjunctive formula} is a conjunction of formulas of the
form  $X = x$ and $X \ne x$.  (The formula $\true$ is viewed as
conjunctive, since it is an empty conjunction.)
A set $C$ of formulas in $\L_{\WW,\R',\I'}(\S)$
is \emph{acceptable for $\<\YY \gets \yy\>\phi$
	with respect to $Z \in \WW$}, where $\phi$ is a conjunctive formula,
if there is some $z \in \R'(Z)$ and a conjunctive formula $\psi$ such that every
conjunct of $\phi$ is a conjunct of $\psi$, $Z=z$ is a conjunct of
%joe18
%$\psi$, and $\<\YY \gets \yy\>\psi \in C$.  $C$ is \emph{acceptable} if
$\psi$ for some $z \in \R'(Z)$, and $\<\YY \gets \yy\>\psi \in C$.  $C$ is \emph{acceptable} if
$C$ is acceptable for every formula of the form $\<\YY \gets \yy\>\phi \in
	C$ such that $\phi$ is conjunctive formula and $Z \in \WW$.
%spencer18: Note to self: intuitively, this means that any formula in
%C can be extended to another formula in C which contains X = x, so
%we're not allowed to rule out all values of X.
\eodefn
% \dfn
% A \emph{conjunctive formula} is a conjunction of formulas of the
% form  $X = x$ and $X \ne x$.  (The formula $\true$ is viewed as
% conjunctive, since it is an empty conjunction.)
% A set $C$ of formulas in $\L_{\WW,\R',\I'}(\S)$
% is \emph{acceptable for $\<\YY \gets \yy\>\phi$
% 	with respect to $Z \in \WW$}, where $\phi$ is a conjunctive formula,
% if there is some $z \in \R'(Z)$ and a conjunctive formula $\psi$ such that every
% conjunct of $\phi$ is a conjunct of $\psi$, $Z=z$ is a conjunct of
% %joe18
% %$\psi$, and $\<\YY \gets \yy\>\psi \in C$.  $C$ is \emph{acceptable} if
% $\psi$ for some $z \in \R'(Z)$, and $\<\YY \gets \yy\>\psi \in C$.  $C$ is \emph{acceptable} if
% $C$ is acceptable for every formula of the form $\<\YY \gets \yy\>\phi \in
% 	C$ such that $\phi$ is conjunctive formula and $Z \in \WW$.
% %spencer18: Note to self: intuitively, this means that any formula in
% %C can be extended to another formula in C which contains X = x, so
% %we're not allowed to rule out all values of X.
% \edfn

%joe10
%The set $C'$ above is not acceptable for $\<Y \gets y>true$ and $X$, and
The set $C'$ above is not acceptable for $\<Y \gets y\>true$ and $X$, and
cannot be extended to an acceptable consistent
set.  Our proof technique involves constructing a model from an
acceptable maximal consistent set.  So we must show that every consistent
formula
%spencer18: typo
% in
%joe26
%
%spencer18*: Would it make more sense to say 'there exists an
%acceptable maximal consistent set'?
%joe18: there does exist an acceptable maximal consistent set, but
%we're making a stronger statement here: every consistent formula is
%contained in some acceptable maximal consistent set.
%spencer19: Oh, ok. Where does this stronger claim get used later in the proof?
%joe19 We certainly use it for the main reult: if \phi is consistent,
%then its satisfiable.  The construction shows that every consistent
%formula is satisfiable in the canonical model that we construct,
%since every consistent formula is a maximal consistent set, and thus
%true in the state on the canonical model corresopnding to that set.
is
included in an acceptable maximal consistent set.
%We actually prove a stronger result, which will be needed for our
%induction hypothesis. Say that a set $C$ of formulas is \emph{safe}
%if, for each variable $X \in \WW$, if $\R'(X)$ is infinite, then there is a
%countable subset $F_X$ of  $\R'(X)$ such
%that none of the values in $F_X$ appear in a formula in $C$.  The
%result we need is summarized in the following lemma.
The next lemma gives the key step for doing this.

\olem{lem:acceptable} If $C$ is a finite subset of
	$\L_{\WW,\R',\I'}(\S)$
	consistent with $AX^*_{basic,A}(\S,\WW,\R',\I')$,
	$\<\YY \gets \yy\>\phi \in C$, and $X \in
		%joe18
		%  \WW$, then we can add a formula $\psi \in \L_{\WW,\R',\I'}(\S)$ such
		\WW$, then there exists a formula $\psi \in \L_{\WW,\R',\I'}(\S)-C$ such
	that $
		%spencer18:
		% \C
		%$
		C
		\union \{\psi\}$ is consistent with $AX^*_{basic,A}(\S,\WW,\R',\I')$
	and acceptable with respect to $\<\YY \gets \yy\>\phi$ and $X$.
  \eolem
} %spencer45: end commentout
%spencer45:
%\prf
\textbf{Proof of Lemma \ref{lem:acceptable}:}
Let $C$ be a finite subset of $\L_{\WW,\R',\I'}(\S)$ consistent
with $AX^*_{basic,A}(\S,\WW,\R',\I')$.
Let $\theta$ be the conjunction of the
formulas in $C$, and let $\{x_1, \ldots, x_k\}$ be the values in
$\R'(X)$ mentioned in $\theta$.
There are two cases.  If there exists
a value $x_{k+1} \in \R'(X) - \{x_1, \ldots, x_k\}$,
then we claim that one of the formulas

$\theta \land \<\YY \gets \yy\>(\phi \land X=x_i)$, $i = 1, \ldots, k+1$
must be consistent with $AX^*_{basic,A}(\S,\WW,\R',\I')$.
For if not, then
$\theta \rimp [\YY \gets \yy](\phi \rimp X \ne x_i)$
must be provable in
$AX^*_{basic,A}(\S,\WW,\R',\I')$ for $i = 1, \ldots, k+1$.
By D2$^+$, so is   $\theta \rimp [\YY \gets \yy]\neg \phi$.  But{}
since, by assumption,  $\<\YY \gets \yy\>\phi$ is one of the
conjuncts of $\theta$, $\theta \rimp \<\YY \gets \yy\>\phi$ is
also provable, so $\theta$ is not consistent with
$AX^*_{basic,A}(\S,\WW,\R',\I')$, contradicting the consistency of $C$.

If there does not exist a value $x_{k+1} \in \R'(X) - \{x_1, \ldots,
	%joe10
	%x_k\}$, then $\R'(X) = \{x_1, \ldots, x_k\}$, so by D2, $[\YY \gets
	x_k\}$, then $\R'(X) = \{x_1, \ldots, x_k\}$, so by D2$^+$, $[\YY \gets
			%joe18
			%  \yy](x_1 \lor \ldots \lor x_k)$
			\yy](X= x_1 \lor \ldots \lor X= x_k)$
is provable in $AX^*_{basic,A}(\S,\WW,\R',\I')$.  Thus, so is $\theta \rimp
	\<\YY \gets \yy\>
	%spencer7:
	%(\phi \land (x_1 \lor \ldots \lor x_k))$.
	%$
	%joe18
	%(\phi \land X = x_1 \lor \ldots \lor X = x_k)$
	(\phi \land X = x_1 \lor \ldots \lor X = x_k)$.
It easily
follows that $\theta \land \<\YY \gets \yy\>(\phi
	%spencer7:
	%\land x_1)$
	%joe18
	%\land x_i)$
	\land X=x_i)$
        %joe26: there seem to be a number of inadvertant paragraph breaks
        %
  %spencer28: Thanks for catching those.
must be
consistent for some $i = 1, \ldots, k$.
In either case, we can add a formula of the form $\<\YY \gets \yy\>(\phi
	%joe18
	%\land x)$ to $C$ while maintaining its consistency.  \eprf
\land X=x)$ to $C$ while maintaining its consistency.
%spencer45:
%\eprf
\qed
%$

%spencer45: removing redundancy
\commentout{
We can now prove completeness.   Given a formula $\phi$ consistent
%joe18
%with $AX^*_{basic,A}(\S,\WW,\R',\I')$ we
with $AX^*_{basic,A}(\S,\WW,\R',\I')$, we
construct a maximal acceptable set $C$ consistent with
$AX^*_{basic,A}(\S,\WW,\R',\I')$ containing $\phi$ as follows.    Let
$\sigma_0, \sigma_1, \sigma_2, \ldots$ be an enumeration of the
formulas in $\L_{\WW,\R',\I'}(\S)$ such that $\sigma_0 = \phi$ and let $X_0, X_1, X_2, \ldots $ be
an enumeration of the variables in $\WW$.  It is well known that
there is a bijection $b$ from $\IN$ to $\IN \times \IN$ such that if
$b(n) = (n_1, n_2)$, then $n_1 \le n$.  We construct a sequence of
sets $C_0, C_1,
	C_2, \ldots$ such that $C_0 = \{\phi\}$, $C_k \subseteq C_{k+1}$, and
(a) either $\sigma_k \in C_k$ or $C_k \cup \{\sigma_k\}$ is
inconsistent with $AX^*_{basic,A}(\S,\WW,\R',\I')$, (b) $C_k$ is
consistent, (c) if $b(k) = (k_1,k_2)$ and $\sigma_{k_1}$ has the form
$\<\YY \gets \yy\>\phi$, then $C_{k+1}$ is acceptable with respect to
$\sigma_{k_1}$ and $X_{k_2}$.  We construct the sequence inductively.
%joe10
%Given $C_k$, then we add $\sigma_{k+1}$ to $C_k$ if C_k \union
Given $C_k$, then we add
%spencer18*: I'm pretty sure you meant
% $\sigma_{k_1}$
$\sigma_{k + 1}$
to $C_k$ if $C_k \union
	\{\sigma_{k+1}\}$ is consistent.  In addition, if $\sigma_{k_1}$ has
the form $\<\YY \gets \yy\>\phi$, then we apply
Lemma~\ref{lem:acceptable} to add a formula if necessary to make
$C_{k+1}$ acceptable with respect to $\sigma_{k_1}$ and $X_{k_2}$.

Let $C= \union_{k=0}^\infty C_k$.  Clearly $C$ contains $\phi$.  It is
consistent, since if not, some finite subset of $C$ must be
inconsistent.  But this finite subset must be contained in $C_k$ for
some $k$, and $C_k$ is consistent, by construction.  Finally, $C$ is
%spencer18: typo
% accsptable.
acceptable.
For suppose that $\<\YY \gets \yy\> \phi \in C$ and $X
	\in \WW$.    There must exist $k_1$ and $k_2$ such that
$\<\YY \gets \yy\> \phi = \sigma_{k_1}$ and $X = X_{k_2}$.  Let $k =
	b^{-1}(k_1,k_2)$. Since $\sigma_{k_1} = \<\YY \gets \yy\> \phi \in C$,
it must already be in $C_{k_1}$ (since it would be added in the
construction of $C_{k_1}$ if it was not already in $C_{k_1-1}$). By
the choice of $b$, $k_1 \le k$, so $\sigma_{k_1} \in C_k$.  By
construction, $C_{k+1}$ is acceptable with respect to $\<\YY \gets
	%joe18
	%\yy\> \phi$ and $X_{k_2}$, hence is $C$.
	\yy\> \phi$ and $X_{k_2}$, hence so is $C$.
  %$
}  %spencer45: end commentout

%spencer45:
%We now construct a model $M^C$ with signature $\S$.
Given the acceptable maximal consistent set $C$ constructed in the sketch,
we now construct a GSEM $M^C$ with signature $\S$.
For
interventions
%spencer7: changed defn above so that \I' subseteq \I
%$\YY \gets \yy in \I \inter \I'$,
%joe10
%$\YY \gets \yy in \I'$,
$\YY \gets \yy \in \I'$,
we take $\FF(\YY \gets \yy, \uu) =
	%spencer18: for consistency
	% \{\vv:$ for all finite subsets $\XX \subseteq \WW$,
	\{\vv \mid$ for all finite subsets $\XX \subseteq \WW$,
%spencer18:
%$\< \yy
$\<\YY
	\gets \yy\>(\XX = \vv[\XX]) \in C\}$.
Now we still need to
%joe10
%define $\F$ on interventions in $\I - \I'$.  If
define $\FF$ on interventions in $\I - \I'$.
Let $\vv^*$ be a fixed assignment. For $I \in \I - \I'$, define
$\FF(I,\uu) = \{\vv^*\}$.
%joe10: removed paragraph break
%
We claim that $M^C \in \G^\A$ and that $\psi \in C$ iff, for all
contexts $\uu$, we have $(M^C,\uu) \sat \psi$.
%spencer18*: added
If these claims hold, then we have produced a model $M^C$ of $\phi$ that satisfies the properties in $\A$, completing the consistency proof.
%spencer37: now duplicated
% \newcommand{\LL}{\L_{\WW, \R', \I'}(\S)}

First we show that for all contexts $\uu$ and formulas $\phi \in \LL$, we have
\begin{equation}
	\label{eq:truth-lemma}
	\phi \in C \dimp (M^C, \uu) \sat \phi.
\end{equation}
%joe18
%Fix an arbitrary $\uu \in \U$. It suffices to show
Fix a context $\uu \in \U$. It suffices to show
(\ref{eq:truth-lemma}) for the
%spencer28*: Is this ok? We no longer say that the <> formulas are
%dual basic causal formulas in the introduction.
%joe21: dual is a fairly standard notion; I think it's fine
dual basic causal formulas $\phi =
	\<\YY \gets \yy\>\rho$. This is because it is easy to see that every
causal formula
is a Boolean combination of dual basic causal formulas. The result
%joe18
%follows for arbitrary causal formulas by a straightforward structural
follows for all causal formulas by a straightforward structural
induction.

%joe18
%Now $\rho$ is a Boolean combination of primitive events $X = x$. We
Now $\rho$ is a Boolean combination of primitive events of the form $X = x$. We
can write $\rho$ in DNF.
That is, there exists $k \geq 0$ and conjunctive events $d_i$ for $1 \leq i \leq k$ such that $\rho \dimp \bor_{1 \leq i \leq k} d_i$ is a propositional tautology.

% That is, there exists $k \geq 1$ and sets of variable-value pairs $I_i^=, I_i^{\neq}$ for $i = 1, 2, \dots, k$ such that $\rho \dimp d$ is a propositional tautology

% $d = \bor_{1 \leq i \leq k} d_i$,
% \[\rho \dimp \bor_{1 \leq i \leq k} ((\band_{(X, x) \in I_i^=} X = x) \land (\band_{(X, x) \in I_i^{\neq}} X \neq x))\]
% is a propositional tautology. Let $d$ denote the RHS above, and $d_i$ denote the $i$-th disjunct of $d$, so that $d = \bor_{1 \leq i \leq k} d_i$
% For $1 \leq i \leq k$, let$d_i$ denote the $i$-th disjunct above.
%joe18: removed paragraph break
%
%Then by the dual form of Lemma~\ref{lemma:and-distrib-box}, we have
%joe26
%By the dual form of Lemma~\ref{lemma:and-distrib-box}, we have
By Lemma~\ref{lemma:and-distrib-box}(b), we have
$AX_{basic} \shows \<\YY \gets \yy\>\rho \dimp \bor_{1 \leq i \leq k}
	\<\YY \gets \yy\>d_i$.
% All the events $(\band_{(X, x) \in I_i^=} X = x) \land (\band_{(X, x) \in I_i^{\neq}} X \neq x)$  appearing in this formula are conjunctive formulas.
So, again by structural induction, it suffices to show
(\ref{eq:truth-lemma}) for formulas $\phi$ of the form $\phi = \<\YY
	\gets \yy\>\gamma$, where $\gamma$ is a conjunctive formula.
%joe18

%spencer19:
% We now do this
We now do this.
%spencer19: removed paragraph break
%
%joe18
%For the backwards direction, suppose $(M^C, \uu) \models \<\YY \gets
For the ``only if'' direction, suppose that $(M^C, \uu) \models \<\YY \gets
	\yy\>\gamma$.
%joe18: removed paragraph break
%
%Then there is $\vv \in M^C(\YY \gets \yy, \uu)$ such that $\vv
Then there is some assignment $\vv \in M^C(\uu, \YY \gets \yy{})$ such that $\vv
	\models \gamma$.
By definition of $M^C$, for all finite sets $\XX \subseteq \WW$, we
have $\<\YY \gets \yy\>(\XX = \vv[\XX]) \in C$. Let $\ZZ$ be the
%joe18*: more complicated than necessary
%(finite) set of variables mentioned in $\gamma$; then in particular,
%$\<\YY \gets \yy\>(\ZZ = \vv[\ZZ]) \in C$. But this formula, along
%with D0, D1, D7, and D8 implies $\<\YY \gets \yy\>\gamma$.  We will
%show equivalently that D0, D1, D7, and D8 imply $(\<\YY \gets
%\yy\>(\ZZ = \vv[\ZZ]) \land [\YY \gets \yy] (\neg \gamma)) \rimp
%false$. D0, D1 and D8 imply that $[\YY \gets \yy](\ZZ = \vv[\ZZ] \rimp
%\gamma)$. Using D0, D7, and D8, it follows $[\YY \gets \yy](\neg
%\gamma \rimp \neg (\ZZ = \vv[\ZZ]))$ Applying D7, it follows $[\YY
%  \gets \yy](\neg (\ZZ = \vv[\ZZ]))$. However, $\<\YY \gets \yy\>(\ZZ
%= \vv[\ZZ])$ is just an abbreviation for $\neg [\YY \gets \yy] (\neg
%(\ZZ = \vv[\ZZ]))$. Combining this with the previous statement yields
%$false$.
(finite) set of variables mentioned in $\gamma$; then, in particular,
$\<\YY \gets \yy\>(\ZZ = \vv[\ZZ]) \in C$. Using
D0, D1, D7, and D8, we get that $AX^*_{basic,A}(\S,\WW,\R',\I') \stur
	\<\YY \gets \yy\>(\ZZ = \vv[\ZZ]) \rimp \<\YY \gets \yy\>\phi$.
Since $C$ is a maximal consistent set, we must have
$\<\YY \gets \yy\>(\ZZ = \vv[\ZZ]) \rimp \<\YY \gets \yy\>\phi \in C$.
The fact that $C$ is a maximal consistent set implies that it is
closed under implication, so since $\<\YY \gets \yy\>(\ZZ = \vv[\ZZ])
	\in C$, we must have $\<\YY \gets \yy\>\phi \in C$, as desired.
%spencer18*: Should we move the paragraph from the appendix explaining
%this reasoning style here?
%joe18: this should probably also go in the appendix (perhps following
%a one-paragraph summary of the proof).  If we do that, then there's
%no need.
%spencer19: makes sense.
%It follows from the maximality of $C$ that $\<\YY \gets \yy\>\gamma \in C$.

Conversely, suppose that $\<\YY \gets \yy\>\gamma \in C$. Fix an
%joe18
%enumeration $X_i, i \geq 1$ of the set of named variables $\WW$. For
enumeration $X_i, i \geq 1$, of the set of named variables $\WW$. For
$i \geq 1$, let $\ZZ_i = \{X_1, \ldots, X_i\}$. Let $\phi_0 =
	\gamma$. Since $C$ is acceptable, we can find formulas $\<\YY \gets
	\yy\>\phi_1, \<\YY \gets \yy\>\phi_2, \ldots \in C$ such that for $i
	\geq 1$, every conjunct of $\phi_{i - 1}$ is a conjunct of $\phi_i$,
%joe18
%and there is some $x_i \in \R(X_i)$ such that $X_i = x_i$ is a
and for some $x_i \in \R(X_i)$,  $X_i = x_i$ is a
conjunct of $\phi_i$. Let $\overline{\ww}$ be any assignment to the
variables $\V - \WW$, and define $\vv = \{(X_i, x_i) \mid i \geq 1\}
	\cup ((\V - \WW) \gets \overline{\ww}).$ Since every conjunct of
$\gamma$ is a conjunct of each $\phi_i$, we have $\vv \sat
	\gamma$. But every finite subset $S$ of $\WW$ is contained in $\ZZ_i$
for some $i$.
%joe18
%Thus $\<\YY \gets \yy\>\phi_i \rimp \<\YY \gets \yy\>(S
%= \vv[S])$. It follows from the maximality of $C$ that for all finite
Thus, $AX_{basic}^* \stur \<\YY \gets \yy\>\phi_i \rimp \<\YY \gets \yy\>(S
	= \vv[S])$ for all $i$ sufficiently large. It follows from the
maximality of $C$ that for all finite
subsets $S$ of $\WW$, $\<\YY \gets \yy\>(S = \vv[S]) \in C$. By
definition of $M^C$, it follows $\vv \in M^C(\YY \gets \yy,
	\uu)$. But then by definition of $\models$, we have $(M^C, \uu)
	\models \gamma$, as desired.
  % END COPY
%spencer31: Added missing \uu's below.
It remains to show that $M^C \in \G^\A$, that is, that $M^C$
satisfies all the properties of $\A$. We proceed one property at a
time. Suppose that $\geq 1 \in \A$. Then $A$ contains D10(a). Fix an
%joe18
%arbitrary
intervention $\YY \gets \yy$ and context $\uu \in
	\R(\U)$. If $\YY \gets \yy \notin \I'$, then $M^C(\uu, \YY \gets \yy)
%joe26
%	= \{v^*\}$, so there is exactly one solution to $\YY \gets \yy$. If
	= \{v^*\}$, so the intervention $\YY \gets \yy$ has exactly
        one outcome. If
$\YY \gets \yy \in \I'$, then by D10(a), we have $\<\YY \gets
	\yy\>true \in C$. Since $C$ is acceptable, we can use the construction
in the proof of the truth lemma to find $\vv$ such that $\<\YY \gets
	\yy\>(\XX = \vv[\XX]) \in C$ for all finite subsets $\XX$ of
%joe18
%$\WW$. Then by definition of $M^C$, $\vv \in M^C(\YY \gets \yy,
$\WW$. Then, by the definition of $M^C$, $\vv \in M^C(\uu, \YY \gets \yy
%joe26
        %	\uu)$, so there is at least one solution to $\YY \gets \yy$.
	)$, so the intervention $\YY \gets \yy$ has at least one outcome.

%joe18: added paragraph break
Next suppose that $\leq 1 \in \A$. Then $A$ contains D10(b). Again,
%joe18
%the case $\YY \gets \yy \notin \I'$ is immediate. Fix arbitrary $\YY
%\gets \yy \in \I'$. Suppose that $\vv, \vv' \in M^C(\YY \gets \yy,
%\uu)$. By definition of $M^C$, we must have $\<\YY \gets \yy\>(X =
the case $\YY \gets \yy \notin \I'$ is immediate. Suppose that $\YY
	\gets \yy \in \I'$. If $\vv, \vv' \in M^C(\uu, {}\YY \gets \yy)$. then by definition of $M^C$, we must have $\<\YY \gets \yy\>(X =
	\vv[X]) \in C$ for all $X \in \V$. Suppose for contradiction that $\vv
	\neq \vv'$. Then there is a variable $Z$ such that $\vv[Z] \neq
	%joe18
	%\vv'[Z]$, and so $\<\YY \gets \yy\>(Z = \vv'[Z])$. But by D10(b), for $
	\vv'[Z]$.  Since $\vv' \in M^C(\YY \gets \yy, \uu)$, we must have
$\<\YY \gets \yy\>(Z = \vv'[Z]) \in C$. By D10(b), for
all $\psi \in \LL$, we have $\<\YY \gets \yy\>\psi \rimp [\YY \gets
		\yy]\psi \in C$.
%spencer19*: added to address issue below
%joe19
%Thus we have $[\YY \gets \yy](Z = \vv'[Z]) \in C$. Then by D2, we have
Thus, $[\YY \gets \yy](Z = \vv'[Z]) \in C$. Then by D2, we have
$[\YY \gets \yy](Z \neq \vv[Z]) \in C$, and so
$[\YY \gets \yy](Z = \vv[Z]) \land [\YY \gets \yy](Z = \vv'[Z]) \in
	C$. It follows from Lemma~\ref{lemma:and-distrib-box} that $[\YY \gets
			%joe18
			%  \yy](Z = \vv[Z] \wedge Z = \vv'[Z]) \in C$. That is, $[\YY \gets
			\yy](Z = \vv[Z] \wedge Z = \vv'[Z]) \in C$, so $[\YY \gets
			%joe18: I'm not sure it exactly follows from anything we said above
			%  \yy]false \in C$.  But from above, we have $\<\YY \gets \yy\>true
			%spencer19: I see what you're saying--fixed above.
			\yy]false \in C$.  But since $\<\YY \gets \yy\>true
	\in C$, and $\<\YY \gets \yy\>true$ is just an abbreviation for $\neg
	%joe18
	%[\YY \gets \yy] false$. So $C$ is inconsistent, a$
	[\YY \gets \yy] false$, $C$ is inconsistent, a
        contradiction. Hence $\vv = \vv'$, and so
%joe26
        %        the model has at most one solution for $\YY \gets \yy$.
the intervention $\YY \gets \yy$ has at most one outcome.

Now suppose that $coh \in \A$. Then $A$ contains D3. Fix an
%joe18
%arbitrary intervention $\YY \gets \yy$ and context $\uu \in$
intervention $\YY \gets \yy$ and context $\uu \in
\R(U)$.
%spencer34: Typo!
% Suppose that $\vv \in M(\uu, \YY \gets \yy)$ and that $\vv[W] = w$.
Suppose that $\vv \in M^C(\uu, \YY \gets \yy)$ and that $\vv[W] = w$.
We want to show that $\vv \in M^C(\uu, \YY \gets \yy; W \gets w)$. First suppose that $\YY \gets \yy \notin \I'$. Then (because
$\I'$ is closed under finite differences with $\I$), we have $\YY
	%joe18
	%\gets \yy; W \gets w \notin I'$ as well. By definition of $M^C$, we
%spencer34: typo!
% \gets \yy; W \gets w \notin I'$. By the definition of $M^C$, we
	\gets \yy; W \gets w \notin \I'$. By the definition of $M^C$, we
have $M^C(\uu, \YY \gets \yy) = M^C(\uu, \YY \gets \yy; W \gets w) =
	%joe18
	%\{v^*\}$. Thus clearly if $\vv \in M^C(\uu, \YY \gets \yy)$ then $\vv
%spencer34: typo
	%\{v^*\}$. Thus, if $\vv \in M^C(\uu, \YY \gets \yy)$, then $\vv
\{\vv^*\}$. Thus, if $\vv \in M^C(\uu, \YY \gets \yy)$, then $\vv
	%joe18
	%= \vv^*$ and so $\vv \in M^C(\uu \YY \gets \yy; W \gets w)$ as
	%desired. Next suppose that $\YY \gets \yy \in \I'$. Then by definition
	= \vv^*$ so $\vv \in M^C(\uu, \YY \gets \yy; W \gets w)$, as
desired. Next, suppose that $\YY \gets \yy \in \I'$. Then by the definition
of $M^C$, we have $\<\YY \gets \yy\>(\XX = \vv[\XX] \wedge W = w) \in
	C$ for all finite subsets $\XX$ of $\WW$. By D3, it follows that
$\<\YY \gets \yy; W \gets w\>(\XX = \vv[\XX]) \in C$ for all finite
%joe18
        %joe26
        %
%subsets $\XX$ of $\WW$. But by definition of $M^C$, this implies that
%$\vv \in M^C(\YY \gets \yy; W \gets w)$ as desired.
subsets $\XX$ of $\WW$. But, by the definition of $M^C$, this implies that
%spencer31:
%$\vv \in M^C(\YY \gets \yy; W \gets w)$, as desired.
$\vv \in M^C(\uu, \YY \gets \yy; W \gets w)$, as desired.
%joe18

%joe34*: added argument for D6
%spencer40: I'm stunned this went through so cleanly. We must have
%been making changes right before the previous deadline and didn't
%realize this would just work. Thanks for remembering that! :)
%joe35: It did require a bit of work (basically, realizing that we
%could start with the canconical model M^c for all the properties
%other than acyclicity, and argue that it must satisfy acyclicity too
%(because of A6+)
%spencer41: It would have been too much work for me to remember how we planned to do it, that's for sure! Great work :)
Finally, suppose that $acyc \in \A$.  Then $A$ contains D6$^+$.
We want to show that the model
%spencer40: typo. Replaced $M^c$ by $M^C$ everywhere below.
%$M^c$ constructed as above.
$M^C$ constructed as above is acyclic.
Suppose that $X_1, X_2, \ldots$ is an ordering of the named variables,
$x^i_1, x^i_2, \ldots$ is an ordering of the named values, for each
named variable $X_i$, and $I_1, I_2, \ldots,$ is an ordering of the
named interventions.  (We can do this since all these sets are countable.)

Say that an ordering (i.e., permutation)
%spencer40: typo (I think?)

%joe35: definitely
%$Y_j, \ldots, Y_n$
$Y_1, \ldots, Y_n$
of $X_1, \ldots, X_n$ is
\emph{$n$-legal} if $\neg (Y_j \leadsto_{\I_n, U_1, \ldots, U_j, Y_1,
  %spencer40: missing paren
  %\ldots, Y_j} Y_{-i}$
\ldots, Y_j} Y_{-i}$)
for $j=1, \ldots,n$, where if $Y_j = X_{k}$, then
%spencer40: missing set bracket
%$U_j = \{x^k_1, \ldots, x^k_n$,
$U_j = \{x^k_1, \ldots, x^k_n\}$,
and $\I_n = \{I_1, \ldots,I_n\}$.
  We now construct a tree whose root is labeled by $\emptyset$ and
  whose nodes at depth $n$ are labeled by
  permutations of $X_1, \ldots, X_n$, and the children of a node
  labeled $Y_1, \ldots, Y_n$ are labeled by extensions of $Y_1,
  \ldots, Y_n$ to include $X_{n+1}$ (i.e., $X_{n+1}$ is slotted
  %spencer40: missing paren
  %somewhere into the permutation.
somewhere into the permutation.)
  Next remove all nodes that are not
  $n$-legal.
 The fact that $M^C$ satisfies all instances of D6$^+$ implies there has to
    be at least one node remaining at each depth. Moreover,
    all the ancestors of that node also remain (because if $(Y_1,
    \ldots, Y_n)$ is an $n$-legal ordering, then $(Y_1, \ldots,
    Y_{i-1}, Y_{i+1}, \ldots, Y_n)$ is easily seen to be an
    $(n-1)$-legal ordering).
    So, by Konig's Lemma, there is an infinite path starting from
    the root.  This path must be a sequence of
    legal orderings; it defines a total well-founded order $\prec^*$ on all the
    variables $X_1,
    %spencer40: is this duplicate comma standard?
%joe35: typo!
    %    X_2, \ldots, $,
        X_2, \ldots, $
    where $X_i < X_j$ if, for
    %spencer40: removing ``some''
    %some $n = \max(i,j)$,
    $n = \max(i,j)$,
    $X_i < X_j$ in the ordering at depth $n$ on the path.  Note that
    if $X_i < X_j$ in the ordering at depth $n$, then $X_i < X_j$ for
    all orderings on the path at depth $n' > \max(i,j)$, since the
    orderings are consistent.

    We claim that this ordering satisfies Acyc1.  For if not, there
    must be some $X \in \V$, some values
    %spencer40:
    %$x, x' of X$,
    $x, x'$ of $X$,
    %spencer40*: This Y is only for Acyc2! Changing to Acyc1.
    % some $Y$ such
    % that $Y \prec^* X$,
    some $I \in \I$, and some context $\uu$ such that
    %spencer40: Acyc1
%$M^C(\uu, I; X \gets x)[Y] \ne M^C(\uu, I, X \gets x')[Y].$
$M^C(\uu, I; X \gets x)[\V_{\prec_\uu X}] \ne M^C(\uu, I, X \gets x')[\V_{\prec_\uu X}].$
%spencer40: Acyc1
%Suppose that $X = X_i$ and $Y= X_j$.
%spencer40: Acyc1
    % Let $n$ be sufficiently large such that $n \ge \max(i,j)$, $I \in
    % \I_n$, $x, x' \in \{x^i_1, \ldots, x^i_n\}$, and
    % $M^C(\uu, I; X \gets x)[Y], M^C(\uu, I, X \gets x')[Y] \in
    % \{x^j_1, \ldots, x^j_n\}$.
    %joe35: you're saying that I \in \I_n twice
%spencer41: good catch! I meant to remove the other one, reverting.
   Let $n$ be sufficiently large that $I \in \I_n$, and that for all of
    % Let $n$ be sufficiently large that, for all of
    the finitely many $i$ such that $Y_i \preceq_\uu X$, the following
hold:
(1) $n \ge i$;
%joe35
%(2) $I \in \I_n$; (3) $x, x', M^C(\uu, I, X \gets x)[Y_i], M^C(\uu, I,
%spencer41:
%(2) $I \in \I_n$; and (3) $x, x', M^C(\uu, I, X \gets x)[Y_i], M^C(\uu, I,
and (2) $x, x', M^C(\uu, I, X \gets x)[Y_i], M^C(\uu, I,
X \gets x')[Y_i] \in \{x^i_1, \ldots, x^i_n\}$.
    This gives us a contradiction to the
    fact that the ordering at the $n$th level of the tree is
    $n$-legal.  Thus, the ordering must satisfy Acyc1,
    %spencer40: added
    and $M^C$ is acyclic,
    as desired.
    \eprf

%joe37: more informative title
    %    \section{Additional results}
    \section{Axioms D5 and D9}
    %spencer46:
    \label{appendix:d5-and-d9}

In this section, we look a little more carefully at axioms D5  and D9.
We start by making precise and proving the claim in the main text that
%joe9
%Halpern's version of D9,
the version of D9 in \cite{Hal20},
$$\<\YY \gets \yy\> true \land (\bor_{x \in \R(X)} [\YY \gets \yy](X =
		x)),$$
which we call HD9,
is equivalent to ours,
$$ \<\YY \gets \yy\> true \land (\<\YY \gets \yy\>\phi \rimp [\YY
		\gets \yy]\phi),$$
in  the presence of the other axioms.  It turns out that the
only other axioms we need are D0, D1, D2, D7, and D8.  So let
%joe17
%$AX_{eq}$ be the aiomatization consisting of these axioms and MP,
$AX_{EQ}$ be the axiomatization consisting of these axioms and MP,
\newcommand{\adnine}{AX_{\text{HD9}}}
\newcommand{\adournine}{AX_{\text{D9}}}
let $\adnine = AX_{EQ} \cup \{\text{HD9}\}$, and let
$\adournine = AX_{EQ} \cup \{\text{D9}\}$.

\begin{theorem} \label{theorem:d9-equivalent}
	%joe16*: simlified theorem statement, given the above
	%  %joe14*: We shoudl say explicitly which axioms we use (in particular,
	%%not D10) and give the axiom system a name.
	%  Let $AX_{eq}$ consist of D0, D1, D2, D7, and D8. Consider a finite
	%  signature $\S$. Then in the presence of $AX_{eq}$,
	%  Halpern's version of D9, namely
	%  \begin{equation} \label{eq:halpern-d9}
	%    \<\YY \gets \yy\> true \land (\bor_{x \in \R(X)} [\YY \gets \yy](X = x))  \tag{HD9}
	%  \end{equation}
	%  is equivalent to our version of D9, namely
	%  begin{equation}
	%spencer17: I like this form of the statement!
	For a finite signature $\S$, $\adnine$ is equivalent to $\adournine$.
\end{theorem}
\prf
%joe16: removed paragraph break
Since $\S$ is finite, without loss of generality we can assume that
$\V = \{X_1, X_2, \dots, X_n\}$ for some $n$.
%joe16: rewrote preamble
%First we prove \ref{eq:halpern-d9} implies \ref{eq:d9}.
% \newcommand{\adnine}{AX_{eq} \cup \{\text{\ref{eq:halpern-d9}}\}}
%\newcommand{\adnine}{AX_{\text{HD9}}}
%More formally, letting $\adnine = AX_{EQ} \cup \{\text{HD9}\}$;
%we show
It clearly
%spencer17:
% suffice
suffices
to show that each instance of D9 is provable from $\adnine$
and that each instance of HD9 is provable from $\adournine$.  For the
first claim, we show that
$$\adnine \vdash \<\YY \gets \yy\>true \land (\<\YY \gets \yy\>\phi
	\rimp [\YY \gets \yy]\phi).$$
%joe9
%Since one conjunct of \ref{eq:halpern-d9} is the formula $\<\YY \gets
%joe16: I think we can just say this
%$
Since one conjunct of HD9 is the formula $\<\YY \gets
	\yy\>true$,
it follows that
$$\adnine \shows \<\YY \gets \yy\>true.$$
%joe16: added this; it's good enough
%spencer17: makes sense.
(Formally, we are using D0, MP, and the observation that $a \land b
	\rimp a$ is a propositional tautology; in the sequel, we omit the
details of such obvious propositional reasoning.)
Thus, it suffices to show
$$\adnine \vdash \<\YY \gets \yy\>\phi \rimp [\YY \gets \yy]\phi.$$
%joe16*: this is unnecesarily complicated.  What I said above is
%enough for the readers of this part of the paper.
\commentout{
	This follows from D0, MP, and two propositional tautologies. From $a
		\land b \rimp a$ it follows \mbox{$\adnine \vdash \<\YY \gets
			\yy\>true$}. From $a \rimp (b \rimp (a \land b))$ it follows that if
	\mbox{$\adnine \vdash (\<\YY \gets \yy\>\phi \rimp [\YY \gets
				\yy]\phi)$} then \mbox{$\adnine \vdash \<\YY \gets \yy\>true \land
			(\<\YY \gets \yy\>\phi \rimp [\YY \gets \yy]\phi)$}.

	%spencer17*: It'd be really nice if we could work in the following
	%style. What do you think?
	%joe16*: You have to be really careful with this. If a system has what's called
	%generalization (from \phi infer []\phi), then the kind of
	%propositional reasoning you're referring to doesn't work:  Assume \phi.
	%Then aplying generalization, you can conclude []\phi.  From which you
	%can get \phi => []\phi, which is not sound.  You can do this here be
	%cause the *only* rule of inference you have is MP; but I'd rather not
	%say that.
	%spencer17: This makes sense. I'm not arguing that you can assume \phi, prove \psi, and infer \phi \rimp \psi, though.  For combining proof steps, I'm just saying that if you have $\shows \phi \rimp \rho$ and $\shows \rho \rimp \psi$, then you can conclude $\shows \phi \rimp \psi$.
	% Regardless, I agree that we don't need this extra detail in the paper.
	Notice that in an analogous way, all the usual logical introduction
	and elimination rules are at our disposal when reasoning about
	provability in an axiom system containing D0 and MP.
	What we argued above is a special case of the fact that if $\adnine \vdash a \wedge b$, then $\adnine \vdash a$ and $\adnine \vdash b$, and vice versa.
	Similar facts hold for the other logical connectives.
	For disjunctions, we can do proof-by-cases. That is, using D0, MP, and the propositional tautology $(a \vee b) \wedge (a \rimp c) \wedge (b \rimp d) \rimp (c \vee d)$, it is easy to show that if $\adnine \vdash (a \vee b)$ and $\adnine \vdash (a \rimp c)$ and $\adnine \vdash (b \rimp d)$, then $\adnine \vdash (c \vee d)$.
	Notice that proving a formula $f$ by cases on $a \vee b$ corresponds to taking $c = d = f$.
	For negations, we can do proof by contradiction using D0, MP, and the tautology $(a \rimp false) \dimp \neg a$.
	These imply that if $\adnine \vdash a \rimp false$, then $\adnine \vdash \neg a$, and vice versa.
	Finally, we can combine proof steps by using D0, MP, and the tautology $(a \rimp b) \wedge (b \rimp c) \rimp (a \rimp c)$. These imply if $\adnine \vdash a \rimp b$ and $\adnine \vdash b \rimp c$, then $\adnine \vdash a \rimp c$. We will use these facts implicitly going forwards.
}
Using an argument almost identical to that given to prove
\eqref{eq:equivalence-improved}, we can show that
\begin{equation} \label{eq:box-phi}
	%joe17: just changing punctuation
	%AX_{EQ} \shows [\YY \gets \yy](\phi \dimp \bor_{\vv \models \phi}  \V = \vv).
	AX_{EQ} \shows [\YY \gets \yy](\phi \dimp \bor_{\vv \models \phi}  \V = \vv);
\end{equation}
%joe17
%We leave details to the reader. (Note that our argument for
we leave details to the reader. (Note that our argument for

\eqref{eq:equivalence-improved} used only the axioms of $AX_{EQ}$.)

From D0, D7, D8, and MP, using standard modal logic arguments, it follows
\begin{equation} \label{and-of-diamond-box-implies-diamond-of-and}
	AX_{EQ} \shows \<\YY \gets \yy\>\phi \land [\YY \gets \yy](\phi \rimp \psi) \rimp \<\YY \gets \yy\>\psi.
\end{equation}
%joe17
%Combining this with the above equivalence, it follows
%$AX_{EQ} \shows \<\YY \gets \yy>\phi \dimp \<\YY \gets \yy>(\bor_{\vv
%$
Combining (\ref{eq:box-phi}) and
(\ref{and-of-diamond-box-implies-diamond-of-and}), it follows that
$AX_{EQ} \shows \<\YY \gets \yy\>\phi \dimp \<\YY \gets \yy\>(\bor_{\vv
		\models \phi} \V = \vv)$.
%joe17
%Applying Lemma~\ref{lemma:and-distrib-box}, it follows
Applying Lemma~\ref{lemma:and-distrib-box}, it follows that
\begin{equation} \label{eq:diamond-eq-bor}
	%joe17: typos
	%  AX_{EQ} \shows \<\YY \gets \yy>\phi \dimp \bor_{\vv \models \phi}
	%  \<\YY \gets \yy>(\V = \vv).
	AX_{EQ} \shows \<\YY \gets \yy\>\phi \dimp \bor_{\vv \models \phi}
	\<\YY \gets \yy\>(\V = \vv).
\end{equation}

%spencer17*: This isn't quite your outline, but I think it's correct. Let me know what you think.
Taking the conjunction over $\V$ of the second conjunct of HD9 gives
%joe17: displaying formula
%$AX_{HD9} \shows \band_{X \in \V} \bor_{x \in \R(X)} [\YY \gets \yy](X = x)$.
%Using D0 to distribute and over or and then applying
$$AX_{HD9} \shows \band_{X \in \V} \bor_{x \in \R(X)} [\YY \gets \yy](X = x).$$
Then using D0 to distribute and over or and then applying
Lemma~\ref{lemma:and-distrib-box} gives
%joe17: typo + displaying formula
%$AX_{HD9} \shows \bor_{\vv \in \R(\V)} [\YY \gets \yy](\V = \vv)}$.
$$AX_{HD9} \shows \bor_{\vv \in \R(\V)} [\YY \gets \yy](\V = \vv).$$

%joe17*: I don't like having a - in the superscript here.  Also
%slighlty simplified proof to get rid of the ``false''
%But for all $\vv^- \models \neg \phi$,
%$AX_{HD9} \shows \<\YY \gets \yy\>\phi \land [\YY \gets \yy](V =
%\vv^-) \rimp false$.
%To see this, observe that
%$AX_{HD9} \shows [\YY \gets \yy](\V = \vv^-) \rimp [\YY \gets
%\yy]\neg \phi$,

%but
%$AX_{HD9} \shows <\YY \gets \yy\>\phi \dimp \neg [\YY \gets \yy] \neg
%\phi$.$
%It follows that
%spencer18: This looks good.
%$
If $\vv \models \neg \phi$, then $\V=\vv \rimp \neg \phi$ is a
propositional tautology.  Thus, using D7 and D8,
we have that
$$AX_{HD9} \shows [\YY \gets \yy](\V =\vv) \rimp [\YY \gets \yy]\neg
	\phi.$$
Taking the contrapositive and using the fact that $\<\YY \gets
	\yy\>\phi$ is an abbreviation for $\neg [\YY \gets \yy]\neg \phi$, we
get that
$$AX_{HD9} \shows \<\YY \gets \yy\>\phi \rimp \neg [\YY \gets \yy](\V =\vv).$$
Since this is true for all $\vv$ such that $\vv \models \neg \phi$, it
follows from (\ref{eq:diamond-eq-bor}) that
%joe17: back to your proof; again, displaying formula
$$AX_{HD9} \shows \<\YY \gets \yy\>\phi \rimp \bor_{\vv \models \phi} [\YY \gets \yy](\V = \vv).$$
% However,
% \begin{equation} \label{eq:diamond-eq-rimp-box-eq}
%   AX_{HD9} \shows \<\YY \gets \yy\>(\V = \vv) \rimp [\YY \gets \yy](\V = \vv).
% \end{equation}
% To see this, observe that $AX_{HD9} \shows <\YY$
%joe17: displaying again
%But for all $\vv^+ \models \phi$,
%$AX_{HD9} \shows [\YY \gets \yy](\V = \vv^+) \rimp [\YY \gets \yy]\phi$.
%It follows
%$AX_{HD9} \shows \<\YY \gets \yy\>\phi \rimp [\YY \gets \yy]\phi$.
If $\vv \models \phi$, then standard arguments using D7 and D8 show that
$$AX_{HD9} \shows [\YY \gets \yy](\V = \vv) \rimp [\YY \gets \yy]\phi.$$
It follows
$$AX_{HD9} \shows \<\YY \gets \yy\>\phi \rimp [\YY \gets \yy]\phi,$$
as desired.

% $AX_{HD9} \shows (\<\YY \gets \yy\>(V = \vv) \land \neg [\YY \gets \yy](\V = \vv)) \dimp
% (\neg [\YY \gets \yy](\neg(V = \vv)) \land \neg [YY \gets \yy](\V = \vv))$.
% %
% Further, $AX_{HD9} \shows (\neg [\YY \gets \yy](\neg(V = \vv)) \land \neg [YY \gets \yy](\V = \vv)) \dimp \neg ([\YY \gets \yy](\neg(\V = \vv)) \lor [\YY \gets \yy](\V = \vv))$.
% %
% Using the fact that $AX_{EQ} \shows [\YY \gets \yy]\phi \lor [\YY \gets \yy]\psi \rimp [\YY \gets \yy](\phi \lor \psi)$, which follows from D0, D7, D8 and MP by standard modal logic arguments, we have
% $AX_{HD9} \shows (\<\YY \gets \yy\>(V = \vv) \land \neg [\YY \gets \yy](\V = \vv)) \dimp \neg [\YY \gets \yy](true)$.
% %
% Finally, using D0 and D8, we have
% $AX_{HD9} \shows [\YY \gets \yy](\true)$.
% Thus $AX_{HD9} \shows (\<\YY \gets \yy\>(V = \vv) \land \neg [\YY \gets \yy](\V = \vv)) \rimp false$, and the claim follows.
%
%spencer17: removing more complex proof
\commentout{
}

For the second claim, we show that
%joe26
%$AX_{D9} \shows \<YY \gets \yy\>true \wedge (\bor_{x \in \R(X)} [\YY
%$
 $AX_{D9} \shows \<\YY \gets \yy\>true \wedge (\bor_{x \in \R(X)} [\YY
  \gets \yy](X = x))$.
As before,  $AX_{D9} \shows \<\YY \gets \yy\>true$, so it suffices to show
$AX_{D9} \shows \bor_{x \in \R(X)} [\YY \gets \yy](X = x)$.
%joe17: removed paragraph break
%
Since $AX_{D9}$ includes D2,
$AX_{D9} \shows [\YY \gets \yy](\bor_{x \in \R(X)} (X = x))$.
Combining this with $AX_{D9} \shows \<\YY \gets \yy\>true$ using \eqref{and-of-diamond-box-implies-diamond-of-and}
gives
$AX_{D9} \shows \<\YY \gets \yy\>(\bor_{x \in \R(X)} (X = x))$.
%joe26*: Again, this refers to a lemma in the next section.  I suggest
%rearranging the sections.
%spencer28: DONE.
Applying Lemma~\ref{lemma:and-distrib-box}, it follows
$AX_{D9} \shows \bor_{x \in \R(X)} \<\YY \gets \yy\>(X = x)$.
But then by D9,
%joe17
%$AX_{D9} \shows \bor_{x \in \R(X)} [\YY \gets \yy](X = x)$
$AX_{D9} \shows \bor_{x \in \R(X)} [\YY \gets \yy](X = x)$,
as desired.
\eprf

% spencer15: moved this here.
% \item D5 follows from D2, D3, D6, D7, D8, D10, and MP.  (This was
% already essentially observed by Galles and Pearl
% \nciteyear{GallesPearl98}.) Indeed, as we show below
% %spencer15:
% % (Proposition~\ref{??}}),
% (Proposition~\ref{theorem:d5}),
% % \todo{}
% in the presence of these other axioms, D5
% holds even without the requirement that $\ZZ = \V -\{\XX,\YY\}$.
%spencer7: I can't immediately see how to prove this.
%joe7*:   Here's the idea: Using D6, there's some ordering on the
%variables.  Suppose without loss of generality that Y has no effect
%on W.  By D10, we can replace <> by [].  Thus, from <\XX<-\xx,
%Y<y>W=w & Z=z), we get [\XX <- \xx](W=w). Now by D2, there must be
%some \zz' such that [\XX <- \xx](W=w & \ZZ = \zz').  By D3, it follows
%that [\XX <- \xx, W <-x](\ZZ = \zz'). Since [\XX <- \xx, W <-x](Y = y
%& \ZZ =\zz), we must have \zz' = \zz.  Similarly, there must be a y'
%such that [\XX <- \xx](W=w & \ZZ = \zz & Y=y').  Again, by D3, we
%have [\XX <- \xx, W <- w](\ZZ = \zz & Y=y'), so we must have y=y'.
%Does that make sense?
%spencer8: Yes, this makes perfect sense, thank you!
%\zz
%joe37: added some glue, assuming we remove D5 from AX_{rec}.
We next show that D5 follows from $AX^+_{rec}(\S)$; in fact from a
subset of the axioms in $AX^+_{rec}(\S)$.
\begin{theorem} \label{theorem:d5}
	%  The axioms D2, D3, D6, and  D10, (along with the modal logic axioms
	%  D0, D7, D8 and MP), imply all instances of D5, that is, all
	%  instances of
	Let $AX^-$ consist of D0, D2, D3, D6, D7, D8, D10, and MP.
	Then $AX^-$ implies D5.  That is,
	\begin{equation}
		\begin{split}
			AX^- \vdash \   &(\<\XX \gets \xx; Y \gets y\>(W = w \wedge \ZZ = \zz) \\
			&\wedge \<\XX \gets \xx; W \gets w\>(Y = y \wedge \ZZ = \zz)) \\
			&\rimp \<\XX \gets \xx\>(W = w \wedge Y = y
%joe26
               %\wedge \ZZ = \zz      \end{split}
\wedge \ZZ = \zz.      \end{split}
	\end{equation}
\end{theorem}
\prf
% Let $\preceq$ be a total ordering of $\V$ such that for all $X, Y
% \in \V$, if $X \preceq Y$ then $A \vdash \neg (Y \leadsto X)$. This
% ordering exists because D6 states that the $\leadsto$ relation is
% acyclic.
%joe14
%D10(a) and D10(b) together imply that $\<\YY \gets \yy\> \dimp [\YY
%  \gets \yy]$.
Using
D10(a) and D10(b), we get that $AX^- \vdash \<\YY \gets \yy\> \phi \dimp [\YY
		\gets \yy]\phi$.
% spencer15*: this also needs the
% and-of-box-diamond-implies-diamond-of-and lemma ^
%joe14*: Not true; this just uses propositional equivalence (D0).
%That is if A<=>A', B<=>B', and C<=>C', then A&B => C is propositinally
%equivalent to A'&B' => C'
%spencer16*: ^ Are you sure? We need to combine [Y <- y]\phi and <Y <- y>true to get <Y <-y>\phi.
% Above you use the lemma I'm talking about (namely  <>\phi & [](\phi
%\psi) => <>\psi) to do that.
Hence, by D0,  it suffices to prove the propositionally equivalent statement
\begin{equation}
	\begin{split}
		%joe14
		%    &([\XX \gets \xx; Y \gets y](W = w \wedge \ZZ = \zz) \\
		AX^-\vdash\    &([\XX \gets \xx; Y \gets y](W = w \wedge \ZZ = \zz) \\
		&\wedge [\XX \gets \xx; W \gets w](Y = y \wedge \ZZ = \zz)) \\
		& \rimp [\XX \gets \xx](W = w \wedge Y = y \wedge \ZZ = \zz).
	\end{split}
\end{equation}
%joe14*: Spencer, this is actually not acceptable.  You can't use
%arguments like ``Assume''.  There is You're impicitly assuing some
%proof rules tha we haven't got (and don't in general have for modal
%logic, although in this case we do, since the only inference rule is
%MP).
Assume $[\XX \gets \xx; Y \gets y](W = w \wedge \ZZ = \zz) \wedge [\XX
		\gets \xx; W \gets w](Y = y \wedge \ZZ = \zz)$.
First observe that D6 implies $\neg (Y \leadsto W) \vee \neg (W \leadsto Y))$.
%joe14*: Again, you can't ``assume without loss of generality''.
%You're trying to show that (A\/ B) => C is provable.  So you need to
%show that A=> C is provable (which you've essentially done, show that
%B => C is provable (which follows by the same argument, and then use
%D0 and MP (using the fact that (A=>C) => ((B=>C) => (A\/B) => C) is
%propositionally valid).  Axiomatic derivations are painful!
%spencer15: this makes perfect sense.
% I was hoping that these steps might be implicit, based on our emails
% about proofs by contradiction in modal logic :)
%have to show tht AX^-
Assume without loss of generality that $\neg (Y \leadsto W)$.
We will use Lemma~\ref{lemma:and-distrib-box} frequently in this proof, to justify treating
$[\TT \gets \ttt](\phi \wedge \psi)$ as $[\TT \gets \ttt]\phi \wedge [\TT \gets \ttt]\psi$ and vice versa.
Thus from $[\XX \gets \xx; Y \gets y](W \gets w \wedge \ZZ = \zz)$ we get $[\XX \gets \xx; Y \gets y](W \gets w)$. Then $\neg (Y \leadsto W)$ implies $[\XX \gets \xx](W \gets w)$.

%joe14*: Again, this is not a legal argument.  You need to use the
%style of reasoning that I used above, to show that a disjuntion
%implies something.  Also, we don't ``have'' formulas, they are
%provable in AX^-.
%joe26
%Second, D2 implies that there must be some solution for $\ZZ$ in the
Second, D2 implies that $\ZZ$ must take on some value in the outcome
resulting from the
intervention $\XX \gets \xx$, that is, $\bor_{\zz \in \R(\ZZ)} \ZZ =
	\zz$.
  To see this, take the conjunction of instances of D2 over $Z \in \ZZ$, that is, $\band_{Z \in \ZZ} [\XX \gets \xx](\bor_{z \in \R(Z) }Z = z)$.
%spencer28:
%Then applying Lemma~\ref{lemma:and-distrib-box}
Then applying Lemma~\ref{lemma:and-distrib-box}(a)
and distributing and over or, we have $[\XX \gets \xx](\bor_{\zz' \in \R(\ZZ)} \ZZ = \zz')$, or equivalently by D10, $\<\XX \gets \xx\>(\bor_{\zz' \in \R(\ZZ)} \ZZ = \zz')$. Then applying
%spencer28:
%the dual form of Lemma~\ref{lemma:and-distrib-box},
Lemma~\ref{lemma:and-distrib-box}(b)
yields $\bor_{\zz' \in \R(\ZZ)} \<\XX \gets \xx\>(\ZZ = \zz')$.

Suppose without loss of generality that
$\<\XX \gets \xx\>(\ZZ = \zz^*)$, or equivalently by D10, $[\XX \gets \xx](\ZZ = \zz^*)$.
But then $[\XX \gets \xx](\ZZ = \zz^* \wedge W = w)$. By D3, this implies $[\XX \gets \xx; W \gets w](\ZZ = \zz^*)$.
However, we assumed $[\XX \gets \xx; W \gets w](Y = y \wedge \ZZ = \zz)$.
Thus $\zz^* = \zz$, and so $[\XX \gets \xx](\ZZ = \zz \wedge W = w)$.

Finally it remains to show $[\XX \gets \xx](Y = y)$.
Using D2 again, there must be $y^*$ such that $[\XX \gets \xx](W = w \wedge \ZZ = \zz \wedge Y = y*)$. Again, by D3, this implies $[\XX \gets \xx; W \gets w](\ZZ \gets \zz \wedge Y \gets y^*)$.
But we assumed $[\XX \gets \xx; W \gets w](Y = y \wedge \ZZ = \zz)$. It follows $y^* = y$. Thus $[\XX \gets \xx](W = w \wedge Y = y \wedge \ZZ = \zz)$ as desired.
\eprf

%spencer2: new theorems
%joe3*: I would cut this.  It's trivial and not a standard way of
%looking at things: Here's  the proof: AX^+(S) is sound and complete
%for SEMs with respect to the language $\L(S)$.  Now suppose that you
%restrict the set of allowed interventions to get the signature \S'.
%\L(S') is a sublanguage of \L(S).  Thus, a formula in \L(S') is valid
%for SEMs iff it's provable in AX^+(\S).
%spencer3*: This makes sense. I got here via a contrast with finite
%GSEMs satisfying AX^+(\S'), but you are right that viewed in
%isolation it's completely trivial.
%joe6*: cut
\commentout{
	\begin{theorem}
		\label{theorem:SEMs-some-interventions-not-allowed}
		For any signature $\S' = (\U, \V, \R, \I)$, consider the corresponding signature $\S = (\U, \V, \R, \I_{univ})$ where all interventions are allowed. The axioms $AX^+(\S)$ are sound and complete for SEMs with signature $\S'$, in the sense that the set of theorems of $AX^+(\S)$ that fall in the language $\L(\S')$ is exactly the set of causal formulas of $\L(\S')$ valid in all SEMs with signature $\S'$.
	\end{theorem}
	\prf
	%spencer2: Need to make this point clear when defining SEMs.
	%spencer2: Also, I feel like I could make it a lot more easily if I
	%define some better notation, e.g. SEM(\S) for the sems with signature
	%\S.
	%joe3: that notation might indeed be useful
	The SEMs $M$ with signature $\S$ and the SEMs $M'$ with signature $\S'$ are in one-to-one correspondence via the bijection
	$((\U, \V, \R, \I_{univ}), \F)) \leftrightarrow ((\U, \V, \R,
        \I), \F)$. This bijection preserves the solutions $M(\uu, I)$
        for any context $\uu$ and $I \in \I$. It follows that the
        bijection preserves the set of causal formulas in $\L(\S')$
        that the model satisfies. Hence, the set of causal formulas in
        $\L(\S')$ satisfied by all SEMs with signature $\S$ is exactly
        the set of causal formulas in $\L(\S')$ satisfied by all SEMs
        with signature $\S'$. In other words, all formulas in
        $\L(\S')$ valid for SEMs with signature $\S$ are valid for
        SEMs with signature $\S'$ and vice versa. But the theorems of
        $AX^+(\S)$ are exactly the valid formulas for SEMs with
        signature $\S$, so the theorems of $AX^+$ in $\L(\S')$ are
        exactly the formulas in $\L(\S')$ valid for SEMs. It follows
        that the theorems of $AX^+$ in $\L(\S')$ are exactly the valid
        formulas for SEMs with signature $\S'$.
	\eprf
}

\commentout{
\section{Acyclicity in GSEMs}
%joe27*: moved out of example and rewrote
%spencer28*: added
%spencer31: added
In SEMs, acyclicity is defined via the notion of
%joe29
%\emph{independence}. Recall from Section \label{sec:review} that given
\emph{independence}. Recall from Section~\ref{sec:review} that given
a SEM $M$ with endogenous variables $X$, $Y$, we say that \emph{$Y$ is
  independent of $X$} (in context $\uu$) if the structural equation
$F_Y(\uu, \dots)$ for $Y$ does not depend on $X$.

%spencer31:
%The definition of independence in SEMs
This definition of independence
has a very natural
interpretation in acyclic SEMs.
In acyclic SEMs, there is always a
  unique outcome, and given variables $X, Y$, if $Y$ is independent of
  $X$, then you cannot change $Y$ by intervening on $X$. That is, for
  any context $\uu$, the outcome $\vv$ of doing $I; X \gets x$ in
  context $\uu$ and the outcome $\vv'$ of doing $I; X \gets x'$ agree
  on $Y$ ($\vv[Y] = \vv'[Y]$).   The following examples show that this
  interpretation can break down in GSEMs,
  %spencer31: added
  %joe29
%  and that as a consequence, it is not obvious whether it is possible
  and that, as a consequence, it is not obvious whether it is possible
  to define acyclicity for GSEMs in a reasonable way.
  %spencer30*: No, the example is not taken from PH20. It is only the
  %little 'acyclic' example that is from PH20.
  % The first example is taken
  % from \cite{PH20} (although it used there for a different purpose.
%  general SEMs, which may have multiple solutions, in a very strong
%  sense.
%joe26*: It's not clear to me what the point of this example is, and
%how it shows some thing about independence.  THere needs to be a
%better story here, and it needs to be tied better into the story in
%the main text.
%spencer28*: You're totally right. The link was clear in my head, but
%definitely not clear on paper. I had to change the example, but I
%think I've managed to tie it into the story. At the least, this is
%what I had in mind.
\begin{example}
	\label{example:killing-solutions}
%joe27
        %  This example illustrates that this interpretation breaks down in
%  general SEMs, which may have multiple solutions, in a very strong
%  sense.
%  In this example,
  %spencer31:
  % We construct a family of SEMs in which
  We construct a SEM with unique solutions and a single context $\uu$ in which
%joe27
%  variables $V_1, \dots, V_n$ are independent of the boolean variable
  %spencer31:
    %variables $V_1, \dots, V_n$ are independent of the binary variable
the binary variable $V$ is independent of the binary variable
$T$, but intervening on $T$ can change
%spencer31:
%the outcomes restricted to $V_1, \dots, V_n$ in an arbitrary way.
the outcome for $V$.
%spencer31:
% That is, given arbitrary sets
% %joe28
% %    of outcomes $S_0, S_1$, we construct $M$ such that $M(\uu, T \gets
%     of outcomes $S_0, S_1$, we construct a GSEM $M$ such that $M(\uu, T \gets
%     0) = S_0$, but $M(\uu, T \gets 1) = S_1$.
%$
That is, we construct a SEM $M$ such that for the unique solution
$\vv \in M(\uu, T \gets 0)$, we have $\vv[V] = 0$,
but for the unique solution $\vv' \in M(\uu, T \gets 1)$,
we have $\vv'[V] \gets 1$.

The set $\V$ of endogenous variables consists of
%spencer31:
%$\WW =\{V_1, \ldots, V_n\}$, $\WW' = \{V'_1, \dots, V'_n\}$,
        %joe28: what is \R(V_i)?  Ifit doesn't matter, say so, or pick a range
        %Wny not make them all binary?
        %spencer31: They don't matter. I said above they could be arbitrary.
        %    with $\R(V'_i) = \R(V_i)$, in addition to binary variables
%spencer31: cut
% with $\R(V'_i) = \R(V_i)$,
%spencer31: added
$V$ and additional binary variables
  %spencer28:
  %$B, Z, Z'$.
  $T, B, Z, Z'$.
  The equations are
  \begin{align*}
    % \label{eq:1}
    %spencer28: added
    &T = 0, \\
    %spencer31:
    % &V_i = V'_i, \quad i = 1, \dots, n, \\
    % &V'_i = V_i, \quad i = 1, \dots, n, \\
    &V = V', \\
    &V' = V, \\
    %spencer28:
    &B =
      \begin{cases}
        %spencer31:
%       1, & \text{\parbox[t]{.4\textwidth}{
% %joe28*:
% %          $(T = 0 \wedge (V_1, \dots, V_n) \in S_0) \vee \newline
%           $\text{if } (T = 0 \wedge (V_1, \dots, V_n) \in S_0) \vee \newline
%           (T = 1 \wedge (V_1, \dots, V_n \in S_1))$}} \\
%$
%joe29
        %        1, & (T = 0 \wedge V = 0) \vee (T = 1 \wedge V = 1) \\
            1 &\mbox{if } (T = 0 \wedge V = 0) \vee (T = 1 \wedge V = 1), \\
%joe27*: this line is a little long
%joe29
            %            0, & \text{otherwise,} \\
                        0 & \text{otherwise,} \\
    \end{cases} \\
    %     &B =
    % \begin{cases}
    %   1, & (V_1, \dots, V_n) \in S,
    %   0, & (V_1, \dots, V_n) \notin S,
             % \end{cases} \\
    %spencer31: Flipping Z and Z' and interchanging the order so that it matches the order in which the equations are solved.
%     &Z = Z', \\
%     &Z'=
%     \begin{cases}
% %      1, & B = 1, \\
%       %      \neg Z, & B = 0. \\
%       1, & \text{if } B = 1, \\
%       1- Z, & \text{if } B = 0. \\
    %     \end{cases} \\
    &Z =
          \begin{cases}
%joe29
            %            1, & \text{if } B = 1, \\
%      1- Z', & \text{if } B = 0, \\
            1 & \text{if } B = 1, \\
      1- Z' & \text{if } B = 0, \\
    \end{cases} \\
    &Z' = Z.
  \end{align*}

  %!! should be independent but not hard to verify
  %spencer31: added
  Looking at these equations, it is clear that $V$ is independent of
  $T$, since $T$ does not appear in the equation for $V$. Indeed, at
  first glance, it is not obvious that $T$ has anything to do with
  $V$; $T$ only appears in the equation for $B$, and $V$ does not
  appear in the equation for $T$ at all. Nevertheless,
  it is not hard to verify that
%joe28: Aligning it this actually make it hard to parse
%  \begin{align*}
  %spencer31:
%   $$\begin{array}{ll}
%     % M(\uu, \emptyset) = \{(&\WW = \ww, \WW' = \ww, \\
%     %   &B = 1, Z = 1, Z' = 1) \\
%     %  \mid &\ww \in S\}.
% %joe28
% %  M(\uu, T \gets 0) = \{(&\WW = \ww, \WW' = \ww, \\
% %      &T = 0, B = 1, Z = 1, Z' = 1) \\
% %     \mid &\ww \in S_0\}.
%   %  \end{align*}
%     M(\uu, T \gets 0) = \\ \{(\WW = \ww, \WW' = \ww,
%       T = 0, B = 1, Z = 1, Z' = 1)
%       \mid \ww \in S_0\}.
%       \end{array}$$
  $M(\uu, T \gets 0) = \{\vv\}$ with $\vv[V] = 0$, and $M(\uu, T \gets 1) = \{\vv'\}$ with $\vv'[V] = 1$.
  %spencer31:
%   Indeed, if $V_1 = v_1, \dots, V_n = v_n$, then if $(v_1, \dots, v_n) \in S_0$, solving the remaining equations gives $V'_1 = v_1, \dots, V'_n = v_n, T = 0, B = 1, Z = 1, Z' = 1$,
%   but if $(v_1, \dots, v_n) \notin S_0$, then the equation for $B$ gives
%   %spencer28: typo
%   % $B = 1$,
%   $B = 0$,
% %joe28
% %  and so the equations for $Z, Z'$ are $Z = Z'$, $Z' = \neg Z$, which
%   and so the equations for $Z, Z'$ are $Z = Z'$, $Z' = 1- Z$, which
  %   clearly have no solution.
  %joe29
%  Indeed, if we try $V = 0$, then solving the remaining equations
  Indeed, if we set $V = 0$, then solving the remaining equations
  gives the unique solution $\vv$ specified by $V = 0$, $T = 0$, $V' =
%joe29
%0$, $B = 1$, $Z = 1, Z' = 1$. However, if we try $V = 1$, then we
  %$
0$, $B = 1$, $Z = 1, Z' = 1$. However, if we set $V = 1$, then we
  get $T = 0$, $V' = 1$, $B = 0$, and the remaining equations are $Z =
  1 - Z'$, $Z' = Z$, which clearly have no solution.
  %spencer28: added
  An identical argument shows that
%joe28
%  \begin{align*}
%      M(\uu, T \gets 1) = \{(&\WW = \ww, \WW' = \ww, \\
%      &T = 1, B = 1, Z = 1, Z' = 1) \\
%     \mid &\ww \in S_1\},
%    \end{align*}
  %spencer31:
% $$\begin{array}{ll}
%     M(\uu, T \gets 1) = \\
%     \{(\WW = \ww, \WW' = \ww,
%       T = 1, B = 1, Z = 1, Z' = 1)
%     \mid \ww \in S_1\},
%       \end{array}$$
  $M(\uu, T \gets 1) = \{\vv'\}$ with $\vv'[V] = 1$,
    as claimed.
    %joe30
    \bbox
\end{example}
  %spencer31: Rewrite.
This example shows that the intuition for independence already
%joe29*  I have a problem with the rest of this paragraph, since the
%point of the example was that the difficulty already arose when there
%were unique solutions (in cyclic SEMs).  So the story does not seem
%quite right.    My temptation would be to skip right to ``To
%illustrate ...''
%breaks down in SEMs with multiple solutions.
breaks down in cyclic SEMs, even if they have unique solutions.
%spencer33*: You're completely right, of course. What I was trying to get at here is that there *were* multiple solutions, but one of them got killed. That's how T can affect V even though V is not a descendent of T. If we define acyclicity in GSEMs in a way that allows for killing of solutions, then we get this problem.
% I think it's

% --as written, this doesn't make sense. But, that actually wasn't the point of this example! What I'm trying to get at here is that there *were* multiple solutions, but one of them got killed; that's how T can affect V even though V is not a descendant of T.  What I should do: replace the final example with an example showing that in the 'acyclic' model we define, we get these types of violations of independence. In fact, maybe we can just have the acyclic GSEM as our only example.
%joe29
  %  However, in SEMs, this   is not an issue for defining
%  acyclicity. This is because acyclic
  %  SEMs always have unique solutions.
There is no difficulty in acyclic SEMs, since solutions in acyclic
SEMs are bound to be unique.
%joe29
%However, although we have not attempted to define acyclicity for
%  GSEMs, we expect that any such definition will provide for acyclic
%  GSEMs having multiple solutions. Example \ref{example:infinite-sem}
Although we have not attempted to define acyclicity for
  GSEMs, we expect that any such definition will allow acyclic
  GSEMs to have multiple solutions. Example \ref{example:infinite-sem}
  below is a GSEM which we feel clearly should be considered
  acyclic. However, it does not have unique solutions. This poses a
  clear problem. To give a definition of acyclicity, we need a
  definition of independence that at least makes sense for acyclic
%joe29*: why is it obvious that for independence to make sense, we
%require unique solutions?  Do we have an example that shows this?
  models. But for independence to make sense, we seem to require
  unique solutions.

  \begin{example}
    \cite{PH20}
    \label{example:infinite-sem}
    Let $M_{chain}$ be the GSEM
    %spencer31:
    with binary endogenous variables
    $X_1, X_2, \dots$
    %spencer31:
    (and a single context $\uu$)
    whose outcomes under any
  intervention are the solutions to the structural equations
%joe27
  %  (possibly modified by the intervention)
  $X_1 = X_2$, $X_2 = X_3$,
  and so on. This GSEM is intuitively acyclic (the dependencies
  encoded by the equations only go from higher-numbered variables to
  lower-numbered ones,
  %spencer31: added
  so there is no cycle of dependencies). However, it has two solutions
  under the empty
  intervention; all zeros, and all ones.
  %joe30
  \bbox
  \end{example}

%spencer30: moved from below, edited
%spencer31:
To illustrate the problems that can arise when attempting to define
acyclicity for GSEMs more concretely, let us consider the following
intuitively reasonable definitions.
% One seemingly reasonable attempt at a definition of acyclicity in GSEMs is as follows.
\fullv{as proposed before the statement of Theorem \ref{theorem:GSEMs}}

\begin{definition}
  \label{definition:independence-GSEM}
%joe28*: I thought that the example above was meant to show that
%defining independence in GSEMs would be hard.  But now you've just
%given a definition.  So I'm still puzzled as to what the previous
%example is showing.  It seems like a better story might be ``To
%define acyclicity we need to define dependence and independence.  But
%defining independence is not so easy.  First, we show that
%independence doesn't work the way you think it should, even
%in SEMs that are not acyclic.  Consider the following example.  [[I
%would simplify the example, replace the set W by the singleton V and
%the set W' by the singleton V', and making both of them binary variables.]]
%Observe that, under any reasonable notion of independence T is
%independent of V.  Yet if T<-0, then V=0 and if T<-0, the V=1.  Go on
%Go on to say that another example of independence might be ... (Are T
%and V indpendent unde this defintion).
  %spencer31: typo! cutting below line
  % T is independent of of W, then changing the value of T would no
%affect W.  Then I wold rewrite the example above so that W and W' are
%singletons (with just V and V'), which should be binary variables,
%and replace S_0 by and S_1 by 1.  Afte you give the equations,
%observe that T and V are independent in any reasonable sense of he word
  Given a GSEM $M$ with endogenous variables $X, Y$, \emph{$Y$ is
      independent of $X$} (in context $\uu$) if there are no $x, x',
    \WW, \ww$ such that the outcomes $M(\uu, \WW \gets \ww; X \gets
    x)$ restricted to $Y$ differ from the outcomes $M(\uu, \WW \gets
    \ww; X \gets x')$ restricted to $Y$.
\end{definition}

\begin{definition}
  \label{definition:acyclic-GSEM}
 $M$ is \emph{acyclic} if for all contexts, there is a total ordering $\prec_\uu$ of the variables with $X \prec Y$ implying that $X$ is independent of $Y$ in context $\uu$.
\end{definition}

%spencer30: added
%joe29: If we cut the example as I suggested above, we shold ct the
%next sentence too.
It is easy to see that under these definitions, the GSEM $M_{chain}$ presented in Example \ref{example:infinite-sem} above is acyclic.
%spencer31: added
Furthermore, these definitions rule out the problem presented by Example \ref{example:killing-solutions}, in the sense that if we view Example \ref{example:killing-solutions} as a GSEM, and apply Definition \ref{definition:independence-GSEM}, we find that $V$ is in fact not independent of $T$. Thus it is perfectly reasonable that intervening on $T$ can change the value of $V$.

%spencer31:
% However, the final example of this section shows that this definition leads to unintuitive results in GSEMs.
However, the final example of this section shows that, as expected from the preceding discussion, this proposed definition provides for acyclic GSEMs in which independence does not behave as it should intuitively behave.

\begin{example}
	\label{example:switching-values}
  Define a GSEM $M$ with binary variables $A, B, C$, allowed
  interventions $\I = \{C \gets 0, C \gets 1\}$, a single context
%joe26
  %  $\uu$, and solutions as follows:
%joe27
  %  $\uu$, and the following outcomes:
    $\uu$, and the outcomes
  %spencer28: typo
  %\[M(\uu, C \gets 0) = (0, 0, 0), (1, 1, 0)\]
%joe27
  %  \[M(\uu, C \gets 0) = \{(0, 0, 0), (1, 1, 0)\}\]
    \[M(\uu, C \gets 0) = \{(0, 0, 0), (1, 1, 0)\} \mbox{ and }\]
%joe26
%  \[M(\uu, C \gets 1) = (0, 1, 1), (1, 0, 1)\]
%spencer28: typo
    %\[M(\uu, C \gets 1) = (0, 1, 1), (1, 0, 1),\]
\[M(\uu, C \gets 1) = \{(0, 1, 1), (1, 0, 1)\},\]
  where $(a, b, c)$ is short for $(A = a, B = b, C = c)$.
  It is easy to see that under Definition \ref{definition:independence-GSEM}, all
  ordered pairs of the variables $A, B, C$ are independent. In
  particular, $A$ and $B$ are independent of $C$, and so $M$ is
%joe26
%  acyclic. However, this language belies our intuition. Even though
    acyclic. However, even though
  $A$ and $B$ are independent of $C$, setting $C$ changes the possible
  values for $(A, B)$ from $(0, 0), (1, 1)$ to $(0, 1), (1, 0)$.
  %spencer30: cut, this was mentioned above
%   It is
%   worth noting that this can also occur in SEMs, but
%   %spencer28: added
%   as noted in Example \ref{example:killing-solutions},
%   it cannot occur in acyclic SEMs. This is because acyclic SEMs always have unique
% %joe27: ??
% %  outcomes; see
%   %spencer30: just a typo. Thanks!
%     outcomes.
  %joe30
  \bbox
\end{example}

%spencer31: added
This example shows that Definitions \ref{definition:independence-GSEM}
and Definition \ref{definition:acyclic-GSEM} have clear
problems. Based on the examples presented above, we think it is
%joe29
%unlikely that good definitions of independence and acyclicity for
%GSEMs, that avoid these problems, exist  (although of course, we
unlikely that  useful definitions of independence and acyclicity for
GSEMs  exist  (although, of course, we
cannot definitively rule out the possibility of finding such
definitions). Therefore, although a class of infinite GSEMs analogous
to acyclic SEMs would be of definite interest,  we do not attempt to
define such a class in this paper.
} %end commentout
} %end fullv

%joe39
\fullv{\bibliography{joe}}

\begin{thebibliography}{11}
\providecommand{\natexlab}[1]{#1}

\bibitem[{Alur et~al.(1992)Alur, Courcoubetis, Henzinger, and
  Ho}]{alur_hybrid_1992}
Alur, R.; Courcoubetis, C.; Henzinger, T.~A.; and Ho, P.-H. 1992.
\newblock Hybrid automata: {An} algorithmic approach to the specification and
  verification of hybrid systems.
\newblock In \emph{Hybrid Systems}, 209--229. Springer.

\bibitem[{Beckers and Halpern(2019)}]{BH19}
Beckers, S.; and Halpern, J.~Y. 2019.
\newblock Abstracting causal models.
\newblock In \emph{Proc.~Thirty-Third AAAI Conference on Artificial
  Intelligence (AAAI-19)}.
\newblock The full version appears at arxiv.org/abs/1812.03789.

\bibitem[{Galles and Pearl(1998)}]{GallesPearl98}
Galles, D.; and Pearl, J. 1998.
\newblock An axiomatic characterization of causal counterfactuals.
\newblock \emph{Foundation of Science}, 3(1): 151--182.

\bibitem[{Halpern(2000)}]{Hal20}
Halpern, J.~Y. 2000.
\newblock Axiomatizing causal reasoning.
\newblock \emph{Journal of A.I. Research}, 12: 317--337.

\bibitem[{Halpern(2016)}]{Hal48}
Halpern, J.~Y. 2016.
\newblock \emph{Actual Causality}.
\newblock Cambridge, MA: MIT Press.

\bibitem[{Halpern and Pearl(2005)}]{HP01a}
Halpern, J.~Y.; and Pearl, J. 2005.
\newblock Causes and explanations: a structural-model approach. {P}art {II}:
  {E}xplanations.
\newblock \emph{British Journal for Philosophy of Science}, 56(4): 889--911.

\bibitem[{Halpern and Peters(2021)}]{HP21}
Halpern, J.~Y.; and Peters, S. 2021.
\newblock Reasoning about causal models with infinitely many variables.
\newblock Available on arxiv.

\bibitem[{Laurent, Yang, and Fontana(2018)}]{laurent_counterfactual_2018}
Laurent, J.; Yang, J.; and Fontana, W. 2018.
\newblock Counterfactual resimulation for causal analysis of rule-based models.
\newblock In \emph{Proc.~Twenty-Seventh International Joint Conference on
  Artificial Intelligence (IJCAI '18)}, 1882--1890.

\bibitem[{Pearl(2000)}]{pearl:2k}
Pearl, J. 2000.
\newblock \emph{Causality: Models, Reasoning, and Inference}.
\newblock New York: Cambridge University Press.

\bibitem[{Peters and Halpern(2021)}]{PH20}
Peters, S.; and Halpern, J.~Y. 2021.
\newblock Causal modeling with infinitely many variables.
\newblock Available on arxiv.

\bibitem[{Rubenstein et~al.(2017)Rubenstein, Weichwald, Bongers, Mooij,
  Janzing, Grosse{-}Wentrup, and Sch{\"{o}}lkopf}]{Rub17}
Rubenstein, P.~K.; Weichwald, S.; Bongers, S.; Mooij, J.~M.; Janzing, D.;
  Grosse{-}Wentrup, M.; and Sch{\"{o}}lkopf, B. 2017.
\newblock Causal consistency of structural equation models.
\newblock In \emph{Proc.~33rd Conference on Uncertainty in Artificial
  Intelligence ({UAI} 2017)}.

\end{thebibliography}
\end{document}